\documentclass[runningheads]{llncs}

\usepackage{eccv}

\usepackage{eccvabbrv}

\usepackage{graphicx}
\usepackage{booktabs}
\usepackage{arydshln}
\usepackage{mathtools}
\usepackage[dvipsnames]{colortbl}
\usepackage[accsupp]{axessibility}  %
\usepackage{multirow}
\usepackage{wrapfig}
\usepackage{amsmath}
\usepackage{cases}
\usepackage{enumitem}
\usepackage{adjustbox}
\usepackage{makecell}
\usepackage[stable]{footmisc}
\DeclareMathOperator*{\E}{\mathbb{E}}

\newcommand{\colorfirst}{255, 153, 153}
\newcommand{\colorsecond}{255, 204, 153}
\newcommand{\colorthird}{255, 255, 153}
\definecolor{colorfirst}{RGB}{255, 153, 153}
\definecolor{colorsecond}{RGB}{255, 204, 153}
\definecolor{colorthird}{RGB}{255, 255, 153}

\definecolor{colorfirst}{RGB}{255, 153, 153}
\definecolor{colorsecond}{RGB}{255, 204, 153}
\definecolor{colorthird}{RGB}{255, 255, 153}

\usepackage[pagebackref,breaklinks,colorlinks,citecolor=eccvblue]{hyperref}

\usepackage{orcidlink}

\begin{document}

\title{SplatFields: Neural Gaussian Splats for\\Sparse 3D and 4D Reconstruction}
\titlerunning{SplatFields} 

\author{Marko Mihajlovic\inst{1}\thanks{Partly done during an internship at Meta.}\orcidlink{0000-0001-6305-3896} \and
Sergey Prokudin\inst{1,3}\orcidlink{0000-0001-6501-8234} \and
Siyu Tang\inst{1}\orcidlink{0000-0002-1015-4770} \and
Robert Maier\inst{2}\orcidlink{0000-0003-4428-1089} \and
Federica Bogo\inst{2}\orcidlink{0009-0003-4991-185X} \and
Tony Tung\inst{2}\orcidlink{0000-0002-0824-0960} \and Edmond Boyer\inst{2}\orcidlink{0000-0002-1182-3729}
\\ETH Zürich$^1$\,\,  Meta Reality Labs$^2$ \,\, Balgrist University Hospital$^3$
}
\authorrunning{M.~Mihajlovic et al.}
\institute{
\vspace{-0.25em}
{\href[pdfnewwindow=true]{https://markomih.github.io/SplatFields}{\nolinkurl{markomih.github.io/SplatFields}} }
\vspace{-1.75em}
}

\maketitle

\begin{abstract} 

Digitizing 3D static scenes and 4D dynamic events from multi-view images has long been a challenge in computer vision and graphics. Recently, 3D Gaussian Splatting (3DGS) has emerged as a practical and scalable reconstruction method, gaining popularity due to its impressive reconstruction quality, real-time rendering capabilities, and compatibility with widely used visualization tools. However, the method requires a substantial number of input views to achieve high-quality scene reconstruction, introducing a significant practical bottleneck. This challenge is especially severe in capturing dynamic scenes, where deploying an extensive camera array can be prohibitively costly. In this work, we identify the lack of spatial autocorrelation of splat features as one of the factors contributing to the suboptimal performance of the 3DGS technique in sparse reconstruction settings. To address the issue, we propose an optimization strategy that effectively regularizes splat features by modeling them as the outputs of a corresponding implicit neural field. This results in a consistent enhancement of reconstruction quality across various scenarios. Our approach effectively handles static and dynamic cases, as demonstrated by extensive testing across different setups and scene complexities. 


\end{abstract}

\section{Introduction}  \label{sec:intro}

\begin{figure}[t]
    \begin{center}
        \includegraphics[width=\textwidth]{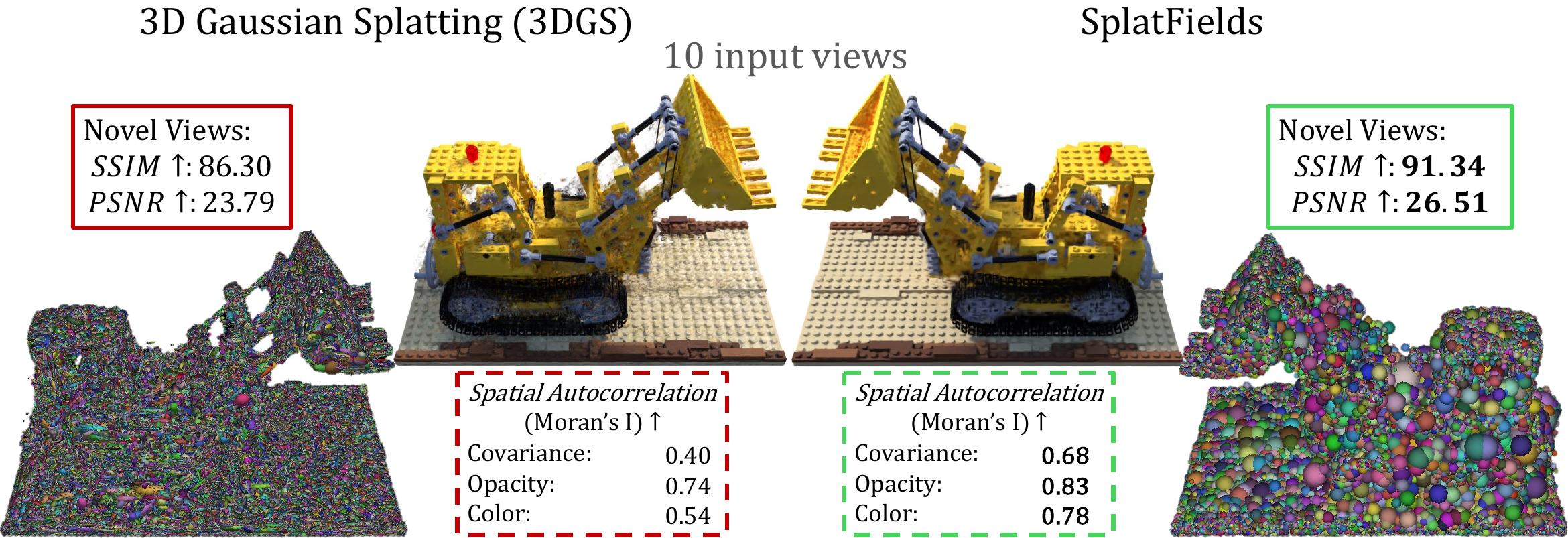}
    \end{center}
    \caption{\textbf{SplatFields} regularizes 3D Gaussian Splatting (3DGS)~\cite{GaussianSplatting} by predicting the splat features and locations via neural fields to improve the reconstruction under unconstrained sparse views. We measure spatial autocorrelation (Moran's I~\cite{moran1950notes}) of splat features in the local neighborhoods to assess their similarity and observe that better reconstruction quality achieved by our method corresponds to higher Moran's I. The figure presents the results of a static reconstruction from ten calibrated images from Blender dataset~\cite{mildenhall2020nerf}. Metrics are reported on the full test set; the rendered view is a novel view. 
    }
    \label{fig:teaser}
\end{figure}

Building a realistic replica of static and dynamic environments can revolutionize the world by transforming the way we interact, work, and engage online~\cite{singh2021digital}. This ambitious vision has motivated a surge in recent research to develop new representations and rendering techniques that allow for comprehensive and photorealistic capture and reconstruction of scenes from multi-view imagery.

Recent advancements, notably the introduction of Neural Radiance Fields (NeRF)\cite{mildenhall2020nerf}, have shown exceptional quality in photorealistic 3D reconstruction from casually captured images. 
This success comes from modeling a 3D scene as a neural field~\cite{xie2022neural} and optimizing it through volume rendering techniques. 
The parameterization of the rendering volume using a continuous differentiable field presents several benefits. 
It enables a compact representation of the scene's geometry and appearance through neural network weights, offering a more practical alternative to explicit volume modeling, which is often unfeasible. 
Crucially, for the focus of this work, the continuous nature and the spectral bias~\cite{rahaman2019spectral} of Multi-Layer Perceptrons (MLPa) introduce \textit{a spatial bias}—nearby primitives are likely to exhibit similar features as predicted by the neural field MLP. This concept of implicitly modeling spatiotemporal signals has captured the research community's attention in recent years~\cite{gao2022nerf}, marking a significant shift in methods for 3D scene reconstruction and novel view synthesis. A substantial portion of research has also focused on adapting these methods for sparse view setups~\cite{yu2021pixelnerf, niemeyer2022regnerf} and enhancing training and rendering efficiency~\cite{muller2022instant, chen2022tensorf, peng2023representing}.

3D Gaussian Splatting (3DGS) \cite{GaussianSplatting} offers an alternative 3D reconstruction framework using point-based rasterization rather than computationally demanding volume rendering. The method quickly gained traction within both the computer vision and graphics communities due to its real-time rendering capabilities, potential compatibility with the standard rasterization pipelines, and the intuitive way of editing and combining the reconstructed scenes. This makes 3DGS a practical and scalable solution that is currently being rapidly adopted and supported by many 3D development platforms and visualization tools \cite{survey3dgs1,survey3dgs2,survey3dgs3}. 

3D Gaussian Splatting represents the 3D scene as a set of unordered 3D Gaussian primitives, rendered from arbitrary views via rasterization, akin to traditional point splatting techniques~\cite{botsch2005high,zwicker2001ewa,zwicker2001surface}. Each rendering primitive comprises trainable parameters such as position, orientation, scale, color, and opacity, which are optimized by rendering the representation with respect to multi-view input images. The flexible parameterization, coupled with the efficient rasterization framework, is key to high-quality novel view synthesis results at scale. However, the flexibility of rendering primitives comes at the cost of requiring a large number of input views to fully constrain the optimization process, making Gaussian splatting unsuitable for more practical captures from sparse views.

We analyze the performance of 3DGS and its 4D variants~\cite{yang4DGS,yang2023deformable3dgs} in sparse input view scenarios. 
We first show that splat-based techniques, with their independently modeled set of rendering primitives, are particularly vulnerable to the training view overfitting in such cases (see Fig.~\ref{fig:teaser}). 
In contrast, volumetric rendering techniques \cite{tewari2022advances} which imply shared feature representations appear to be more robust in such scenarios as demonstrated in \cite{ResFields}, at the expense of a significantly increased training time and suboptimal rendering efficiency. 
This key observation provides the basis for the method introduced in this work.

Our key idea is to regularize the behavior of independent Gaussian primitives by utilizing neural networks that regress splat features at different levels. First, inspired by~\cite{zerorf}, we aim to enforce the spatial bias through a hierarchical convolutional decoder~\cite{peng2020convolutional} that outputs a tri-plane representation \cite{chan2022efficient} of deep features associated with each splat. Please note that the tri-plane representation and the associated network are utilized only during the optimization phase to constrain the attributes of the Gaussian primitives; both are discarded thereafter for accelerated rendering and compatibility with established splat rasterization pipelines. The produced deep splat features are then utilized to condition neural fields \cite{xie2022neural} that model the geometric and appearance properties of Gaussian splats at various locations and time steps. This design is equipped with positional encoding \cite{mildenhall2020nerf} to represent high-frequency details while retaining the good spatial properties required to regularize Gaussian splatting.

We thoroughly analyze our representation (dubbed \textit{SplatFields}) and demonstrate its superior reconstruction quality under sparse input views compared to alternative 3D Gaussian splatting techniques \cite{sugar,lightgaussian,mipgs,GaussianSplatting}. We further present an effective extension of our optimization framework to model dynamic 4D scenes and propose a new forward-flow field formulation to model the dynamics of Gaussian splats, warping rendering primitives into the observation space. We observe that existing techniques that model 3D splat deformations either lack the modeling capacity due to simplified assumptions on scene motion \cite{yang2023deformable3dgs} or have an insufficient spatial bias in the model \cite{wu4dgaussiansRealTime}, leading to suboptimal performance in sparse setups. Therefore, we introduce a forward-flow neural network for 3D Gaussians based on the recent ResFields MLP architecture \cite{ResFields}. Our method outperforms recent baselines \cite{yang4DGS,yang2023deformable3dgs,wu4dgaussiansRealTime} while retaining the key properties of Gaussian splatting, such as rendering efficiency and compatibility with existing frameworks.
In summary, our key contributions are:
\begin{itemize}[noitemsep,topsep=0pt] %
    \item We propose a novel optimization strategy, named \textit{SplatFields}, which introduces spatial bias into the 3D Gaussian Splatting technique to stabilize the optimization process under sparse views.
    \item We extend our formulation to dynamic scenes, demonstrating superior reconstruction quality compared to recent state-of-the-art methods \cite{yang2023deformable3dgs,wu4dgaussiansRealTime,yang4DGS}.
    \item We provide a detailed analysis of various modeling strategies, confirming the optimality of our framework for the tasks of sparse multi-view reconstruction.
\end{itemize}
The code is publicly available: \href[pdfnewwindow=true]{https://markomih.github.io/SplatFields}{\nolinkurl{markomih.github.io/SplatFields}}.

\section{Related Work} \label{sec:related_work} 

\textbf{Implicit volumetric rendering} \cite{drebin1988volume}\textbf{.} Novel View Synthesis (NVS) enables the generation of new images from arbitrary viewpoints using a given set of input images \cite{mildenhall2020nerf,tewari2022advances}. Over the past few years, the predominant method of choice for NVS has been the Neural Radiance Field (NeRF) \cite{mildenhall2020nerf}, which represents a 3D scene as a continuous neural field \cite{xie2022neural}. This field takes as input a location and view direction and predicts color and density. Then, the color of a pixel from an arbitrary viewpoint is rendered by casting a ray and employing volume rendering, which requires sampling multiple points along the ray and converting them into color and density values by querying the neural field. Numerous extensions have been proposed to handle various scenarios and setups such as sparse view reconstruction~\cite{niemeyer2022regnerf,yu2021pixelnerf,keypointnerf,ssdnerf,jain2021putting,deng2022depth,zhang2021ners,deng2023nerdi}, dynamic~\cite{DNeRF,DyNeRF,DCTNeRF,HyperNeRF,Nerfies,NDR} and large scale unbounded scenes~\cite{barron2022mip,tancik2022block}. However, the implicit volume rendering process is inherently expensive due to the large number of sampled points whose predictions need to be integrated. Despite recent efforts to accelerate NeRFs \cite{muller2022instant,chen2022tensorf,chan2022efficient,li2023nerfacc,reiser2021kilonerf,yu_and_fridovichkeil2021plenoxels}, achieving interactive rendering capabilities for regular scenes remains challenging without additional post-processing or compression \cite{yariv2023bakedsdf,yu2021plenoctrees,merf,smerf}. 

\textbf{Point-based rendering.} The limitations of volumetric rendering methods have led to a resurgence in point-based techniques~\cite{alexa2004point}. The seminal work by \cite{grossman1998point} introduced the rasterization of fixed-size, unstructured point samples for NVS. However, this naive rendering approach often results in aliasing artifacts and images with holes. These issues have been partially addressed by employing splatting techniques, where points are rendered with extended sizes to cover multiple pixels, using shapes like circular ellipsoids or surfels~\cite{zwicker2001ewa,zwicker2001surface,botsch2005high}. The era of deep learning led to a new wave of point-based neural rendering methods which allowed differentiable point rendering~\cite{wiles2020synsin,ruckert2022adop,yifan2019differentiable,lassner2021pulsar} and combined point-based rasterizers with 2D convolutional networks~\cite{aliev2020neural,KPLD21,deepsurfels}.

3D Gaussian Splatting (3DGS~\cite{GaussianSplatting}) utilizes the volumetric composition of ordered splats to merge the advantages of volumetric representations with exceptional real-time rendering capabilities. Gaining rapid popularity due to its efficiency, 3DGS has been incorporated into a wide range of downstream tasks~\cite{survey3dgs1,survey3dgs2}. Modifications to the original framework have enhanced its robustness to novel views~\cite{mipgs}, improved geometry reconstruction~\cite{sugar}, and reduced model size~\cite{lightgaussian}. Nevertheless, 3DGS's dependence on numerous \textit{independent} splatting primitives necessitates a large number of views for effective optimization, impacting its performance in sparse view reconstructions. Our work introduces neural networks to regularize splat behavior by regressing splat features based on their 3D location, introducing the spatial autocorrelation bias which substantially enhances reconstruction in sparse scenarios, as demonstrated in our experiments.

\textbf{Dynamic Gaussian splatting.} Several recent modifications of 3DGS have been proposed to extend its capabilities to dynamic sequences. The dynamic 3DGS~\cite{luiten2023dynamic} extends the basic pipeline by optimizing the motion of each splatting primitive and the change in its corresponding features. Although regularizations such as the enforcement of local rigidity and isometry losses~\cite{kilian2007geometric,prokudin2023dynamic} help stabilize the learning process to a certain extent, the overall pipeline still requires a large number of input views for each step and struggles to reconstruct scenes faithfully from sparse observations. The works closest to ours are~\cite{yang2023deformable3dgs,yang4DGS}, which utilize MLPs to model time-dependent deformations of Gaussian splats. However, the single MLP used for modeling deformation in~\cite{yang2023deformable3dgs} lacks the capacity to represent non-trivial scene dynamics, while the parameterizations used in \cite{yang4DGS,wu4dgaussiansRealTime} lack a substantial spatial and temporal autocorrelation bias, leading to suboptimal reconstructions in sparse view scenarios. We thoroughly analyze the behavior of the aforementioned methods and compare them with our approach on several dynamic scenes of varying complexity and view sparsity. We demonstrate that our model, which combines a triplane-based CNN generator~\cite{chan2022efficient} for splat features with ResFields-based~\cite{ResFields} dynamics modeling, offers the optimal combination of expressivity and robustness in sparse capture scenarios.

A growing body of work also addresses dynamic scenes~\cite{das23npg,li2023spacetime,yu2023cogs,huang2023sc,lu2024gagaussian,lin2024gaussian} and template-based approaches~\cite{lei2023gart} for modeling full-body~\cite{qian20233dgs,xu2023gaussian,hu2023gaussianavatar,hugs,hu2024gauhuman,pang2024ash} and head avatars~\cite{saito2024rgca,qian2023gaussianavatars}. In contrast, our model is capable of handling unbounded, topologically varying generic dynamic scenes.

\section{Preliminaries: 3D Gaussian Splatting}  \label{sec:preliminaries} 

In the following, we provide a brief overview of the Gaussian splatting rendering technique, which is a fundamental building block of our model.

\textbf{\emph{Scene representation.}} 3DGS~\cite{GaussianSplatting} parametrizes the 3D scene via static 3D Gaussian primitives $\{\mathcal{G}_k\}_{k=1}^K$ that contain the geometric and appearance information. 
These rendering primitives are utilized for efficient differentiable rasterization-based volume splatting. 

The geometry of each Gaussian splat $\mathcal{G}_k$ is defined by the mean location $\mathbf{p}_k \in \mathbb{R}^{3 \times 1}$, the opacity value $\alpha_k \in [0, 1]$, and the covariance matrix $\mathbf{\Sigma}_k \in \mathbb{R}^{3 \times 3}$ defined in the world space. Each splatting primitive $\mathcal{G}_k$ then induces the following Gaussian distribution in 3D space:
\begin{equation}
    \mathcal{G}_k(\mathbf{x}) \propto \exp \left(   -\frac{1}{2}(\mathbf{x}-\mathbf{p}_k)^T \mathbf{\Sigma}_k^{-1} (\mathbf{x}-\mathbf{p}_k)\right)\,, 
\end{equation}
where the covariance matrix is modeled by the scaling vector $\mathbf{s}_k \in \mathbb{R}^{3}$ and the rotation matrix $\mathbf{O}_k \in \mathbb{R}^{3 \times 3}$ (parameterized via quaternions) to ensure positive semi-definiteness:
\begin{equation} \label{eq:gaussian_cov}
    \mathbf{\Sigma}_k = \mathbf{O}_k \mathbf{s}_k\mathbf{s}_k^T \mathbf{O}_k^T\,.
\end{equation}

The appearance of splats is view-dependent and described by $C$ coefficients that are converted to color $\mathbf{c}_k \in \mathbb{R}^3$ via spherical harmonics, similar to~\cite{yu2021plenoctrees}. 

\textbf{\emph{Rendering.}} 
Given an arbitrary camera viewpoint described by the rotation $\mathbf{R} \in \mathbb{R}^{3 \times 3}$ and translation $\mathbf{t} \in \mathbb{R}^{3 \times 1}$, we can obtain the 2D coordinates of the splat center $\mathbf{p}_k^{\prime} \in \mathbb{R}^2$ on the image plane:
\begin{equation}
    \mathbf{p}_k^{\prime} = (\mathbf{R}   \mathbf{p}_k + \mathbf{t})_{1:2}/(\mathbf{R}   \mathbf{p}_k + \mathbf{t})_{3}.
\end{equation}

Further, we can obtain the 2D projection $\mathbf{\Sigma}_k^{\text{2D}}$ of the covariance matrix on the image plane:
\begin{eqnarray} 
 \mathbf{\Sigma}_k^{\text{2D}} = \left(\mathbf{J}_k \mathbf{\Sigma}_k^{\prime} \mathbf{J}_k^T\right)_{1:2,1:2}\, \in \mathbb{R}^{2 \times 2}, \text{  where  } \mathbf{\Sigma}_k^{\prime} =  \mathbf{R}   \mathbf{\Sigma}_k \mathbf{R}^T,
\end{eqnarray} 
and the Jacobian $\mathbf{J}_k \in \mathbb{R}^{3 \times 3}$ is an affine approximation to the projective transformation (see~\cite{zwicker2001ewa} for details).  The subscript $_{1:2}$ denotes row and column selection. 

Using the image-space splat center and 2D covariance matrix, we obtain the 2D image-space Gaussian distribution induced by the corresponding splat: 
\begin{equation} \label{eq:sigma2d}
    \mathcal{G}^{\text{2D}}_k(\mathbf{x}^{\prime}) \propto \exp \left( -\frac{1}{2}  (\mathbf{x}^{\prime}-\mathbf{p}^{\prime}_k)^T (\mathbf{\Sigma}_k^{\text{2D}})^{-1} (\mathbf{x}^{\prime}-\mathbf{p}_k^{\prime})  \right)\,, 
\end{equation}

Finally, we can predict the color $\mathbf{c}(\mathbf{x}^{\prime}) \in \mathbb{R}^3$ at each pixel location $\mathbf{x}^{\prime}\in \mathbb{R}^2$ by blending the splats, sorted according to their projection depth: 
\begin{equation} \label{eq:render}
    \mathbf{c}(\mathbf{x}^{\prime}) = \sum\nolimits_{k=1}^{K} \mathbf{c}_k \alpha_k \mathcal{G}^{\text{2D}}_k(\mathbf{x}^{\prime}) \prod\nolimits_{j=1}^{k-1}\Big(1-\alpha_j \mathcal{G}^{\text{2D}}_j(\mathbf{x}^{\prime})\Big)\,,
\end{equation}
where $\alpha_k$ is the learned opacity of the splat.

\textbf{\emph{Training.}} 
The collection of splats $\{\mathcal{G}_k\}_{k=1}^K$ is optimized by minimizing the following rendering loss w.r.t the input images via the Adam~\cite{Adam:ICLR:2015} optimizer:
\begin{equation} \label{eq:gaussian_loss}
    \mathcal{L} = (1-\lambda)\mathcal{L}_1 + \lambda \mathcal{L}_{\text{D-SSIM}},
\end{equation}
where the first term is a standard $\mathcal{L}_1$ loss between the target and rendered images, and  $\mathcal{L}_{\text{D-SSIM}}$ is a differentiable version of structural similarity index \cite{wang2004image}. As this optimization is highly sensitive to a local minima, 3DGS additionally employs periodic adaptive densification and pruning of splats through randomized sampling. We refer to \cite{GaussianSplatting} for further details. 

\section{SplatFields: Neural Gaussian Splats} \label{sec:splatFields} 
\textbf{Limitations of 3DGS.} 
Modeling 3D scenes with irregularly spaced point primitives offers significant flexibility and facilitates rapid and efficient optimization when an extensive number of training views is provided. 
However, with limited views, these independent point primitives are prone to overfitting. 
Therefore, we advocate for integrating a spatial autocorrelation bias within the splats. 
This can be accomplished by deriving splat features through implicit neural models, presenting a viable method to constrain and regularize the otherwise ill-posed optimization in sparse-view environments.

\textbf{Key insight.} 
We analyze the optimization procedure of 3DGS under sparse view inputs and observe (Fig.~\ref{fig:teaser}) that the splats do not exhibit any local structure and display incoherent patterns. 
To quantify the local spatial autocorrelation of each splat, we select the five nearest neighbors and measure Moran's I~\cite{moran1950notes} of the splat's attributes (color, opacity, covariance). 
We note that a low level of spatial autocorrelation is associated with overfitting to training views, which impedes the learning of a structured 3D representation. This correlation is further showcased in our discussion and empirical evidence presented in Tab.~\ref{tab:exp_morans}.

The core idea of our method is to introduce a \textit{spatial bias} during the optimization phase, which encourages nearby primitives to share similar features, thereby emulating the more continuous behavior characteristic of widely used implicit representations for volumetric rendering~\cite{mildenhall2020nerf,tewari2022advances,ResFields}. 
However, directly enforcing this constraint—ensuring local neighborhoods exhibit common patterns—yields sub-optimal performance (Tab~\ref{tab:exp_static_blender}) for volumetric point representations. 
To overcome this, we propose a novel neural framework, termed \textit{SplatFields}, designed to adaptively regularize the optimization of 3DGS. Importantly, SplatFields straightforwardly extends to 4D, facilitating the reconstruction of dynamic scenes.

\begin{figure}[t]
    \begin{center}
        \includegraphics[width=\textwidth]{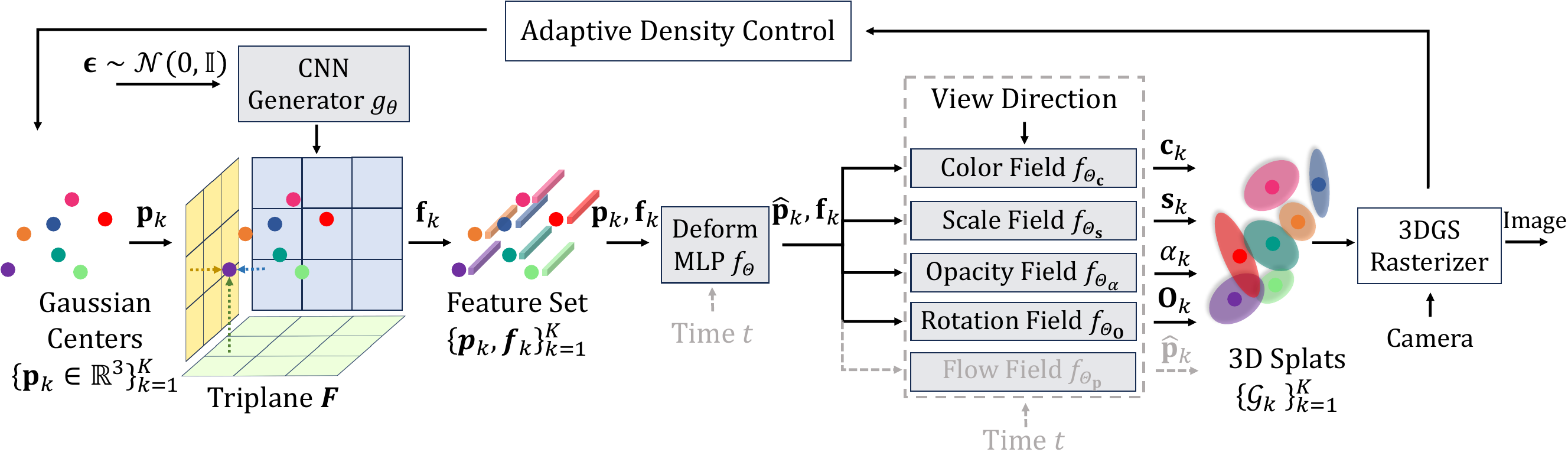}
    \end{center}
    \caption{\textbf{Overview.} SplatFields takes as input a point cloud (\textit{e.g.}, initialized from SfM~\cite{schoenberger2016sfm}), for which it models the geometric (position $\mathbf{p}_k$, scale $\mathbf{s}_k$, rotation $\mathbf{O}_k$) and appearance attributes (color $\mathbf{c}_k$, opacity $\alpha_k$). These attributes represent the point set as 3D splats that are then rendered with the 3DGS rasterizer~\cite{GaussianSplatting}. First, the point location set $\{\mathbf{p}_k\in \mathbb{R}^3\}_{k=1}^K$ is encoded into features $\{\mathbf{f}_k\}_{k=1}^{K}$ by sampling the tri-plane representation generated by a CNN generator $g_\theta$ to provide a deep structural prior~\cite{ulyanov2018deep} on the feature values. These values are then propagated through a deformation MLP $f_\Theta$ to refine the point locations $\hat{\mathbf{p}}_k$. The new point set, along with the features, is then propagated through a series of compact neural fields to predict the properties of rendering primitives $\{\mathcal{G}_k\}_{k=1}^K$ that are rendered with respect to arbitrary viewpoints. During the optimization, we adopt the adaptive density control~\cite{GaussianSplatting} to periodically prune and densify the point set. SplatFields seamlessly adapts to 4D reconstruction by conditioning neural fields on the time step $t$ and introducing an extra time-conditioned flow field. Gray blocks indicate learnable modules.
    }
    \label{fig:overview}
\end{figure}

SplatFields (Fig.~\ref{fig:overview}) builds on the core property of neural networks to discover local patterns and fit low frequencies of a signal first \cite{rahaman2019spectral,ulyanov2018deep}. 
To that end, we implement SplatFields as a neural generator that infers the attributes of Gaussian splats.
The neural generator combines key properties of convolutional neural networks, which model local structured patterns, with multi-layer perceptrons that serve as global approximators. This approach straightforwardly extends to 4D reconstruction by conditioning the MLP networks on time $\textcolor{gray}{t}$. 

\textbf{Deep structural prior.} First, we follow the idea of a deep image prior~\cite{ulyanov2018deep,zerorf} and aim to utilize CNNs to model locally structured patterns of splat features. In the original work~\cite{ulyanov2018deep}, the CNN takes as input low-dimension Gaussian noise $\mathbf{\epsilon} \sim \mathcal{N}(0,\mathbb{I})$ and gradually upsamples it into the desired image resolution; the weights of the network are then optimized to fit the observed noisy image. In our case, we aim to generate a 3D field of splat features; as 3D CNNs are computationally prohibitive, we use 2D CNNs that generate axis-aligned tri-plane representations \cite{peng2020convolutional,chan2022efficient}. Overall, the step is a splat-based variation of the approach utilized in~\cite{zerorf} for a fully volumetric NeRF-based sparse rendering.
\begin{table}[t]
  \caption{
    \textbf{Impact of the spatial autocorrelation} on static scene reconstruction. Results on Owlii~\cite{xu2017owlii} dataset. See Section~\ref{subsec:exp_static} for discussion
    }
  \label{tab:exp_morans}
  \centering
  \setlength{\tabcolsep}{3.0pt} %

\scriptsize
\resizebox{\textwidth}{!}{%
\begin{tabular}{@{}l|ccc|ccc|ccc@{}}

\toprule

& \multicolumn{3}{c}{Train} & \multicolumn{3}{c}{Test}   & \multicolumn{3}{c}{Spatial Autocorrelation}\\
& \multicolumn{3}{c}{View Synthesis} & \multicolumn{3}{c}{Novel View Synthesis}   & \multicolumn{3}{c}{(Moran’s I) $\uparrow$}\\
& SSIM$\uparrow$                & PSNR$\uparrow$ & LPIPS$\downarrow$   & SSIM$\uparrow$ & PSNR$\uparrow$ & LPIPS$\downarrow$   &   Color        &      Opacity & Covariance \\ \midrule

3DGS~\cite{GaussianSplatting}   & 99.85 &  44.07 & 0.493 & 91.68            &  27.50          & 8.881          & 0.547          & 0.670         & 0.232 \\
SplatFields3D                   & 98.87 &  37.58 & 3.274 & \textbf{96.13}   &  \textbf{30.33} & \textbf{5.973} & \textbf{0.935} & \textbf{0.874}& \textbf{0.431} \\

\bottomrule
\end{tabular}
}
\end{table}

Specifically, given a randomly initialized noise $\mathbf{\epsilon}$, the convolutional network $g_{\theta}$ regresses the three $H \times W$-resolution planes $\mathbf{F}$: 
\begin{equation} \label{eq:conv}
    \mathbf{F} = g_{\theta}(\mathbf{\epsilon}) \in \mathbb{R}^{3 \times H \times W \times l}\,,
\end{equation}
where $l$ denotes the feature dimension and $\theta$ indicates learnable network weights. The overall CNN structure resembles the one originally proposed in~\cite{zerorf}. 

\textbf{Neural splat fields.} Next, the splat center $\mathbf{p}_k$ is projected onto each of the three feature planes to obtain feature vectors via bilinear interpolation. These features are then concatenated along the feature dimension and denoted as $\mathbf{f}_k \in \mathbb{R}^{3l}$. 
The feature and the initial point are propagated through a deformation MLP $f_{\Theta}$ which refines the position of the input point: 
\begin{equation}
     \hat{\mathbf{p}}_k = f_{\Theta}(\mathbf{p}_k, \mathbf{f}_k, \textcolor{gray}{t}),
\end{equation}
where \textcolor{gray}{$t$} indicates an optional time step input provided in the case of dynamic 4D reconstruction. Finally, the updated point location, along with the inferred feature vector, is provided as input to a set of compact (5-6 layers, 64-128 neurons) neural fields $\{f_{\Theta_{\mathbf{c}}}$, $f_{\Theta_{\mathbf{s}}}$, $f_{\Theta_{\alpha}}$, $f_{\Theta_{\mathbf{O}}}\}$ to obtain properties of Gaussian splats. 
Rather than using spherical harmonics to model color, we directly predict view-dependent color. 
The obtained splats are then rendered w.r.t. the input views to optimize the learnable modules by minimizing the photometric loss (Eq.~\ref{eq:gaussian_loss}). 

\textbf{Splat norm regularization.} For static reconstruction, we add additional norm regularization $||\hat{\mathbf{p}}_k||_2$ to the loss function to bias the resulting splats to not deviate significantly from the origin, similar to the floor loss considered in \cite{luiten2023dynamic}. 

\textbf{4D reconstruction.} SplatFields is a flexible representation that straightforwardly extends to 4D reconstruction of dynamic scenes. 
It models temporal variations in the splat features by conditioning the corresponding neural field MLPs $\{f_{\Theta}, f_{\Theta_{\mathbf{c}}}$, $f_{\Theta_{\mathbf{s}}}$, $f_{\Theta_{\alpha}}$, $f_{\Theta_{\mathbf{O}}}\}$ on the time step \textcolor{gray}{$t$}. 
Additionally, we add a time-conditioned flow field \textcolor{gray}{$f_\mathbf{p}$} to warp the center of Gaussians $\mathbf{p}_k$ to the desired time step \textcolor{gray}{$t$}.
To enhance the expressivity of the neural fields and allow for complex geometry changes, we utilize the recently proposed ResField MLP architecture~\cite{ResFields}. 

\section{Experiments} 
\subsection{Static Scene Reconstruction} \label{subsec:exp_static} 
\textbf{Impact of the spatial autocorrelation.} 
First, we conduct a toy experiment to verify our intuition that the absence of the \textit{spatial bias} hampers the reconstruction quality from sparse views. 
We utilize four sequences from the Owlii dynamic dataset~\cite{xu2017owlii}\footnote{All experiments presented in this publication were performed by ETH Zürich. ETH Zürich obtained the licenses for the data used in such experiments.} and select the first frame from each. 
Each scene comprises nine training views and one validation view, on which we report training and test metrics. 
We compare 3DGS~\cite{GaussianSplatting} and SplatFields (both initialized from visual hulls) and observe (Tab.~\ref{tab:exp_morans}) that 3DGS demonstrates extremely high fitting quality on the training views while poorly generalizing to the novel views. 
In contrast, SplatFields demonstrates a slightly lower training quality while achieving higher reconstruction quality on novel views. 
This observation is followed by computing the Moran’s I metric~\cite{moran1950notes} which shows the amount of spatial correlation between nearby splat features; as hypothesized, the lower test-time error is in line with the increased level of spatial consistency of all groups of splat features. Fig.~\ref{fig:teaser} presents both qualitative and quantitative results from the same experiment conducted on a scene from the Blender dataset~\cite{mildenhall2020nerf}.

\begin{figure} %
  \centering
  \begin{minipage}{1.0\textwidth}
    \scriptsize
    \setlength{\tabcolsep}{0.2mm} %
    \newcommand{\sz}{0.160}  %
    \renewcommand{\arraystretch}{0.0} 
    \caption{\textbf{Static reconstruction} of Blender~\cite{mildenhall2020nerf} scenes for the setup from Tab.~\ref{tab:exp_static_blender}}
\resizebox{\textwidth}{!}{%
    \begin{tabular}{ccccccc}  %
           & \text{SuGaR}~\cite{sugar} & \text{Mip3DGS}~\cite{mipgs} & \text{3DGS}~\cite{GaussianSplatting} & \text{Light3DGS}~\cite{lightgaussian} & \text{SplatFields3D} & GT \\ 

           \rotatebox{90}{\phantom{+}12 Views}
& 
\begin{tikzpicture} 
\node[anchor=south west,inner sep=0] (image) at (0,0) {
\includegraphics[width=\sz\linewidth, trim={50 50 100 220},clip]{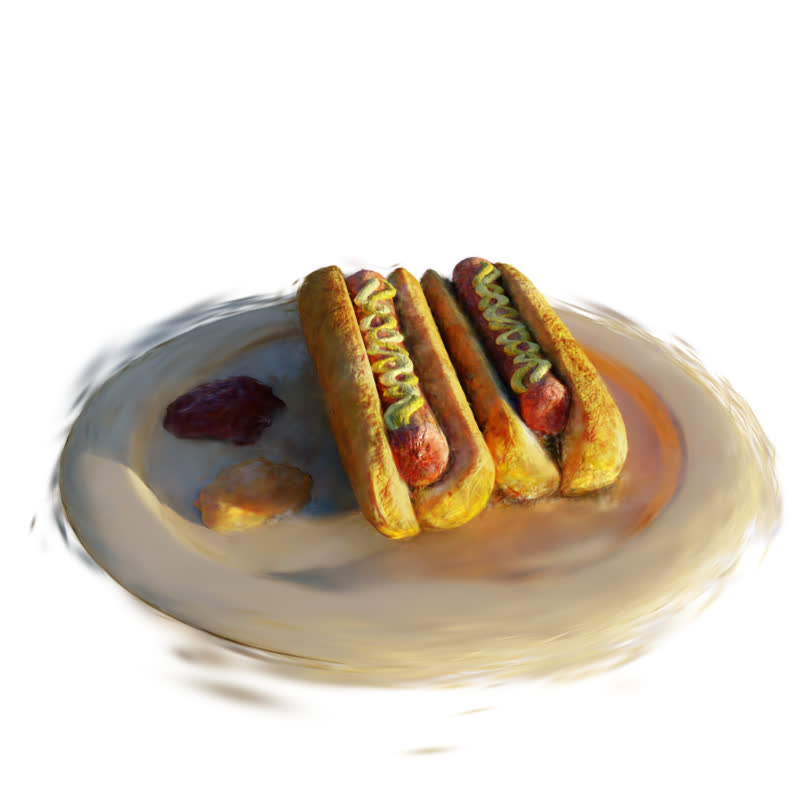} 
}; 
\begin{scope}[x={(image.south east)},y={(image.north west)}]
\draw[red,thick] (0.25,0.20) rectangle (0.80,0.70); %
\end{scope} 
\end{tikzpicture} 
& 
\begin{tikzpicture} 
\node[anchor=south west,inner sep=0] (image) at (0,0) {
\includegraphics[width=\sz\linewidth, trim={50 50 100 220},clip]{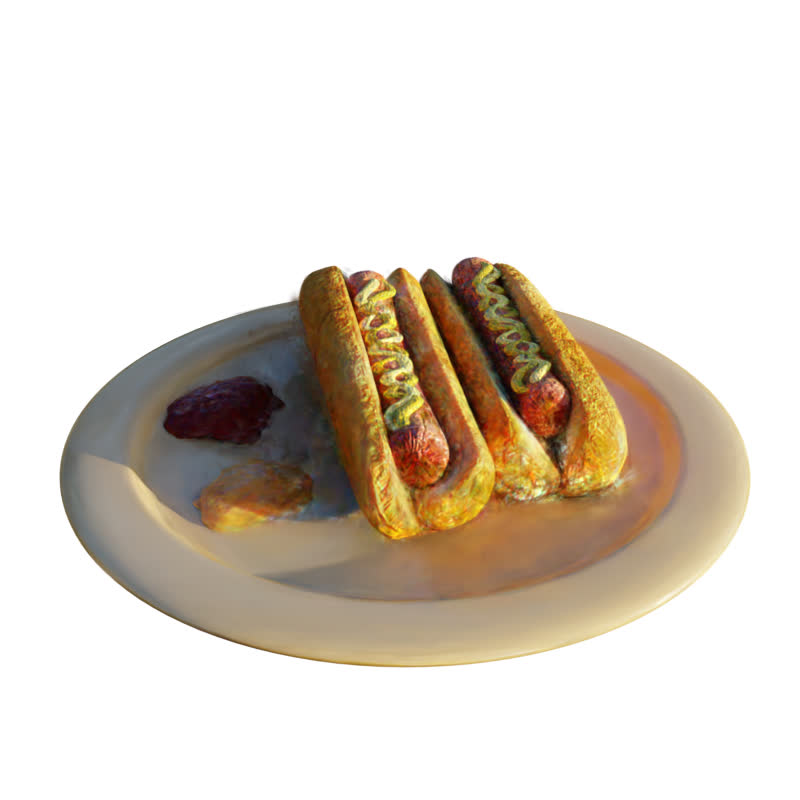} 
}; 
\begin{scope}[x={(image.south east)},y={(image.north west)}]
\draw[red,thick] (0.25,0.20) rectangle (0.80,0.70); %
\end{scope} 
\end{tikzpicture} 
& 
\begin{tikzpicture} 
\node[anchor=south west,inner sep=0] (image) at (0,0) {
\includegraphics[width=\sz\linewidth, trim={50 50 100 220},clip]{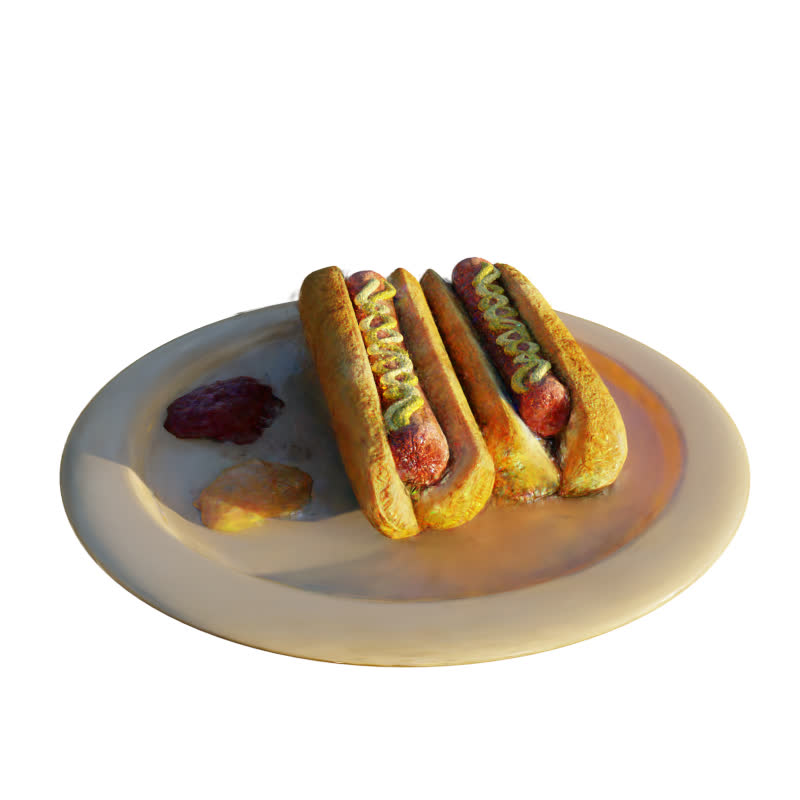} 
}; 
\begin{scope}[x={(image.south east)},y={(image.north west)}]
\draw[red,thick] (0.25,0.20) rectangle (0.80,0.70); %
\end{scope} 
\end{tikzpicture} 
& 
\begin{tikzpicture} 
\node[anchor=south west,inner sep=0] (image) at (0,0) {
\includegraphics[width=\sz\linewidth, trim={50 50 100 220},clip]{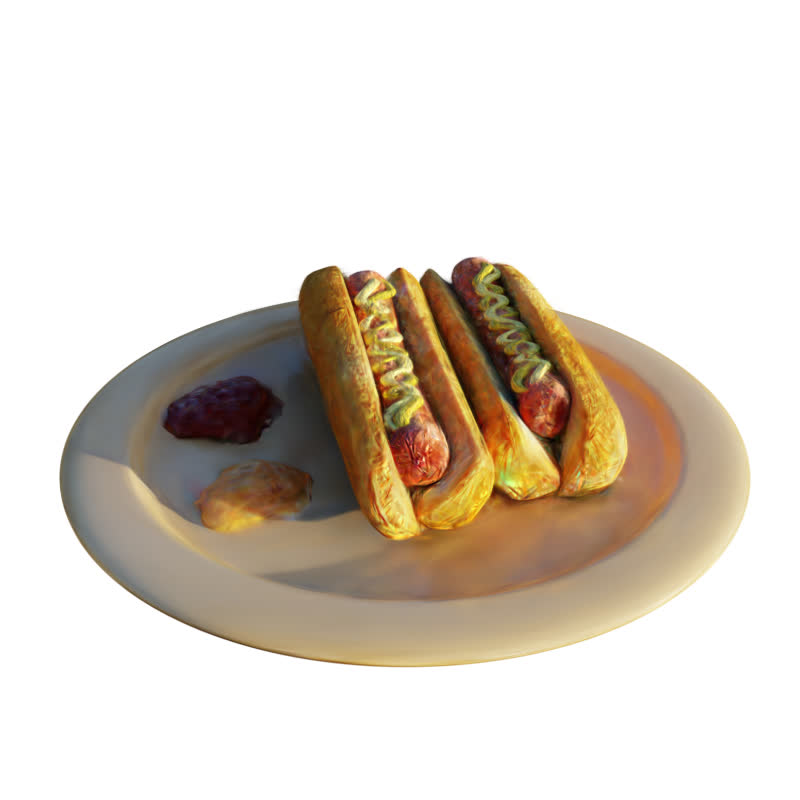} 
}; 
\begin{scope}[x={(image.south east)},y={(image.north west)}]
\draw[red,thick] (0.25,0.20) rectangle (0.80,0.70); %
\end{scope} 
\end{tikzpicture} 
& 
\begin{tikzpicture} 
\node[anchor=south west,inner sep=0] (image) at (0,0) {
\includegraphics[width=\sz\linewidth, trim={50 50 100 220},clip]{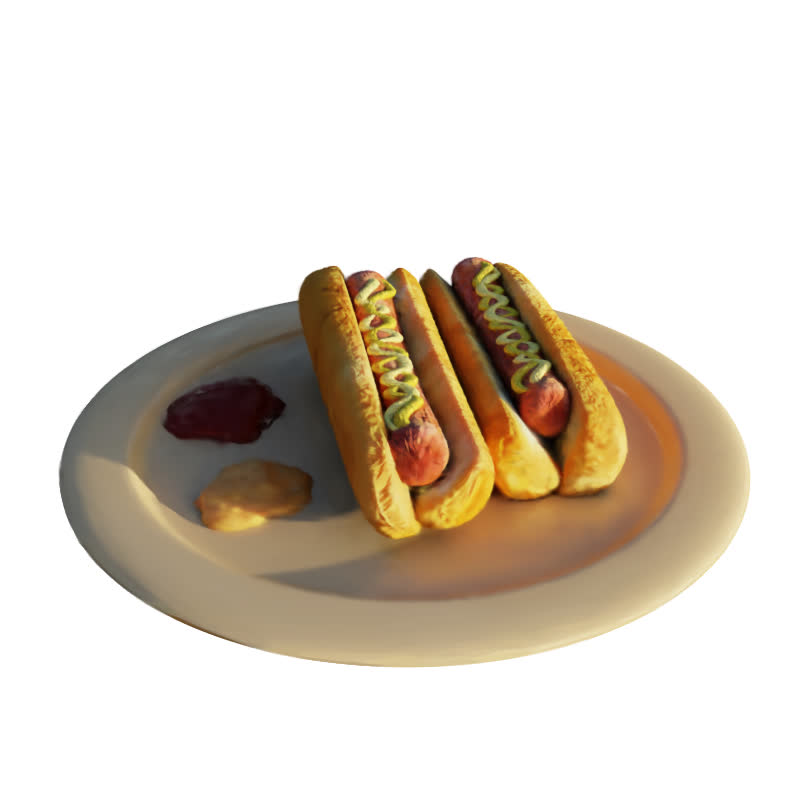} 
}; 
\begin{scope}[x={(image.south east)},y={(image.north west)}]
\draw[green,thick] (0.25,0.20) rectangle (0.80,0.70); %
\end{scope} 
\end{tikzpicture} 
&
\begin{tikzpicture} 
\node[anchor=south west,inner sep=0] (image) at (0,0) {
\includegraphics[width=\sz\linewidth, trim={50 50 100 220},clip]{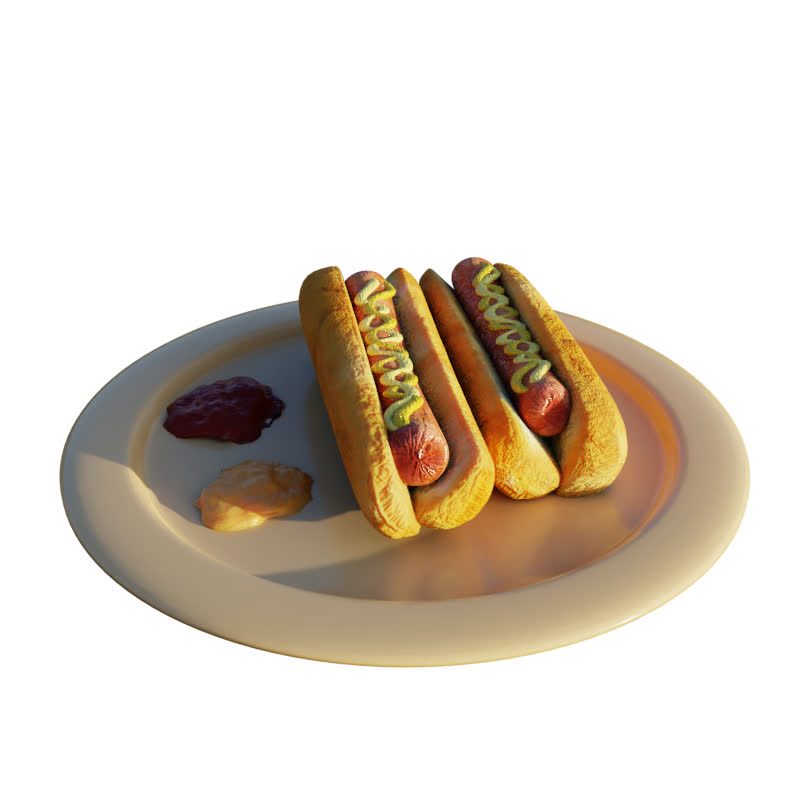}
}; 
\begin{scope}[x={(image.south east)},y={(image.north west)}]
\draw[black,thick] (0.25,0.20) rectangle (0.80,0.70); %
\end{scope} 
\end{tikzpicture} 
\\
            \rotatebox{90}{\phantom{+}6 Views}
& 
\begin{tikzpicture} 
\node[anchor=south west,inner sep=0] (image) at (0,0) {
\includegraphics[width=\sz\linewidth, trim={50 50 100 220},clip]{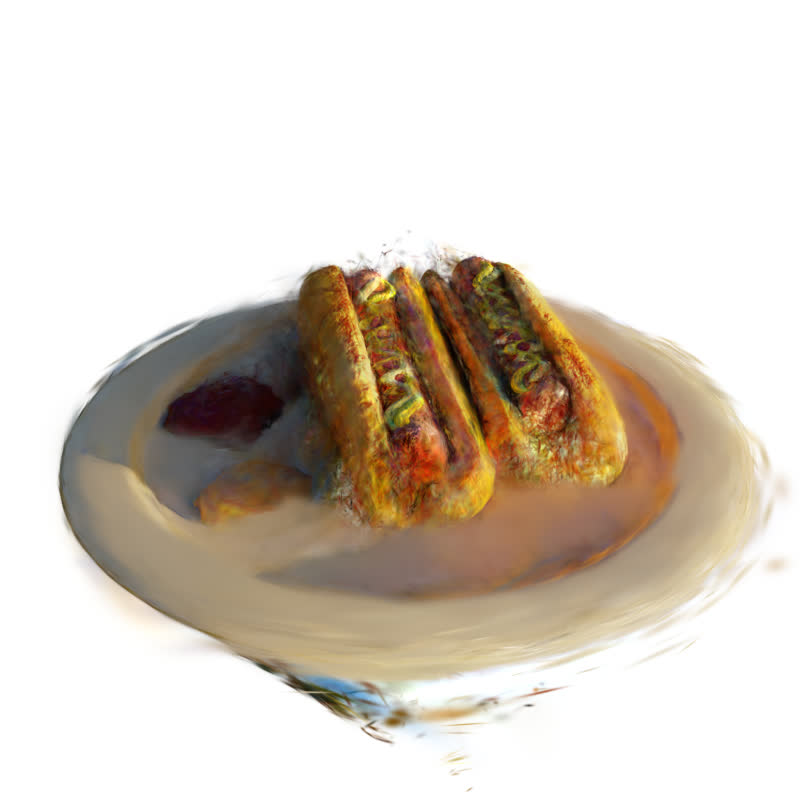}
}; 
\begin{scope}[x={(image.south east)},y={(image.north west)}]
\draw[red,thick] (0.15,0.01) rectangle (0.92,0.40); %
\end{scope} 
\end{tikzpicture} 
& 
\begin{tikzpicture} 
\node[anchor=south west,inner sep=0] (image) at (0,0) {
\includegraphics[width=\sz\linewidth, trim={50 50 100 220},clip]{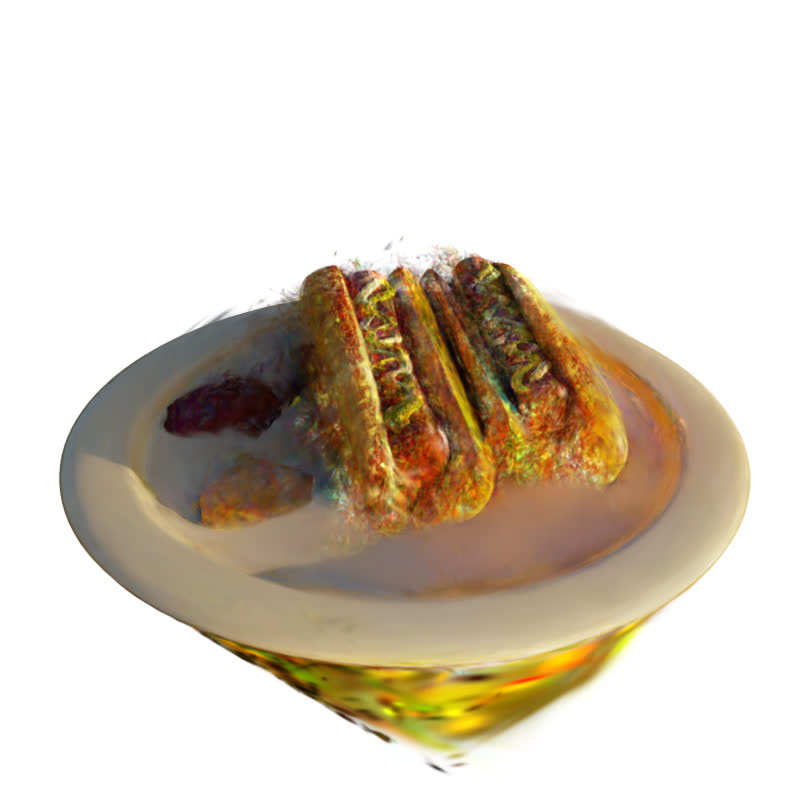}
}; 
\begin{scope}[x={(image.south east)},y={(image.north west)}]
\draw[red,thick] (0.15,0.01) rectangle (0.92,0.40); %
\end{scope} 
\end{tikzpicture} 
& 
\begin{tikzpicture} 
\node[anchor=south west,inner sep=0] (image) at (0,0) {
\includegraphics[width=\sz\linewidth, trim={50 50 100 220},clip]{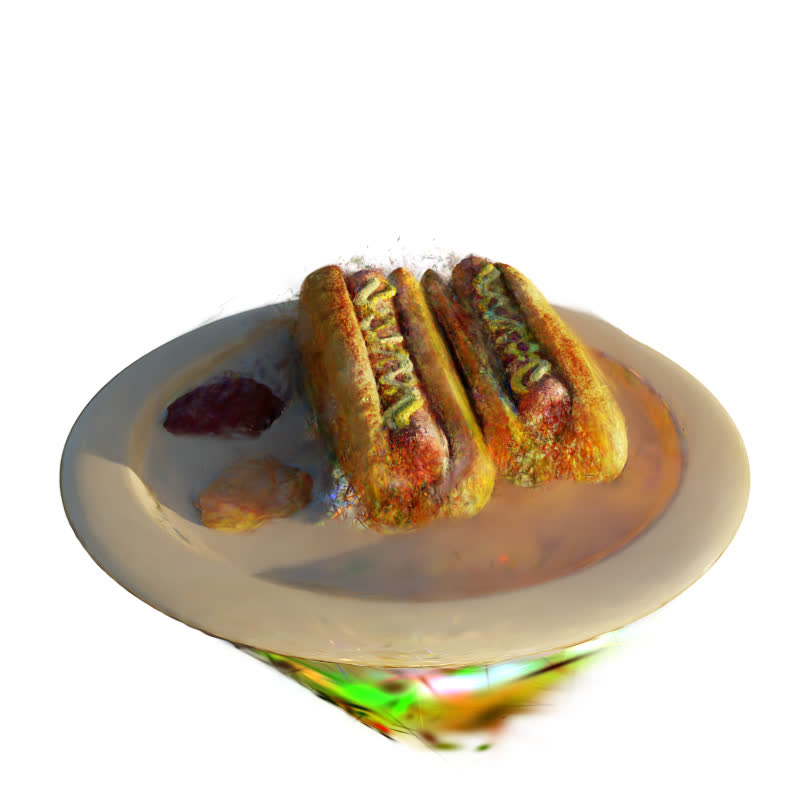}
}; 
\begin{scope}[x={(image.south east)},y={(image.north west)}]
\draw[red,thick] (0.15,0.01) rectangle (0.92,0.40); %
\end{scope} 
\end{tikzpicture} 
& 
\begin{tikzpicture} 
\node[anchor=south west,inner sep=0] (image) at (0,0) {
\includegraphics[width=\sz\linewidth, trim={50 50 100 220},clip]{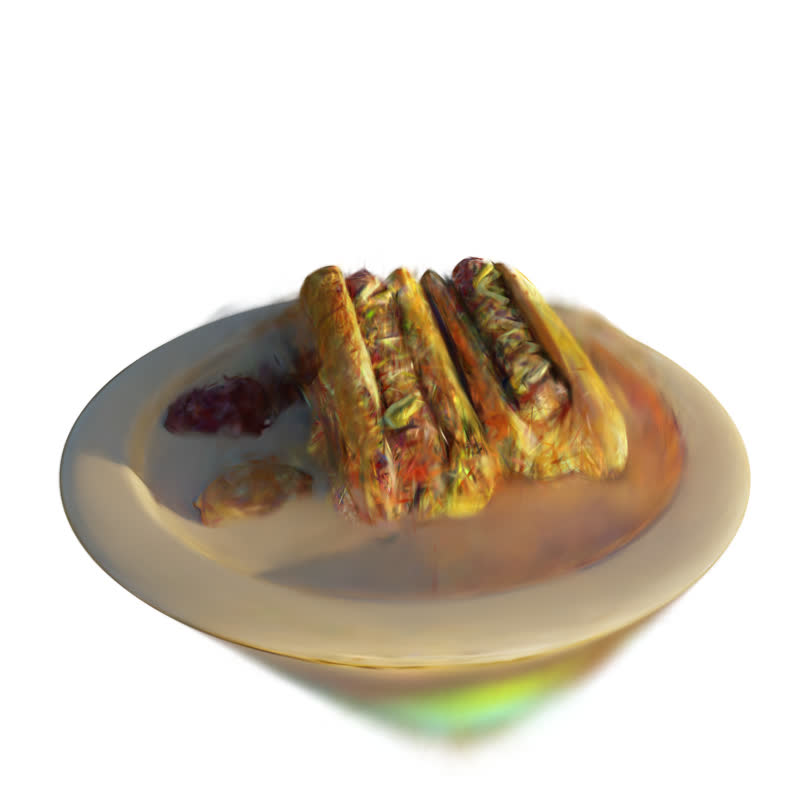}
}; 
\begin{scope}[x={(image.south east)},y={(image.north west)}]
\draw[red,thick] (0.15,0.01) rectangle (0.92,0.40); %
\end{scope} 
\end{tikzpicture} 
& 
\begin{tikzpicture} 
\node[anchor=south west,inner sep=0] (image) at (0,0) {
\includegraphics[width=\sz\linewidth, trim={50 50 100 220},clip]{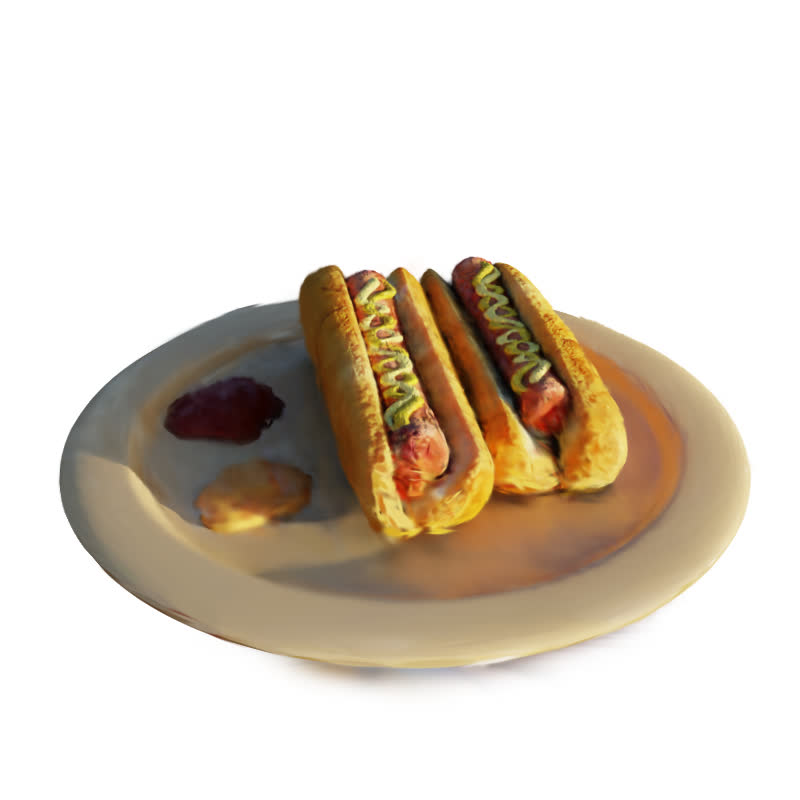}
}; 
\begin{scope}[x={(image.south east)},y={(image.north west)}]
\draw[green,thick] (0.15,0.01) rectangle (0.92,0.40); %
\end{scope} 
\end{tikzpicture} 
& 
\begin{tikzpicture} 
\node[anchor=south west,inner sep=0] (image) at (0,0) {
\includegraphics[width=\sz\linewidth, trim={50 50 100 220},clip]{results/Blender/hotdog_6views_gt/00047.jpg}
}; 
\begin{scope}[x={(image.south east)},y={(image.north west)}]
\draw[black,thick] (0.15,0.01) rectangle (0.92,0.40); %
\end{scope} 
\end{tikzpicture} 
\\

          \rotatebox{90}{\phantom{+++}12 Views}
&
\begin{tikzpicture} 
\node[anchor=south west,inner sep=0] (image) at (0,0) {
\includegraphics[width=\sz\linewidth, trim={130 10 130 140},clip]{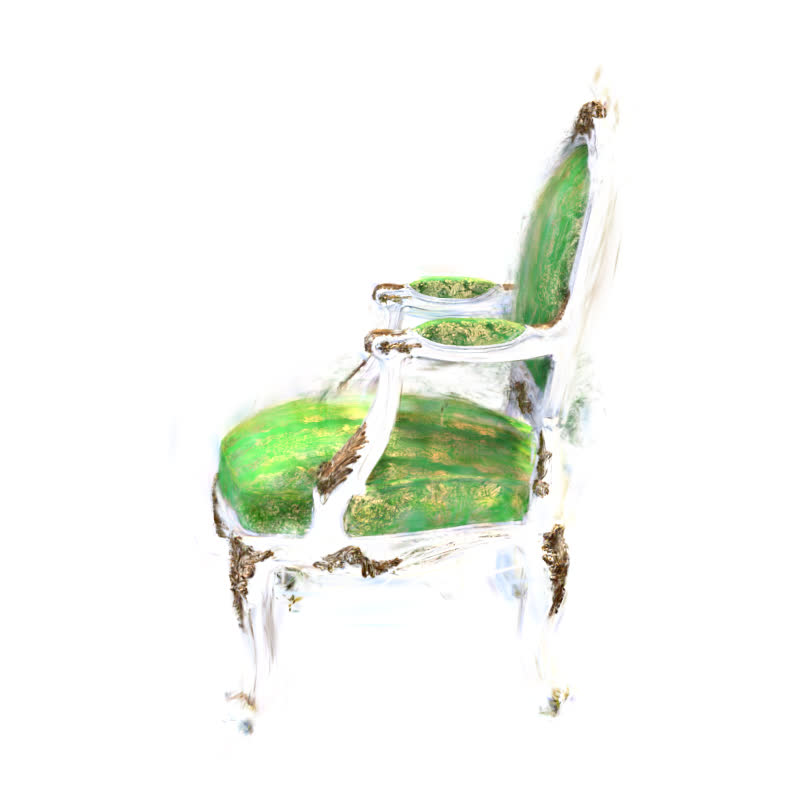}
}; 
\begin{scope}[x={(image.south east)},y={(image.north west)}]
\draw[red,thick] (0.62,0.45) rectangle (0.93,0.75); %
\end{scope} 
\end{tikzpicture} 
&
\begin{tikzpicture} 
\node[anchor=south west,inner sep=0] (image) at (0,0) {
\includegraphics[width=\sz\linewidth, trim={130 10 130 140},clip]{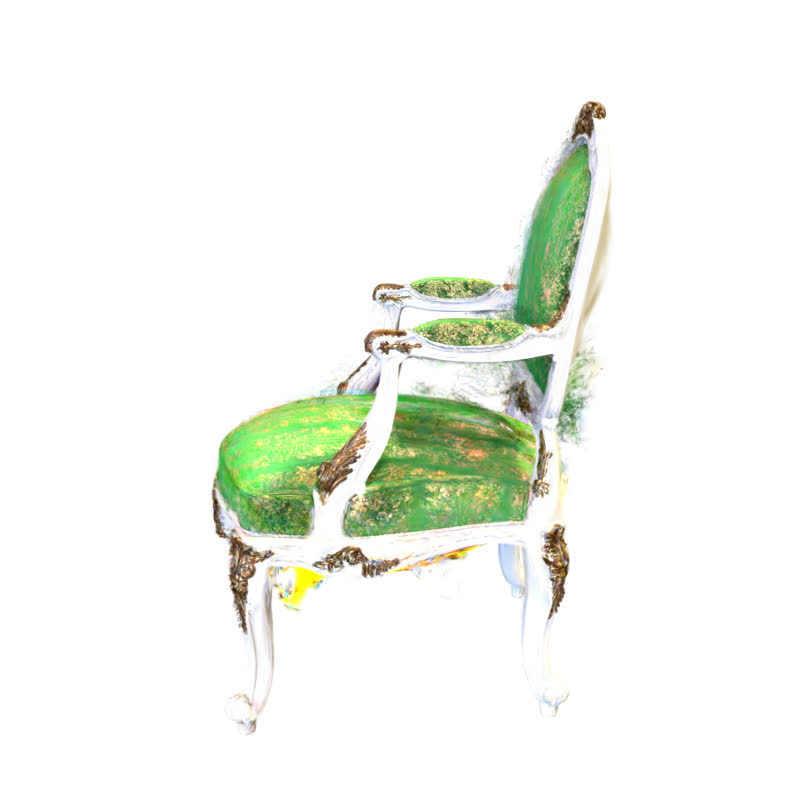}
}; 
\begin{scope}[x={(image.south east)},y={(image.north west)}]
\draw[red,thick] (0.62,0.45) rectangle (0.93,0.75); %
\end{scope} 
\end{tikzpicture} 
&
\begin{tikzpicture} 
\node[anchor=south west,inner sep=0] (image) at (0,0) {
\includegraphics[width=\sz\linewidth, trim={130 10 130 140},clip]{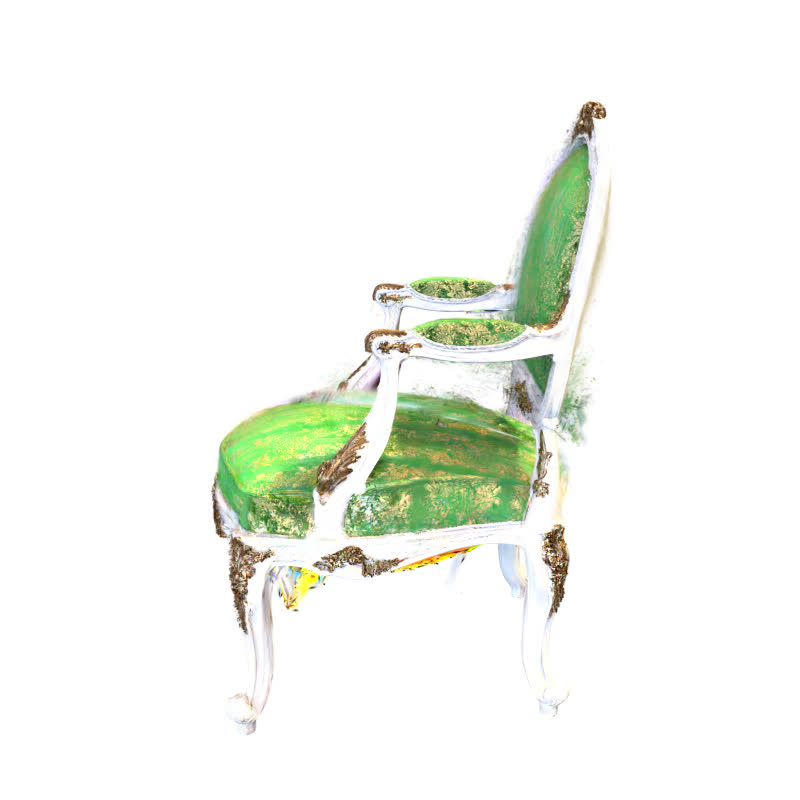}
}; 
\begin{scope}[x={(image.south east)},y={(image.north west)}]
\draw[red,thick] (0.62,0.45) rectangle (0.93,0.75); %
\end{scope} 
\end{tikzpicture} 
&
\begin{tikzpicture} 
\node[anchor=south west,inner sep=0] (image) at (0,0) {
\includegraphics[width=\sz\linewidth, trim={130 10 130 140},clip]{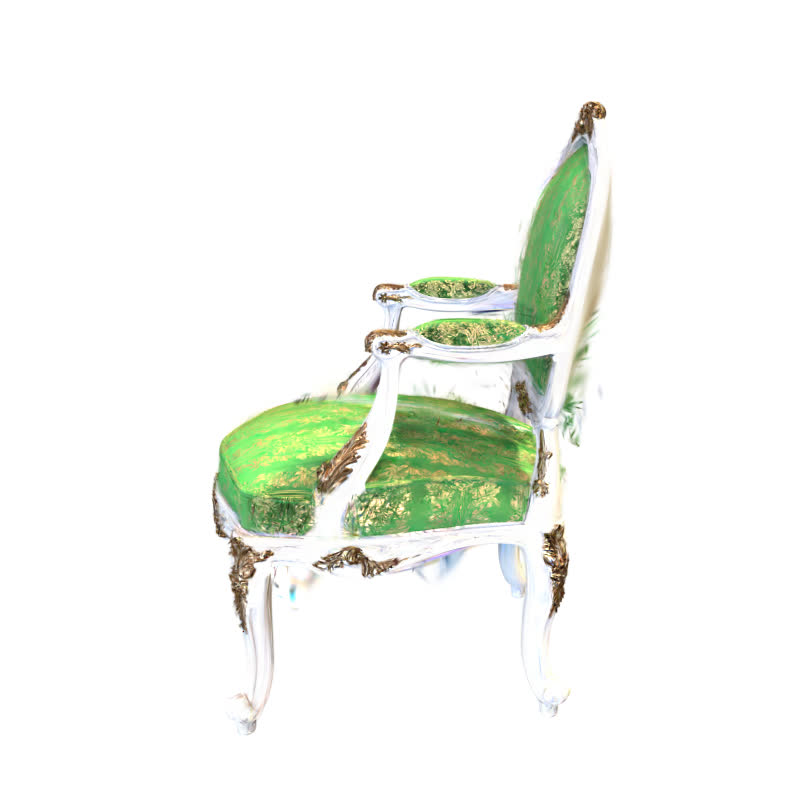}
}; 
\begin{scope}[x={(image.south east)},y={(image.north west)}]
\draw[red,thick] (0.62,0.45) rectangle (0.93,0.75); %
\end{scope} 
\end{tikzpicture} 

&
\begin{tikzpicture} 
\node[anchor=south west,inner sep=0] (image) at (0,0) {
\includegraphics[width=\sz\linewidth, trim={130 10 130 140},clip]{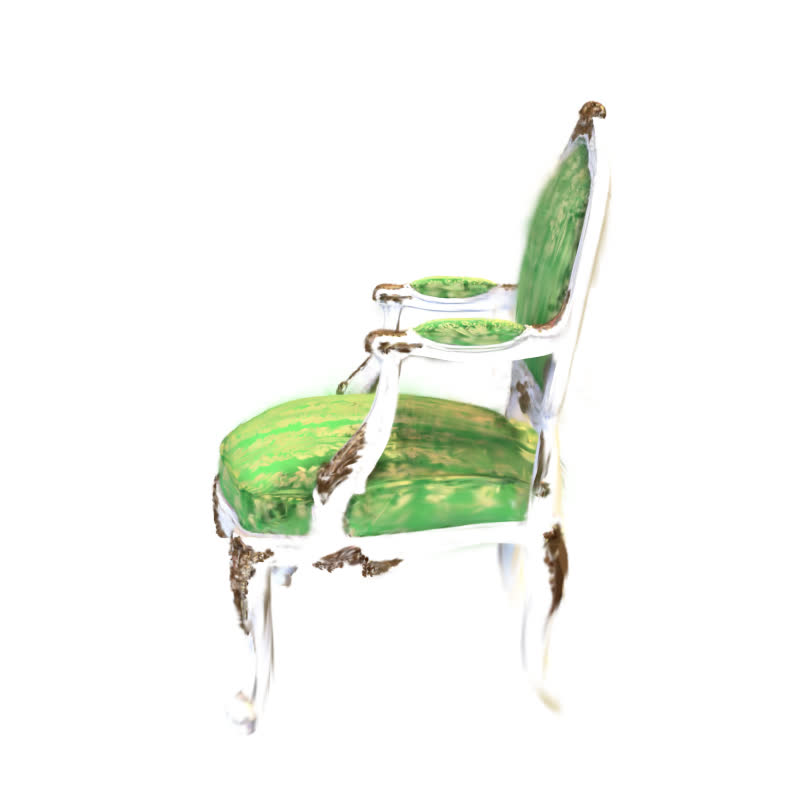}
}; 
\begin{scope}[x={(image.south east)},y={(image.north west)}]
\draw[green,thick] (0.62,0.45) rectangle (0.93,0.75); %
\end{scope} 
\end{tikzpicture} 
&
\begin{tikzpicture} 
\node[anchor=south west,inner sep=0] (image) at (0,0) {
\includegraphics[width=\sz\linewidth, trim={130 10 130 140},clip]{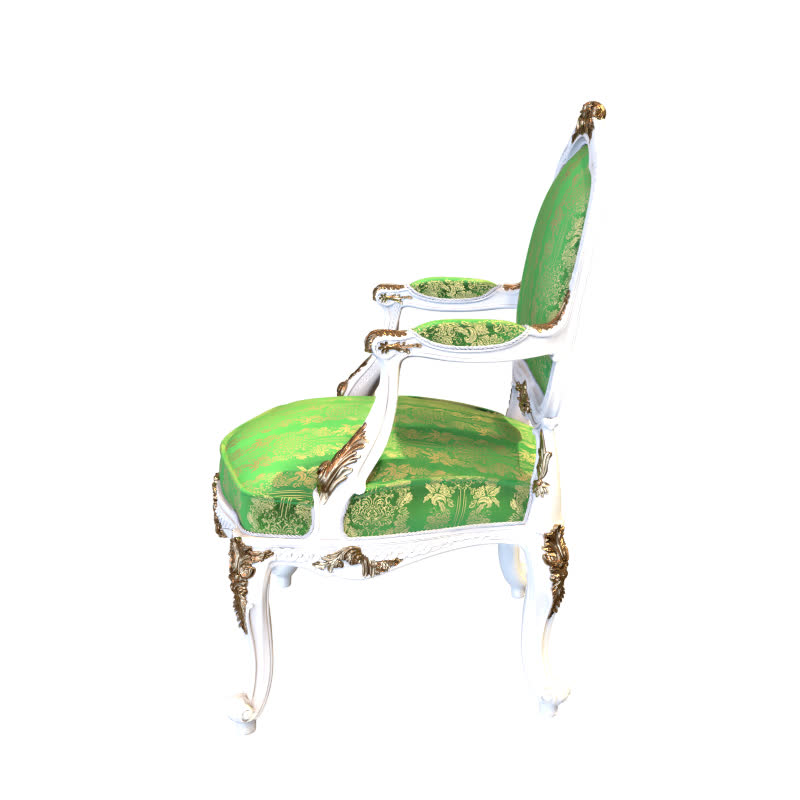}
}; 
\begin{scope}[x={(image.south east)},y={(image.north west)}]
\draw[black,thick] (0.62,0.45) rectangle (0.93,0.75); %
\end{scope} 
\end{tikzpicture} 
\\

            \rotatebox{90}{\phantom{++++}6 Views}
&
\begin{tikzpicture} 
\node[anchor=south west,inner sep=0] (image) at (0,0) {
\includegraphics[width=\sz\linewidth, trim={130 10 130 140},clip]{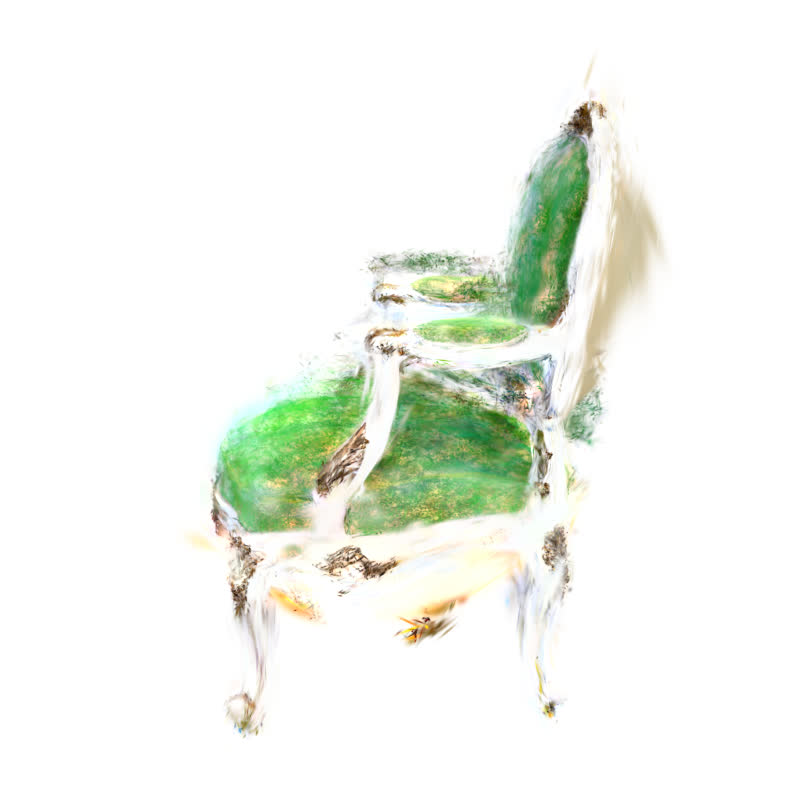}
}; 
\begin{scope}[x={(image.south east)},y={(image.north west)}]
\draw[red,thick] (0.3,0.2) rectangle (0.8,0.5); %
\end{scope} 
\end{tikzpicture} 
&
\begin{tikzpicture} 
\node[anchor=south west,inner sep=0] (image) at (0,0) {
\includegraphics[width=\sz\linewidth, trim={130 10 130 140},clip]{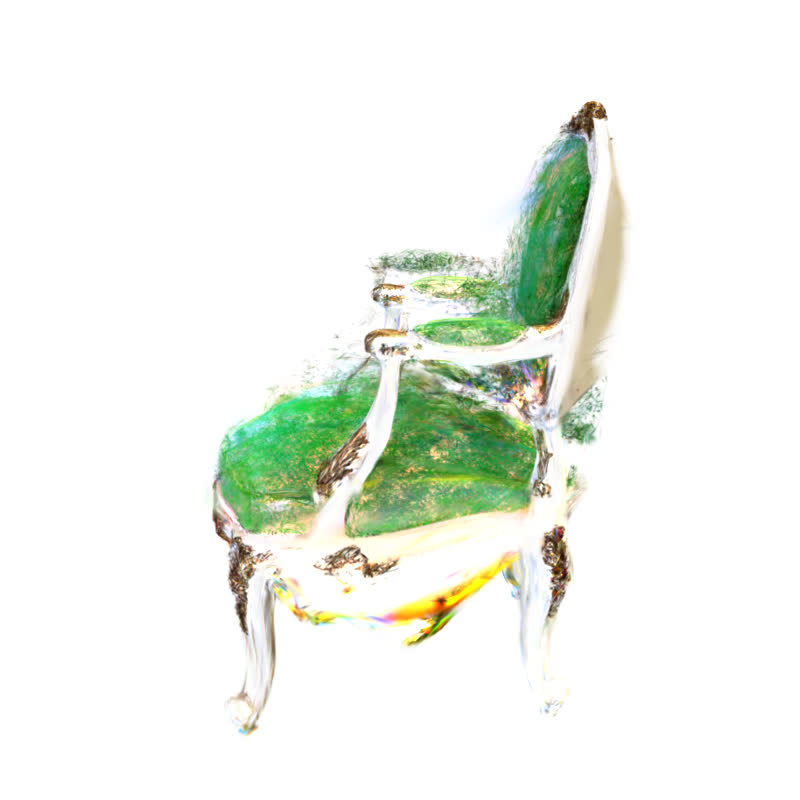}
}; 
\begin{scope}[x={(image.south east)},y={(image.north west)}]
\draw[red,thick] (0.3,0.2) rectangle (0.8,0.5); %
\end{scope} 
\end{tikzpicture} &
\begin{tikzpicture} 
\node[anchor=south west,inner sep=0] (image) at (0,0) {
\includegraphics[width=\sz\linewidth, trim={130 10 130 140},clip]{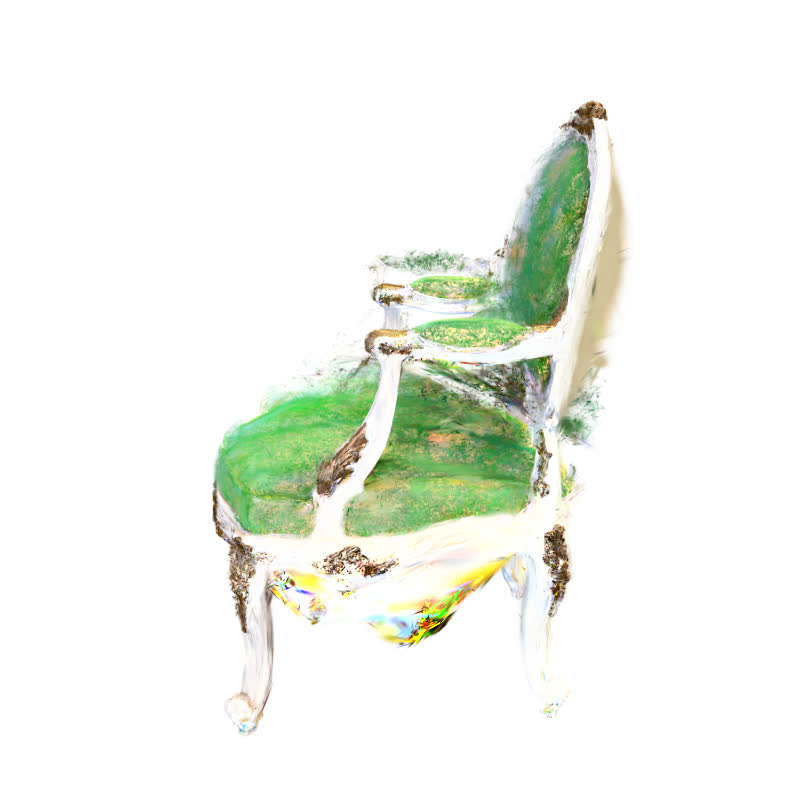}
}; 
\begin{scope}[x={(image.south east)},y={(image.north west)}]
\draw[red,thick] (0.3,0.2) rectangle (0.8,0.5); %
\end{scope} 
\end{tikzpicture} &
\begin{tikzpicture} 
\node[anchor=south west,inner sep=0] (image) at (0,0) {
\includegraphics[width=\sz\linewidth, trim={130 10 130 140},clip]{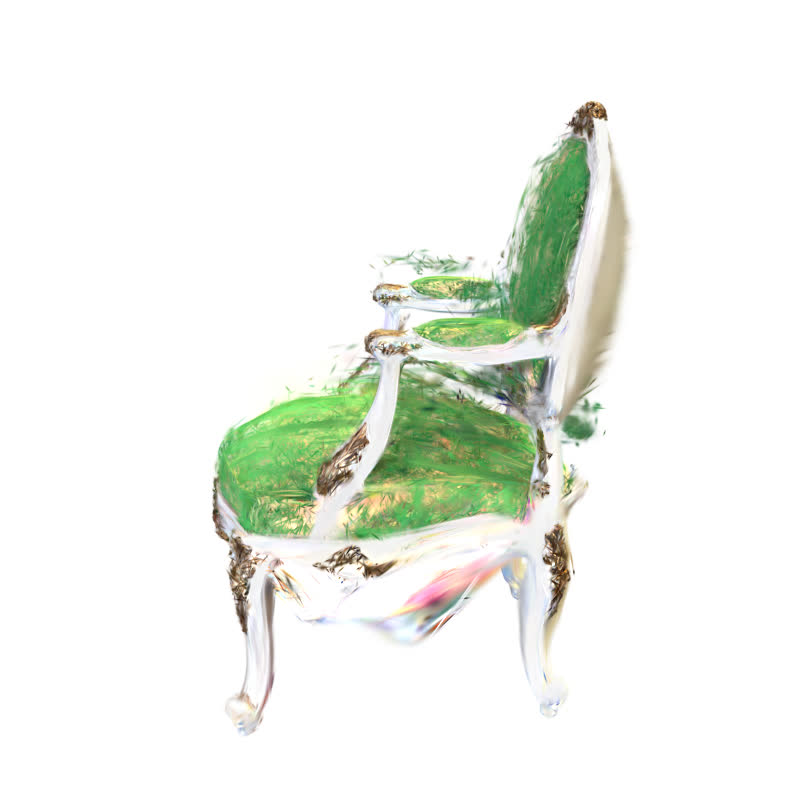}
}; 
\begin{scope}[x={(image.south east)},y={(image.north west)}]
\draw[red,thick] (0.3,0.2) rectangle (0.8,0.5); %
\end{scope} 
\end{tikzpicture} 

&
\begin{tikzpicture} 
\node[anchor=south west,inner sep=0] (image) at (0,0) {
\includegraphics[width=\sz\linewidth, trim={130 10 130 140},clip]{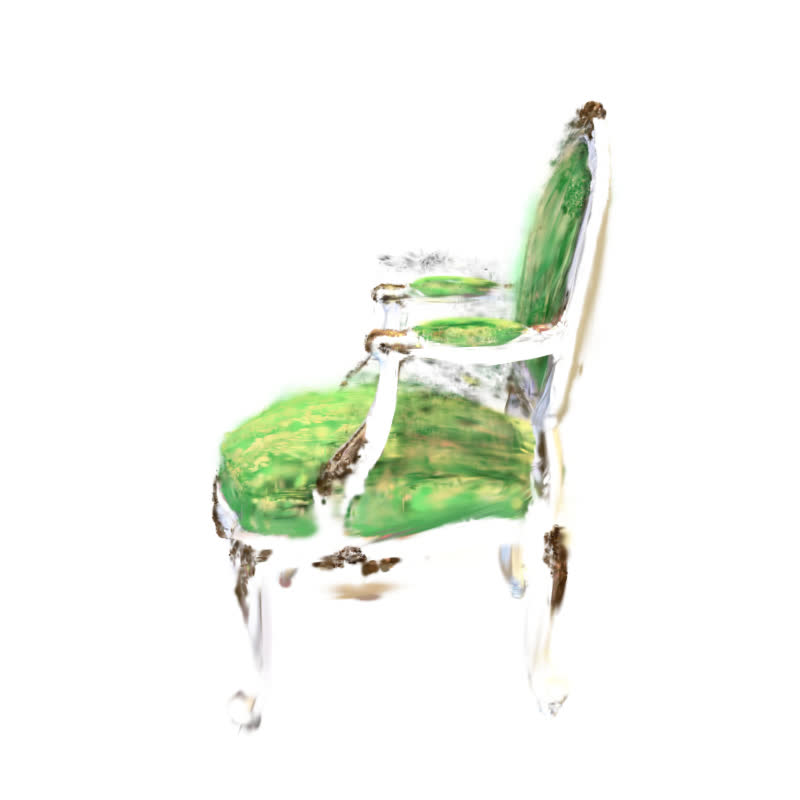}
}; 
\begin{scope}[x={(image.south east)},y={(image.north west)}]
\draw[green,thick] (0.3,0.2) rectangle (0.8,0.5); %
\end{scope} 
\end{tikzpicture} 
&
\begin{tikzpicture} 
\node[anchor=south west,inner sep=0] (image) at (0,0) {
\includegraphics[width=\sz\linewidth, trim={130 10 130 140},clip]{results/Blender/chair_6views_gt/00075.jpg}
}; 
\begin{scope}[x={(image.south east)},y={(image.north west)}]
\draw[black,thick] (0.3,0.2) rectangle (0.8,0.5); %
\end{scope} 
\end{tikzpicture} 
\\

    \label{fig:exp_static}

\end{tabular}
}
  \end{minipage}\hfill
  
  \vspace{1em}
  
  \begin{minipage}{1.0\textwidth}
  \centering
  \scriptsize
  \setlength{\tabcolsep}{3.9pt} %

\resizebox{\textwidth}{!}{%
\begin{tabular}{@{}l|ccccccccc@{}}
\toprule
&\multicolumn{9}{c}{12 Input Views} \\
 & \textit{mean} & Toy  &       Ficus & Hotdog &        Chair & Mic &   Ship &  Drums & Materials \\
SparseNeRF~\cite{sparsenerf}& - &    23.02 & 18.19 & - & 26.20 & 23.26 & 20.81 & 19.21 & 20.80 \\
SparseNeRF \textit{wo.} depth & 22.92 &    24.00 & 18.84 & 27.52 & 27.11 & 23.35 & 21.84 & 19.17 & 21.50 \\ 
\text{SuGaR}~\cite{sugar} &21.78 &   23.77 & 23.08 & 22.36 & 25.72 & 18.72 & 21.09 & 19.55 & 19.94 \\
\text{ScaffoldGS}~\cite{scaffoldgs} &23.82 &      23.65 & 22.78 & 26.34 & 25.80 & \cellcolor[RGB]{\colorthird}28.28 & 21.17 & 20.47 & 22.06 \\
\text{Mip3DGS}~\cite{mipgs} &24.86 & 24.65 & 25.62 & 26.53 & 26.25 & \cellcolor[RGB]{\colorsecond}28.40 & 22.52 & 21.98 & \cellcolor[RGB]{\colorfirst}22.94 \\
\text{3DGS}~\cite{GaussianSplatting} &25.29 &    25.14 & 25.92 & 27.51 & 27.10 & \cellcolor[RGB]{\colorfirst}29.02 & 22.79 & \cellcolor[RGB]{\colorthird}22.10 & 22.71 \\
\text{Light3DGS}~\cite{lightgaussian} &25.39 & 25.08 & \cellcolor[RGB]{\colorfirst}27.53 & 27.10 & 27.40 & 28.04 & \cellcolor[RGB]{\colorthird}23.02 & 22.07 & \cellcolor[RGB]{\colorsecond}22.90 \\ 
\text{2DGS}~\cite{2dgs} &\cellcolor[RGB]{\colorsecond}25.62 &    \cellcolor[RGB]{\colorsecond}25.50 & 25.62 & \cellcolor[RGB]{\colorsecond}29.24 & \cellcolor[RGB]{\colorfirst}28.52 & 28.07 & \cellcolor[RGB]{\colorsecond}23.08 & \cellcolor[RGB]{\colorsecond}22.19 & \cellcolor[RGB]{\colorthird}22.75 \\
\noalign{\vskip 0.2em}\cdashline{1-10}\noalign{\vskip 0.2em}
\text{3DGS \textit{w.} $\mathcal{L}_{Moran}$} &\cellcolor[RGB]{\colorthird}25.44 &  \cellcolor[RGB]{\colorthird}25.26 & \cellcolor[RGB]{\colorsecond}26.55 & \cellcolor[RGB]{\colorthird}28.96 & \cellcolor[RGB]{\colorsecond}27.91 & 27.87 & 22.33 & 21.98 & 22.65 \\
\text{SplatFields3D} &\cellcolor[RGB]{\colorfirst}25.80 &    \cellcolor[RGB]{\colorfirst}26.98 & \cellcolor[RGB]{\colorthird}26.27 & \cellcolor[RGB]{\colorfirst}29.45 & \cellcolor[RGB]{\colorthird}27.42 & 27.60 & \cellcolor[RGB]{\colorfirst}23.78 & \cellcolor[RGB]{\colorfirst}22.55 & 22.32 \\
\midrule
&\multicolumn{9}{c}{6 Input Views} \\
SparseNeRF~\cite{sparsenerf}                    & -     & \cellcolor[RGB]{\colorthird}20.86 & 18.03 & -     & \cellcolor[RGB]{\colorthird}22.75 & 22.40 & \cellcolor[RGB]{\colorsecond}19.33 & 16.24 & \cellcolor[RGB]{\colorsecond}19.54 \\
SparseNeRF \textit{wo.} depth                   & \cellcolor[RGB]{\colorthird}20.86 & \cellcolor[RGB]{\colorfirst}22.62 & 17.63 & \cellcolor[RGB]{\colorsecond}25.84 & 22.65 & 20.72 & \cellcolor[RGB]{\colorfirst}19.85 & 17.25 & \cellcolor[RGB]{\colorfirst}20.30 \\
\text{SuGaR}~\cite{sugar}                       & 19.07 & 19.89 & 20.61 & 20.80 & 21.92 & 18.26 & 17.72 & 16.86 & 16.53 \\
\text{ScaffoldGS}~\cite{scaffoldgs}             & 19.65 & 18.21 & 20.72 & 19.48 & 22.20 & 24.31 & 16.47 & 17.21 & 18.62 \\
\text{Mip3DGS}~\cite{mipgs}                     & 20.04 & 19.39 & 21.81 & 19.70 & 21.72 & 24.44 & 17.02 & 17.72 & 18.52 \\
\text{3DGS}~\cite{GaussianSplatting}            & 20.62 & 19.80 & 22.25 & 21.16 & \cellcolor[RGB]{\colorthird}22.75 & \cellcolor[RGB]{\colorfirst}25.21 & 17.58 & 17.77 & 18.48 \\
\text{Light3DGS}~\cite{lightgaussian}           & 20.76 & 20.25 & \cellcolor[RGB]{\colorfirst}23.12 & 20.66 & 22.69 & \cellcolor[RGB]{\colorsecond}24.89 & 17.83 & \cellcolor[RGB]{\colorthird}18.02 & 18.63 \\
\text{2DGS}~\cite{2dgs}                         & 20.74 & 19.38 & 21.93 & 23.85 & \cellcolor[RGB]{\colorsecond}23.26 & 24.48 & 16.92 & 17.91 & 18.17 \\
\noalign{\vskip 0.2em}\cdashline{1-10}\noalign{\vskip 0.2em}
\text{3DGS \textit{w.} $\mathcal{L}_{Moran}$}   & \cellcolor[RGB]{\colorsecond}21.03 & 20.34 & \cellcolor[RGB]{\colorsecond}23.05 & \cellcolor[RGB]{\colorthird}23.92 & 22.50 & 24.64 & 17.20 & \cellcolor[RGB]{\colorsecond}18.14 & 18.48 \\
\text{SplatFields3D}                            & \cellcolor[RGB]{\colorfirst}22.26 & \cellcolor[RGB]{\colorsecond}22.41 & \cellcolor[RGB]{\colorthird}22.26 & \cellcolor[RGB]{\colorfirst}26.19 & \cellcolor[RGB]{\colorfirst}25.03 & \cellcolor[RGB]{\colorthird}24.84 & \cellcolor[RGB]{\colorsecond}19.33 & \cellcolor[RGB]{\colorfirst}18.97 & \cellcolor[RGB]{\colorthird}19.05 \\

\bottomrule
\end{tabular}
}
    \captionof{table}{
    \textbf{Sparse static scene reconstruction} of Blender~\cite{mildenhall2020nerf} scenes. 
    Reported numbers indicate PSNR metric on the novel views (``-'' denotes failed runs). Colors denote the \colorbox[RGB]{\colorfirst}{1st}, \colorbox[RGB]{\colorsecond}{2nd}, and \colorbox[RGB]{\colorthird}{3rd} best-performing model. See Sec.~\ref{subsec:exp_static} for discussion
    }
    \label{tab:exp_static_blender}
  \end{minipage}
\end{figure}

\begin{table}[t]
  \caption{
    \textbf{Ablation study of SplatFields}. Blender dataset~\cite{mildenhall2020nerf}, setup from Tab.~\ref{tab:exp_static_blender}  }
  \label{tab:exp_static_blender_ablation}
  \centering
  \setlength{\tabcolsep}{2.0pt} %

\scriptsize
\resizebox{\textwidth}{!}{%
\begin{tabular}{@{}l|cccccccccc@{}}

\toprule

& \multicolumn{2}{c}{12  Views}  & \multicolumn{2}{c}{10  Views}   & \multicolumn{2}{c}{8  Views}   & \multicolumn{2}{c}{6  Views}  & \multicolumn{2}{c}{4  Views} \\
& SSIM$\uparrow$ & PSNR$\uparrow$  & SSIM$\uparrow$ & PSNR$\uparrow$   & SSIM$\uparrow$ & PSNR$\uparrow$   & SSIM$\uparrow$ & PSNR$\uparrow$  & SSIM$\uparrow$ & PSNR$\uparrow$ \\ \midrule

basic (MLP-only) & 89.63 & 24.82& 88.69 & 23.93& 87.79 & \cellcolor[RGB]{\colorsecond}23.32& 85.70 & \cellcolor[RGB]{\colorthird}21.58& \cellcolor[RGB]{\colorthird}81.98 & \cellcolor[RGB]{\colorthird}19.07\\
+$\mathcal{L}_2$-norm reg.& \cellcolor[RGB]{\colorthird}89.66 & \cellcolor[RGB]{\colorthird}24.98& \cellcolor[RGB]{\colorthird}88.81 & \cellcolor[RGB]{\colorthird}24.21& \cellcolor[RGB]{\colorthird}87.98 & \cellcolor[RGB]{\colorthird}23.64& \cellcolor[RGB]{\colorsecond}85.93 & \cellcolor[RGB]{\colorsecond}21.79& \cellcolor[RGB]{\colorfirst}82.44 & \cellcolor[RGB]{\colorfirst}19.46\\
+tri-CNN  & \cellcolor[RGB]{\colorsecond}90.83 & \cellcolor[RGB]{\colorsecond}25.23& \cellcolor[RGB]{\colorsecond}90.04 & \cellcolor[RGB]{\colorsecond}24.66& \cellcolor[RGB]{\colorsecond}88.01 & 23.19& \cellcolor[RGB]{\colorthird}85.91 & 21.46& 81.28 & 18.72\\
\noalign{\vskip 0.2em} \cdashline{1-11} \noalign{\vskip 0.2em}
full model & \cellcolor[RGB]{\colorfirst}91.18 & \cellcolor[RGB]{\colorfirst}25.80& \cellcolor[RGB]{\colorfirst}90.32 & \cellcolor[RGB]{\colorfirst}24.94& \cellcolor[RGB]{\colorfirst}88.94 & \cellcolor[RGB]{\colorfirst}23.98& \cellcolor[RGB]{\colorfirst}86.62 & \cellcolor[RGB]{\colorfirst}22.26& \cellcolor[RGB]{\colorsecond}82.27 & \cellcolor[RGB]{\colorsecond}19.16\\

\bottomrule
\end{tabular}
}
\end{table}

\textbf{Static reconstruction from sparse views.} 
We benchmark SplatFields on Blender~\cite{mildenhall2020nerf} under 6 and 12 input views (see Sup. Mat. for more extensive benchmarking). The main goal of this section is to showcase the efficiency of the utilized spatial regularization for 3DGS methods. We, therefore, focus on comparison against the recent splat-based techniques \cite{sugar,mipgs,GaussianSplatting,lightgaussian} and SparseNeRF~\cite{sparsenerf}, leaving the comprehensive comparison against a broader range of NeRF-based methods for more challenging dynamic scenarios considered in Sec.~\ref{subsec:exp_dyn}.

\begin{figure}[t]
    \centering
    \scriptsize
    \setlength{\tabcolsep}{0.0mm} %
    \newcommand{\sz}{0.2}  %
    \renewcommand{\arraystretch}{0.0} 
    \resizebox{\textwidth}{!}{%
    \begin{tabular}{ccccc}  %
        GT & SplatFields3D & 2DGS~\cite{2dgs} & 3DGS~\cite{GaussianSplatting} & ZeroRF~\cite{zerorf} \\

\begin{tikzpicture} 
\node[anchor=south west,inner sep=0] (image) at (0,0) {
    \includegraphics[width=\sz\linewidth, trim={50 0 50 0},clip]{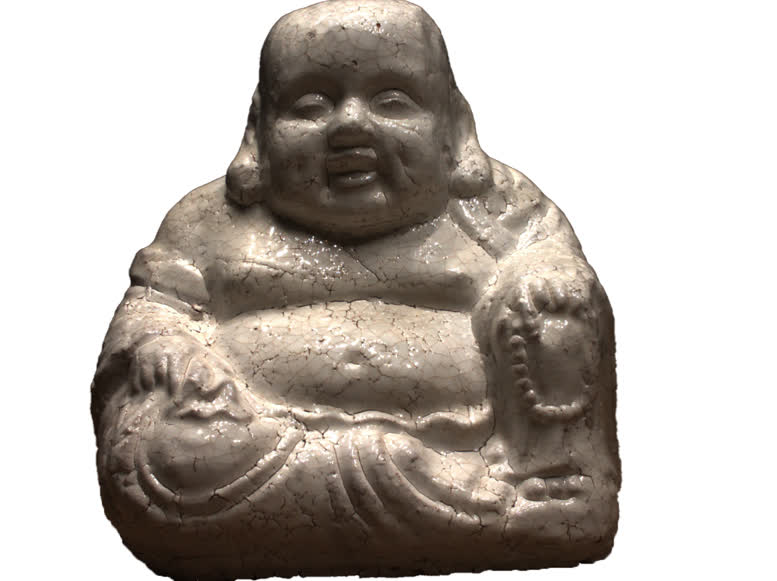}
}; 
\begin{scope}[x={(image.south east)},y={(image.north west)}]
\draw[black,thick] (0.25,0.45) rectangle (0.7,0.85); %
\end{scope} 
\end{tikzpicture}

& %
\begin{tikzpicture} 
\node[anchor=south west,inner sep=0] (image) at (0,0) {
    \includegraphics[width=\sz\linewidth, trim={50 0 50 0},clip]{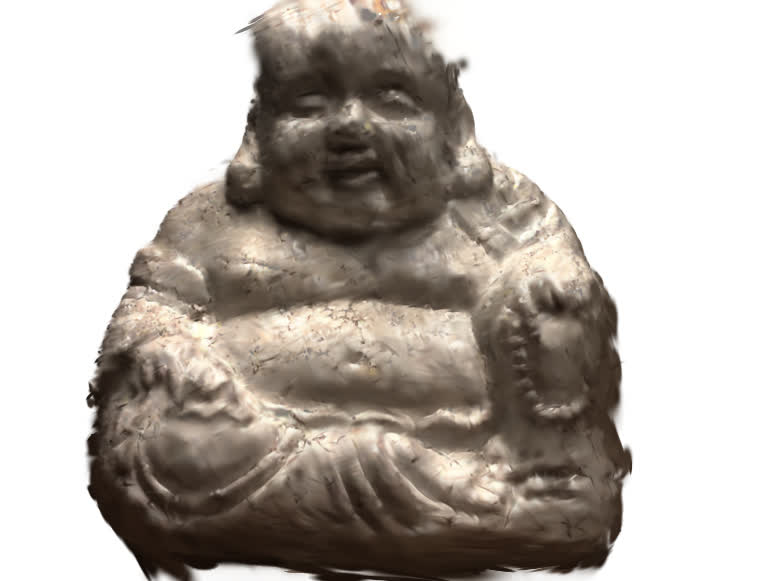}
}; 
\begin{scope}[x={(image.south east)},y={(image.north west)}]
\draw[green,thick] (0.25,0.45) rectangle (0.7,0.85); %
\end{scope} 
\end{tikzpicture}

& %
\begin{tikzpicture} 
\node[anchor=south west,inner sep=0] (image) at (0,0) {
    \includegraphics[width=\sz\linewidth, trim={50 0 50 0},clip]{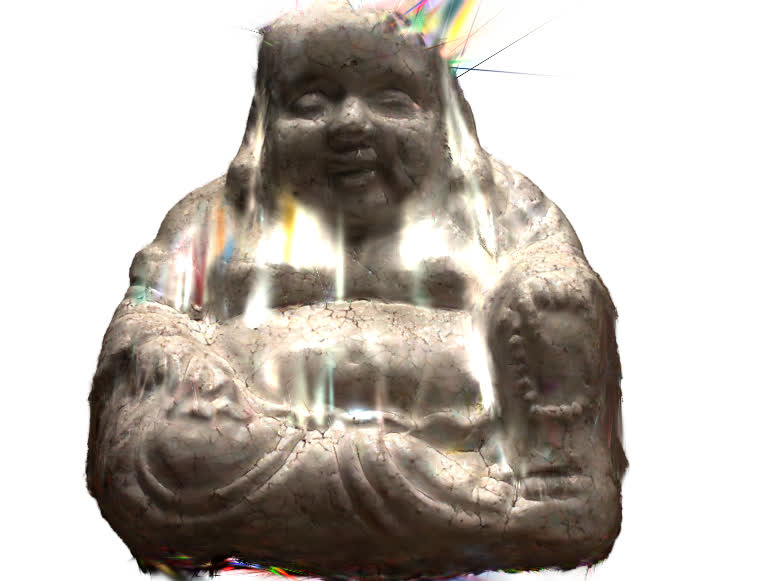}
}; 
\begin{scope}[x={(image.south east)},y={(image.north west)}]
\draw[red,thick] (0.25,0.45) rectangle (0.7,0.85); %
\end{scope} 
\end{tikzpicture}

& %
\begin{tikzpicture} 
\node[anchor=south west,inner sep=0] (image) at (0,0) {
    \includegraphics[width=\sz\linewidth, trim={50 0 50 0},clip]{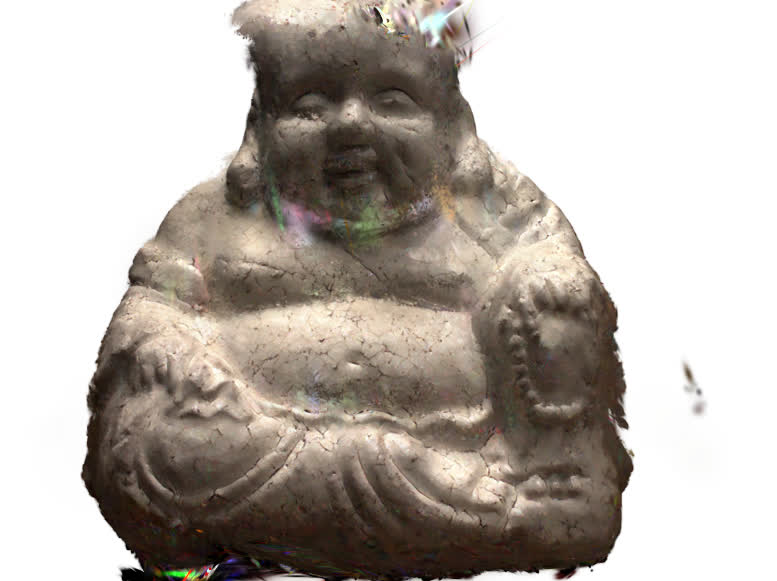}
}; 
\begin{scope}[x={(image.south east)},y={(image.north west)}]
\draw[red,thick] (0.25,0.45) rectangle (0.7,0.85); %
\end{scope} 
\end{tikzpicture}
& %
\begin{tikzpicture} 
\node[anchor=south west,inner sep=0] (image) at (0,0) {
    \includegraphics[width=\sz\linewidth, trim={50 0 50 0},clip]{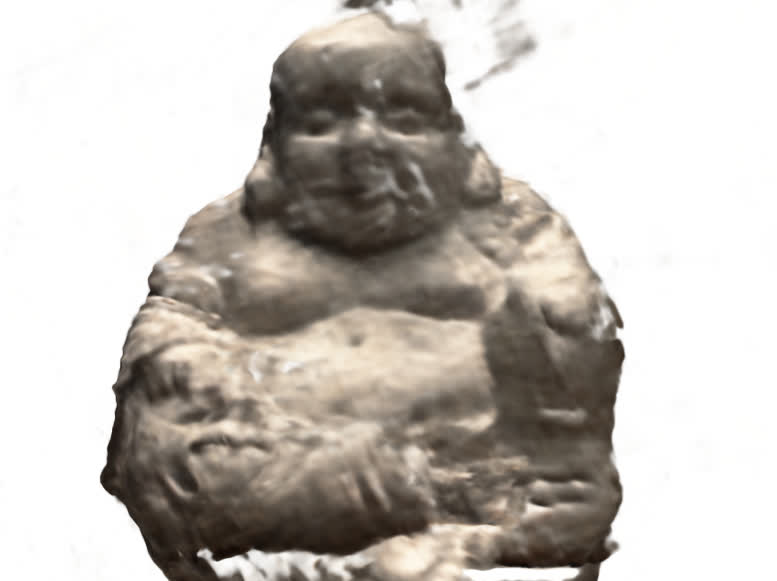}
}; 
\begin{scope}[x={(image.south east)},y={(image.north west)}]
\draw[red,thick] (0.25,0.45) rectangle (0.7,0.85); %
\end{scope} 
\end{tikzpicture}
\\ \noalign{\vskip 1.0em}

    \includegraphics[width=\sz\linewidth, trim={50 0 50 125},clip]{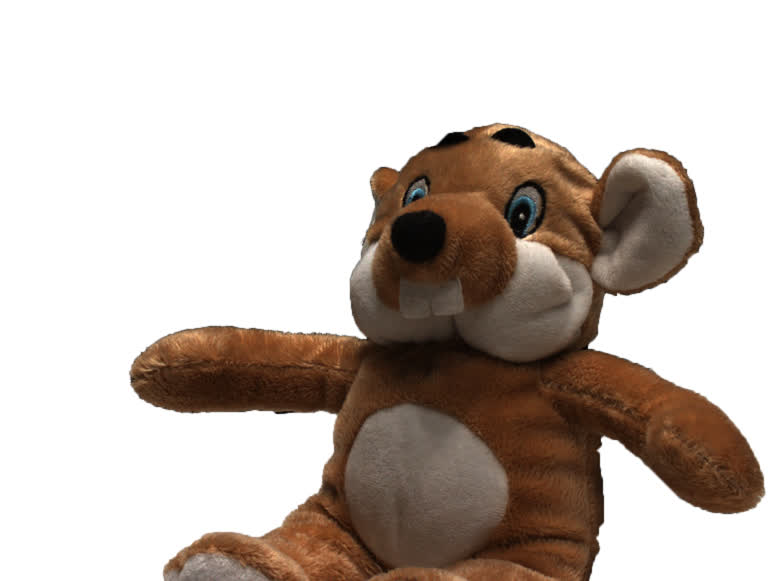}

& %
    \includegraphics[width=\sz\linewidth, trim={50 0 50 125},clip]{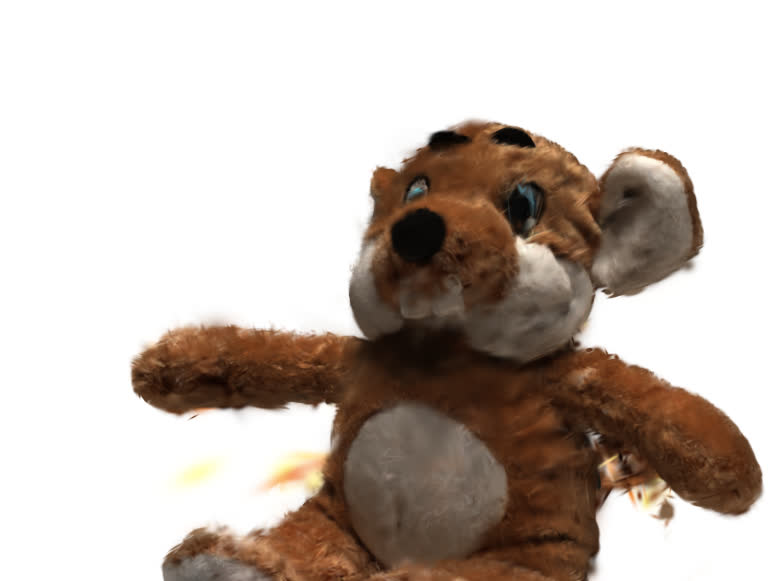}

& %
    \includegraphics[width=\sz\linewidth, trim={50 0 50 125},clip]{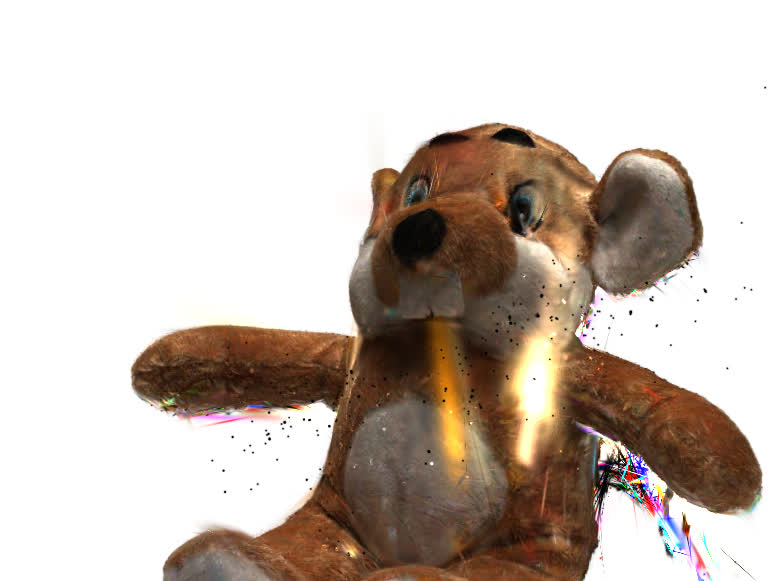}

& %
    \includegraphics[width=\sz\linewidth, trim={50 0 50 125},clip]{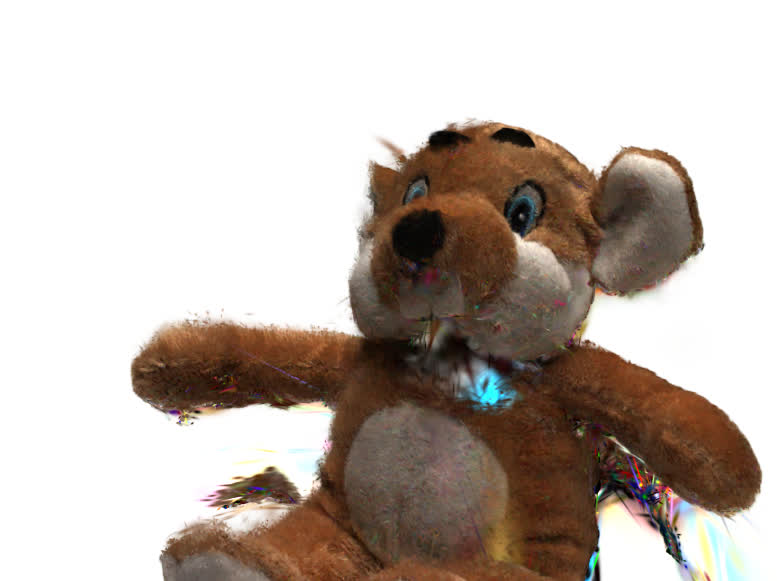}
& %
    \includegraphics[width=\sz\linewidth, trim={50 0 50 125},clip]{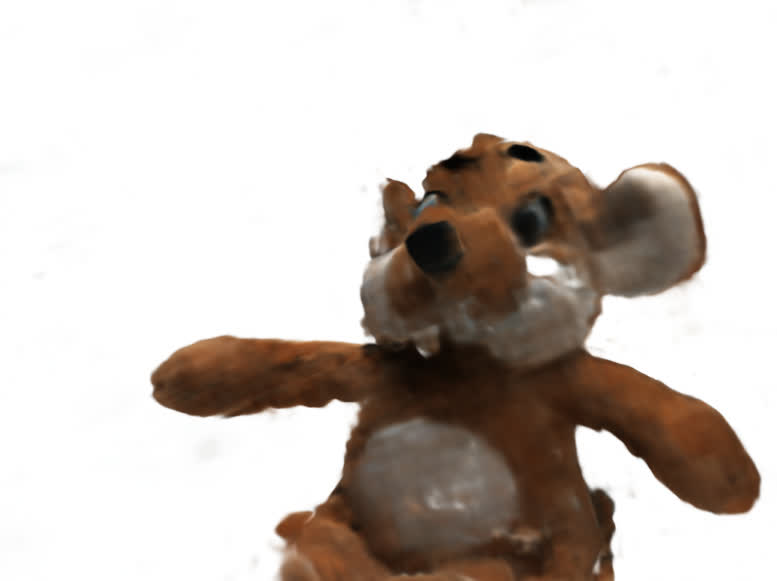}
\\ \noalign{\vskip 1.0em}
    \includegraphics[width=\sz\linewidth, trim={50 0 50 0},clip]{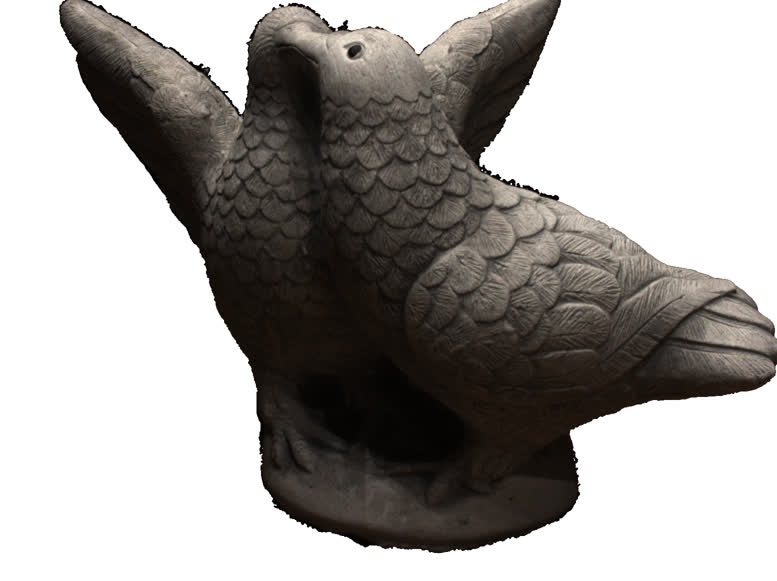}

& %
    \includegraphics[width=\sz\linewidth, trim={50 0 50 0},clip]{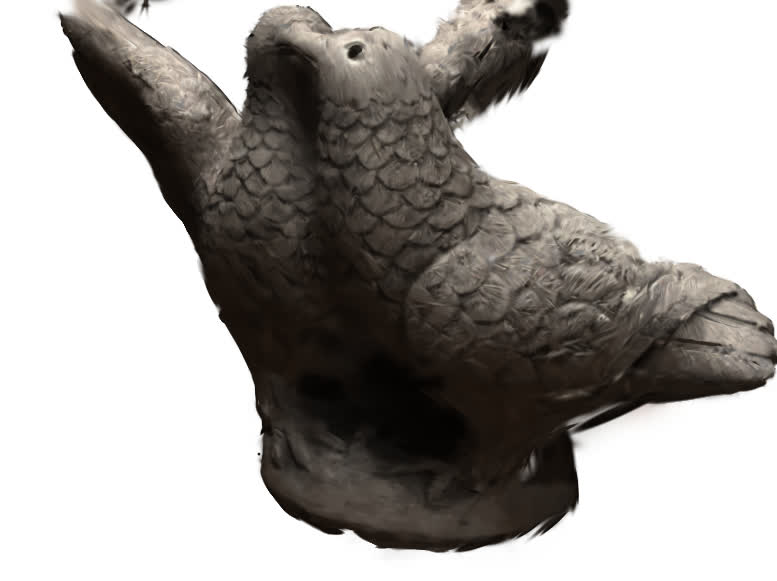}

& %
    \includegraphics[width=\sz\linewidth, trim={50 0 50 0},clip]{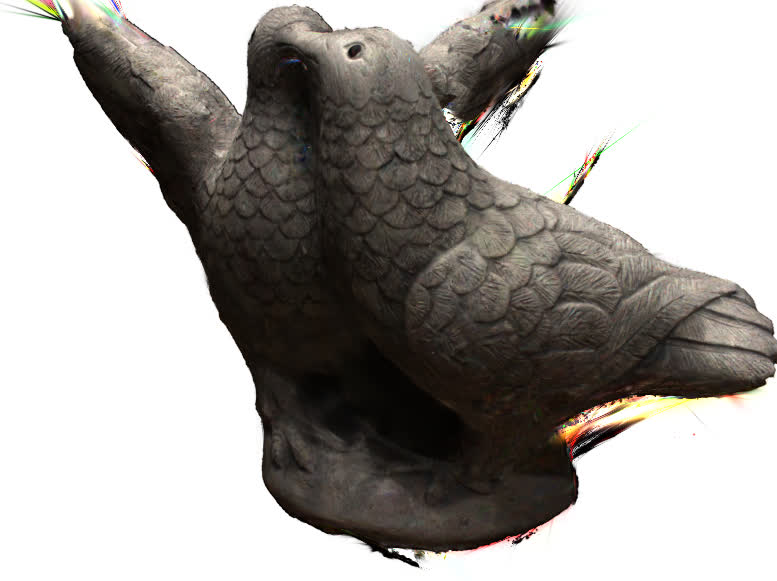}

& %
    \includegraphics[width=\sz\linewidth, trim={50 0 50 0},clip]{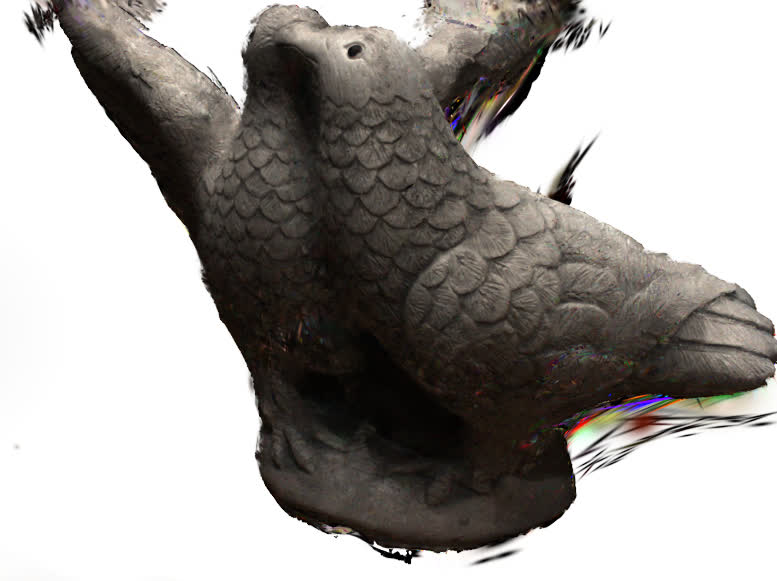}
& %
    \includegraphics[width=\sz\linewidth, trim={50 0 50 0},clip]{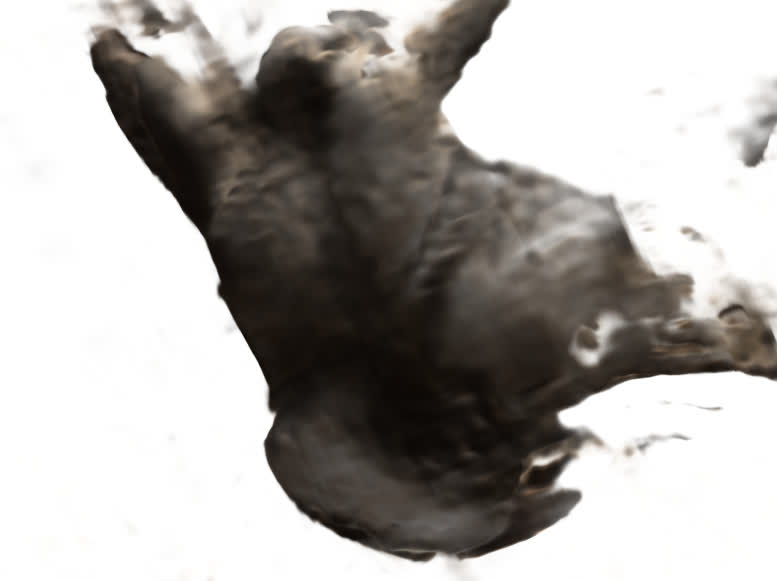}
\\ \noalign{\vskip 1.0em}
    \includegraphics[width=\sz\linewidth, trim={50 0 50 0},clip]{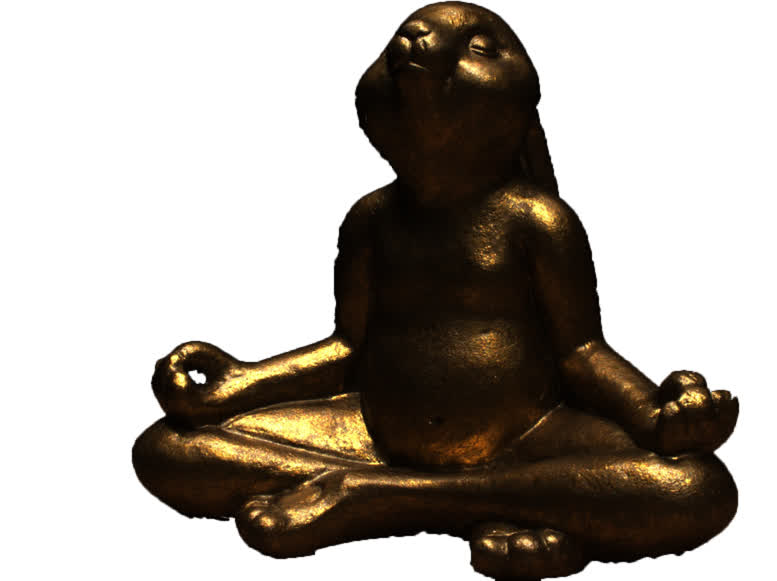}

& %
    \includegraphics[width=\sz\linewidth, trim={50 0 50 0},clip]{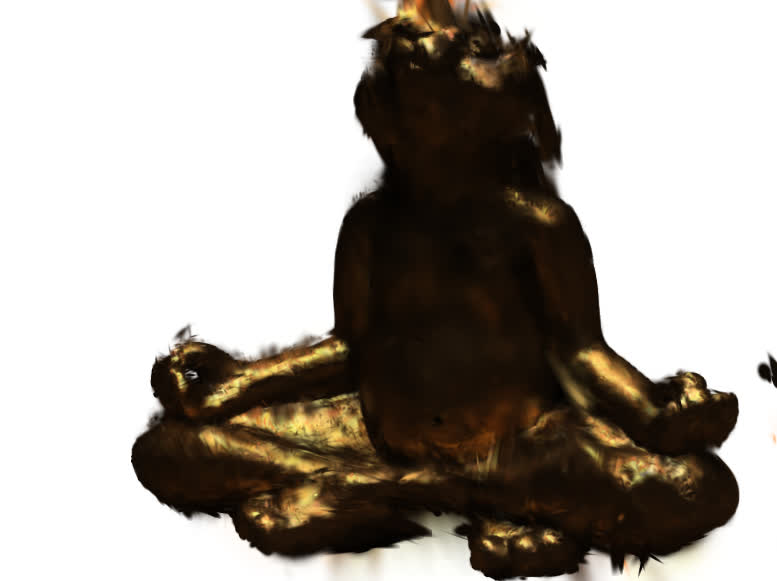}

& %
    \includegraphics[width=\sz\linewidth, trim={50 0 50 0},clip]{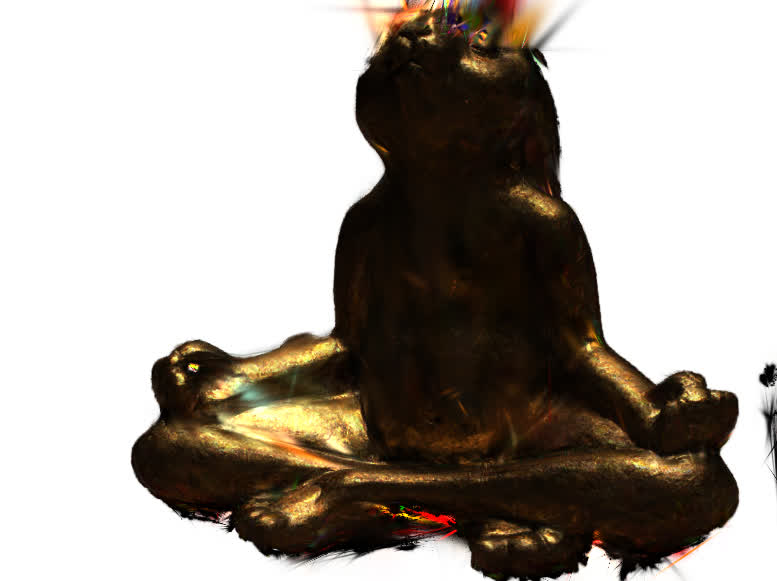}

& %
    \includegraphics[width=\sz\linewidth, trim={50 0 50 0},clip]{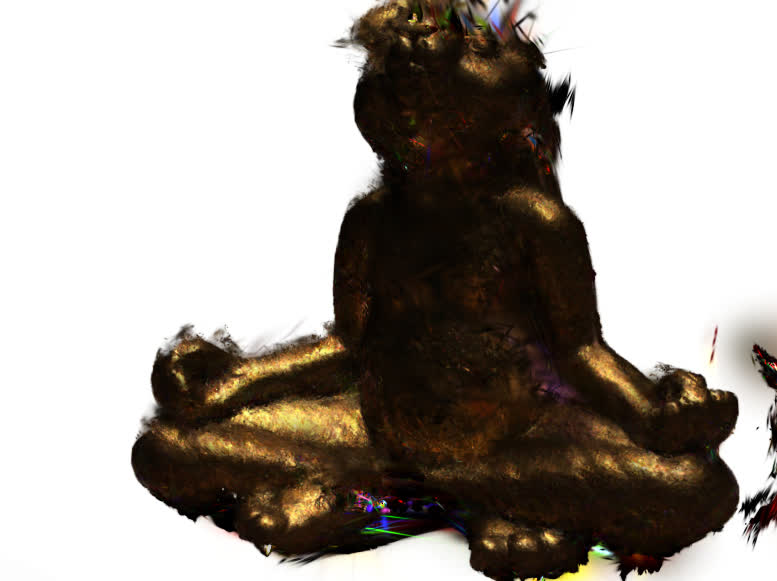}
& %
    \includegraphics[width=\sz\linewidth, trim={50 0 50 0},clip]{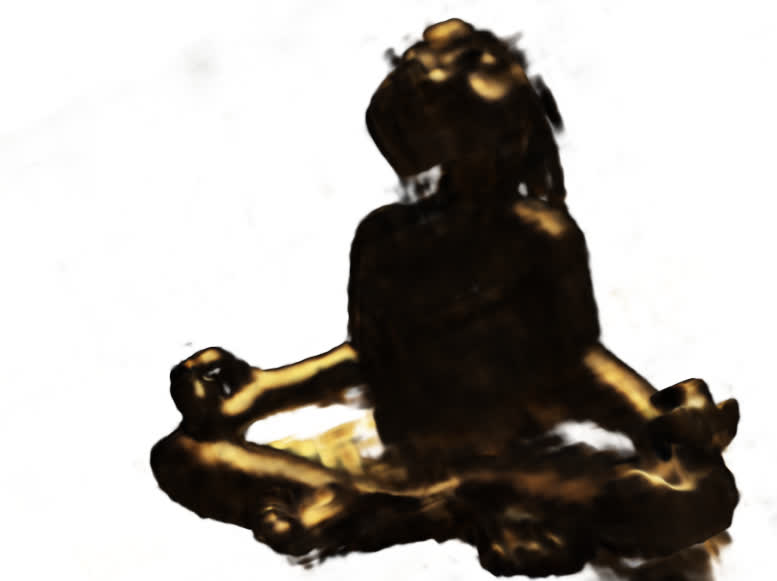}
\\ \noalign{\vskip 1.0em}

    \includegraphics[width=\sz\linewidth, trim={50 0 50 0},clip]{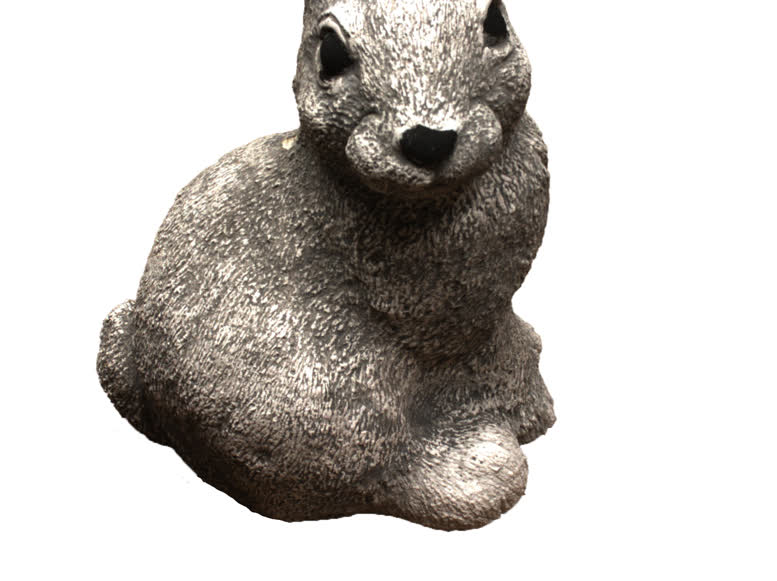}

& %
    \includegraphics[width=\sz\linewidth, trim={50 0 50 0},clip]{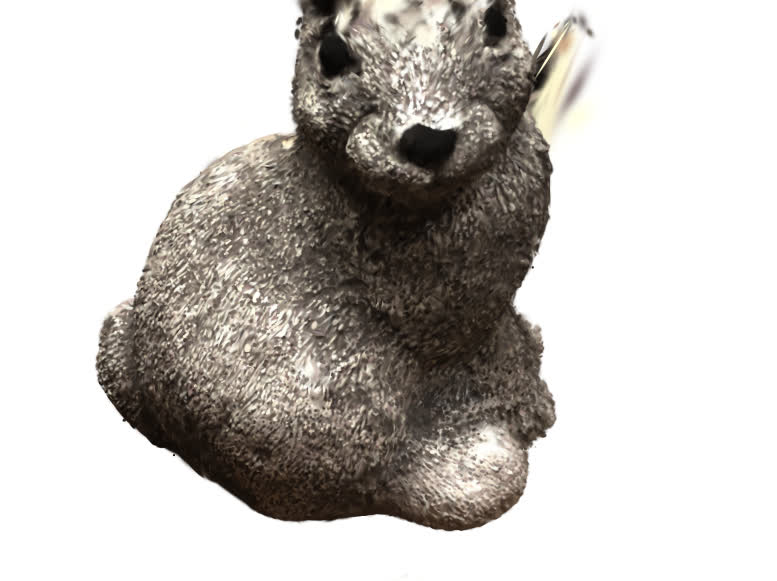}

& %
    \includegraphics[width=\sz\linewidth, trim={50 0 50 0},clip]{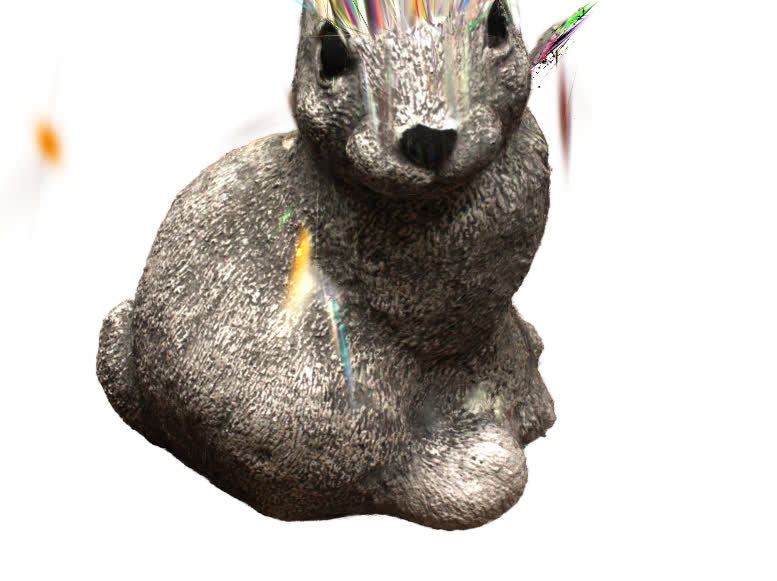}

& %
    \includegraphics[width=\sz\linewidth, trim={50 0 50 0},clip]{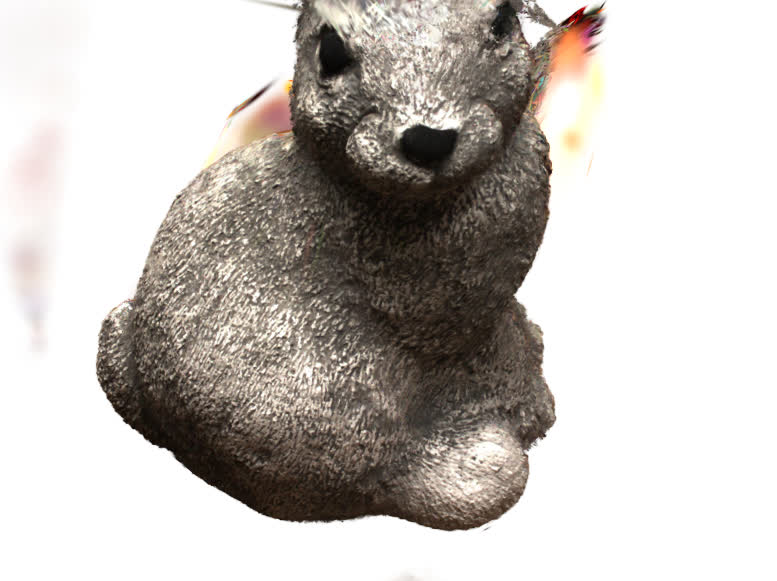}
& %
    \includegraphics[width=\sz\linewidth, trim={50 0 50 0},clip]{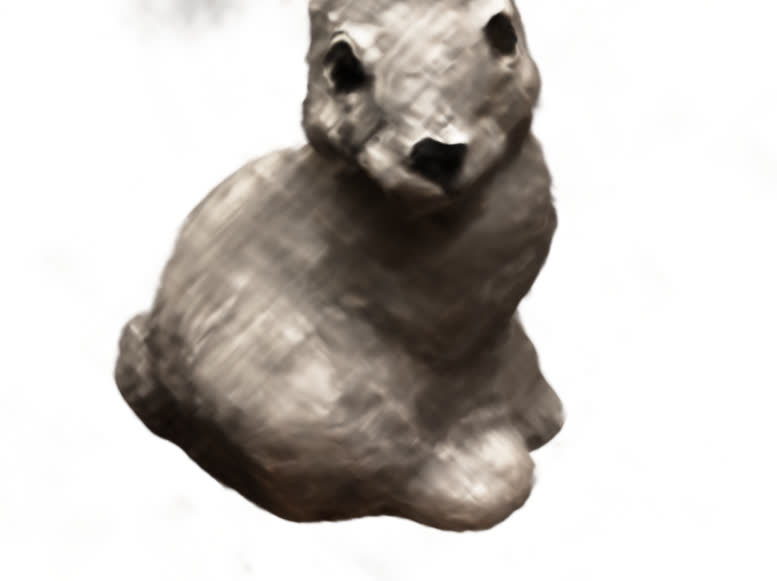}
\\

        Mean PSNR$\uparrow$ & \textbf{21.07} & 20.70 & 19.40 & 19.10 \\
    \end{tabular}
    }
    \caption{\textbf{Three-view reconstruction on DTU~\cite{DTU}}; PSNR are averaged across all 15 scenes. See Tab.~\ref{tab:app:dtu_comparison} for the individualized scores.}
    \label{fig:dtu_static}
\end{figure}

Extensive quantitative results presented in Tab.~\ref{tab:exp_static_blender} demonstrate that SplatFields consistently outperforms the respective baselines across varying numbers of input views. The achieved improvement is also verified by visually sharper reconstructions outlined in Fig.~\ref{fig:exp_static}. More importantly, we observe that the relative gap in performance between our method and the baselines is increasing as the input views become more scarce, which confirms our intuition of the spatial bias as a powerful regularizer in such scenarios. 
We further validate our improvement on the real-world DTU dataset~\cite{DTU} for the challenging task of 3-view reconstruction (Fig.~\ref{fig:dtu_static}) and demonstrate consistent improvement over NeRF-~\cite{zerorf} and splatting-based~\cite{2dgs,GaussianSplatting} baselines. See Sup. Mat. for more extensive evaluation. 

Please note that all the methods, including ours, demonstrate real-time rendering performance with high interactive rates (120+FPS) during test time since the generator $g_\theta$ is discarded after the training completion. We refer the reader to the supplementary for further details. 

\textbf{Ablation of the triplane CNN generator.} We validate the impact of the proposed triplane CNN generator on the performance of the SplatFields model in Tab.~\ref{tab:exp_static_blender_ablation}. Here, the basic pipeline implies using only the set of MLPs to directly predict the splat rendering features (opacity, scale, etc.) and point displacements from the initial splat locations, without conditioning on the deep features produced by the triplane CNN.  Results indicate that utilizing the deep features regressed by a CNN improves the quality, with the splat $\mathcal{L}_2$-norm regularization term further benefiting the reconstruction. Note that the regularization has a marginal improvement on the results of our pipeline that does not utilize the CNN feature generator, demonstrating the synergy of both modeling strategies. 

\subsection{Dynamic Scene Reconstruction}  \label{subsec:exp_dyn}

\begin{figure}[t] 
    \centering
    \begin{minipage}{1.0\textwidth}
      \scriptsize
      \setlength{\tabcolsep}{0.1mm} %
      \newcommand{\sz}{0.242}  %
      \renewcommand{\arraystretch}{0.0} 
      \setlength{\tabcolsep}{1.2pt} %
  
      \captionof{table}{
      \textbf{Monocular reconstruction} of dynamic sequences from the NeRF-DS dataset~\cite{nerfds} with recent state-of-the-art methods.  
          The forward slash in FPS indicates the rendering speed without the neural network inference when the rendering primitives are extracted and stored for each frame \textit{vs.} with the neural network inference 
      }
      \resizebox{\textwidth}{!}{%
      \begin{tabular}{@{}l|cc|cc|cccccccc@{}}
      \toprule
      
      &\multicolumn{2}{c}{Resources} &\multicolumn{2}{c}{\textit{mean}$\uparrow$} & \multicolumn{8}{c}{LPIPS$\downarrow$ ($\times 10^2$)} \\
      & FPS $\uparrow$ & t $\downarrow$ & PSNR & SSIM &\textit{mean} & Sieve & Plate & Bell & Press & Cup & As & Basin \\
      \midrule
      3D-GS~\cite{GaussianSplatting}           &120+&15 \textit{m.}& 20.29&78.16&29.20 & 22.47 & 40.93 & 25.03 & 29.04 & 25.48 & 29.94 & 31.53 \\ 
      TiNeuVox~\cite{TiNeuVox}        &< 1 & 30 \textit{m.} & 21.61&82.34&27.66 & 31.76 & 33.17 & 25.68 & 30.01 & 36.43 & 39.67 & 26.90 \\ 
      4DGaussians~\cite{wu4dgaussiansRealTime} &120+/50&30 \textit{m.} & 23.68&83.22&21.06 & 16.39 & \cellcolor[RGB]{\colorthird}23.80 & 21.84 & 21.68 & 19.06 & 22.06 & 22.57 \\
      HyperNeRF~\cite{HyperNeRF}       &< 1&1 \textit{d.}& 23.45&\cellcolor[RGB]{\colorthird}84.88&19.90 & 16.45 & 29.40 & 20.52 & \cellcolor[RGB]{\colorsecond}19.59 & \cellcolor[RGB]{\colorthird}16.50 & \cellcolor[RGB]{\colorthird}17.77 & \cellcolor[RGB]{\colorthird}19.11 \\ 
      
      Deformable3DGS~\cite{yang2023deformable3dgs} &120+/30& 1 \textit{h.} & \cellcolor[RGB]{\colorthird}23.54&84.05&\cellcolor[RGB]{\colorthird}19.79 & \cellcolor[RGB]{\colorsecond}15.30 & 25.04 & \cellcolor[RGB]{\colorfirst}15.93 & 29.89 & \cellcolor[RGB]{\colorfirst}15.38 & 17.88 & \cellcolor[RGB]{\colorsecond}19.10 \\ 
      NeRF-DS~\cite{nerfds} &< 1&1 \textit{d.} & \cellcolor[RGB]{\colorsecond}23.60&\cellcolor[RGB]{\colorsecond}84.94&\cellcolor[RGB]{\colorsecond}18.16 & \cellcolor[RGB]{\colorfirst}14.72 & \cellcolor[RGB]{\colorfirst}19.96 & \cellcolor[RGB]{\colorthird}18.67 & \cellcolor[RGB]{\colorthird}20.47 & 17.37 & \cellcolor[RGB]{\colorfirst}17.41 & \cellcolor[RGB]{\colorfirst}18.55 \\ \noalign{\vskip 0.2em}
      \cdashline{1-13} \noalign{\vskip 0.2em}
      SplatFields4D  &120+/30&1 \textit{h.}& \cellcolor[RGB]{\colorfirst}23.84&\cellcolor[RGB]{\colorfirst}85.17& \cellcolor[RGB]{\colorfirst}17.86 & \cellcolor[RGB]{\colorfirst}14.72 & \cellcolor[RGB]{\colorsecond}22.43 & \cellcolor[RGB]{\colorsecond}16.10 & \cellcolor[RGB]{\colorfirst}19.26 & \cellcolor[RGB]{\colorsecond}15.67 & \cellcolor[RGB]{\colorsecond}17.71 & \cellcolor[RGB]{\colorthird}19.11 \\ 
      
      \bottomrule
      \end{tabular}
      }
      \label{tab:exp_nerfds}
    \end{minipage}\hfill
    \vspace{2.0em}
    \begin{minipage}{1.0\textwidth}
      \scriptsize
      \setlength{\tabcolsep}{0.1mm} %
      \newcommand{\sz}{0.242}  %
      \renewcommand{\arraystretch}{0.0} 
      \resizebox{\textwidth}{!}{%
      \begin{tabular}{ccccc}  %
              & \text{4DGaussians}~\cite{wu4dgaussiansRealTime} & \text{Deformable3DGS}~\cite{yang2023deformable3dgs} & SplatFields4D & Ground Truth \\[0.5em]
          \rotatebox{90}{\phantom{++..}Press} 
  & 
  \begin{tikzpicture} 
  \node[anchor=south west,inner sep=0] (image) at (0,0) {
      \includegraphics[width=\sz\linewidth]{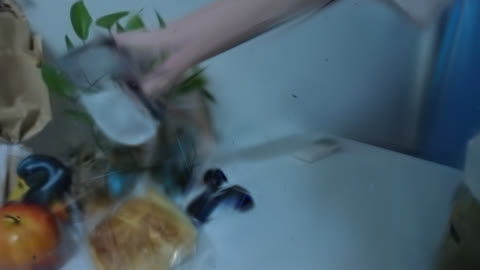}
  }; 
  \begin{scope}[x={(image.south east)},y={(image.north west)}]
  \draw[red,thick] (0.15,0.02) rectangle (0.45,0.40); %
  \end{scope} 
  \end{tikzpicture}
  & 
  \begin{tikzpicture} 
  \node[anchor=south west,inner sep=0] (image) at (0,0) {
      \includegraphics[width=\sz\linewidth]{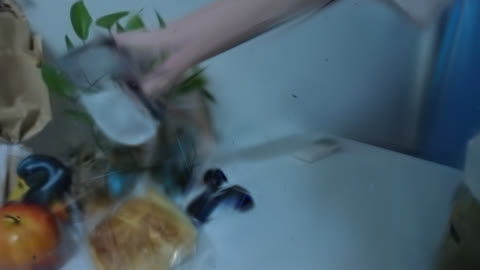}
  }; 
  \begin{scope}[x={(image.south east)},y={(image.north west)}]
  \draw[red,thick] (0.15,0.02) rectangle (0.45,0.40); %
  \end{scope} 
  \end{tikzpicture}
  & 
  \begin{tikzpicture} 
  \node[anchor=south west,inner sep=0] (image) at (0,0) {
      \includegraphics[width=\sz\linewidth]{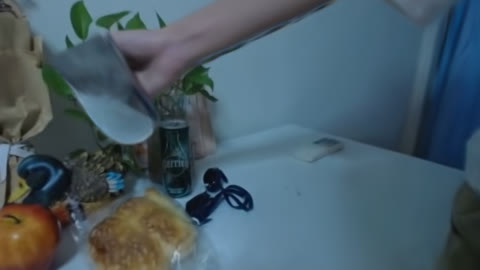}
  }; 
  \begin{scope}[x={(image.south east)},y={(image.north west)}]
  \draw[green,thick] (0.15,0.02) rectangle (0.45,0.40); %
  \end{scope} 
  \end{tikzpicture}
  & 
  \begin{tikzpicture} 
  \node[anchor=south west,inner sep=0] (image) at (0,0) {
      \includegraphics[width=\sz\linewidth]{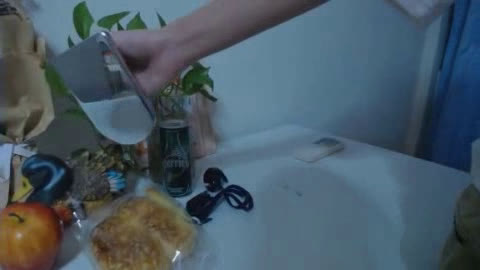}
  }; 
  \begin{scope}[x={(image.south east)},y={(image.north west)}]
  \draw[black,thick] (0.15,0.02) rectangle (0.45,0.40); %
  \end{scope} 
  \end{tikzpicture}
  \\[0.5em]
  
     \rotatebox{90}{\phantom{++.}Plate} 
  & 
  \begin{tikzpicture} 
  \node[anchor=south west,inner sep=0] (image) at (0,0) {
      \includegraphics[width=\sz\linewidth]{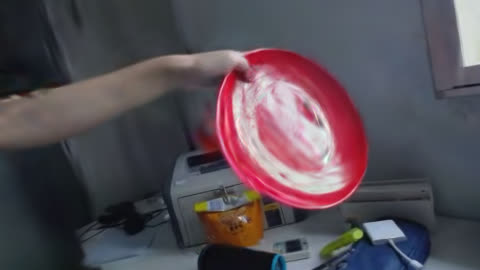}
  }; 
  \begin{scope}[x={(image.south east)},y={(image.north west)}]
  \draw[red,thick] (0.32,0.40) rectangle (0.65,0.75); %
  \end{scope} 
  \end{tikzpicture}
  & 
  \begin{tikzpicture} 
  \node[anchor=south west,inner sep=0] (image) at (0,0) {
      \includegraphics[width=\sz\linewidth]{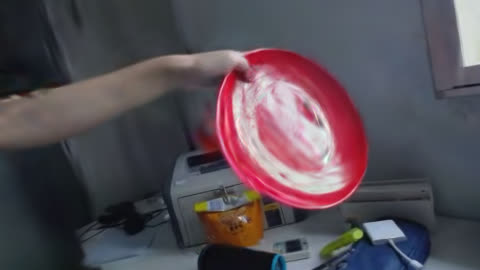}
  }; 
  \begin{scope}[x={(image.south east)},y={(image.north west)}]
  \draw[red,thick] (0.32,0.40) rectangle (0.65,0.75); %
  \end{scope} 
  \end{tikzpicture}
  &
  \begin{tikzpicture} 
  \node[anchor=south west,inner sep=0] (image) at (0,0) {
      \includegraphics[width=\sz\linewidth]{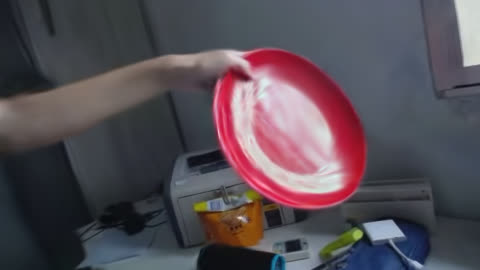}
  }; 
  \begin{scope}[x={(image.south east)},y={(image.north west)}]
  \draw[green,thick] (0.32,0.40) rectangle (0.65,0.75); %
  \end{scope} 
  \end{tikzpicture}
  &
  \begin{tikzpicture} 
  \node[anchor=south west,inner sep=0] (image) at (0,0) {
      \includegraphics[width=\sz\linewidth]{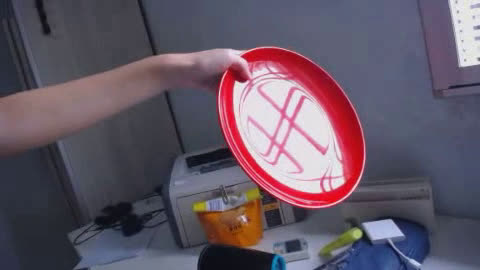}
  }; 
  \begin{scope}[x={(image.south east)},y={(image.north west)}]
  \draw[black,thick] (0.32,0.40) rectangle (0.65,0.75); %
  \end{scope} 
  \end{tikzpicture}
  \\
      \end{tabular}
      }
      \caption{\textbf{Monocular reconstruction} of sequences from \cite{nerfds} on the setup from Tab.~\ref{tab:exp_nerfds}. }
      \label{fig:exp_nerfds}
    \end{minipage}
  \end{figure}

\textbf{Monocular dynamic reconstruction.} 
We further evaluate our method on seven sequences of varying lengths (ranging from 424 to 881 frames) from the NeRF-DS dataset~\cite{nerfds} and compare our method against both NeRF- \cite{HyperNeRF,TiNeuVox,nerfds} and 3DGS-based~\cite{wu4dgaussiansRealTime,yang2023deformable3dgs} dynamic reconstruction baselines. 

The results are reported in Tab.~\ref{tab:exp_nerfds}, and the qualitative comparison of the 3DGS-based methods is presented in Fig.~\ref{fig:exp_nerfds}.
The results demonstrate that our method achieves competitive or superior reconstruction quality across all sequences, while also maintaining real-time rendering capabilities and facilitating accelerated training processes.
However, we observed that the sequences in the dataset involve relatively small motion and large static parts. Therefore, we further analyze SplatFields on multi-view sequences with more challenging dynamics.

\begin{table}
  \caption{
    \textbf{Multi-view reconstruction} of dynamic sequences from the Owlii dataset ~\cite{mildenhall2020nerf} under varying number of input views. The reported metric is PSNR averaged across novel views. 
    The forward slash in FPS indicates the rendering speed without the neural network inference when the rendering primitives are extracted and stored for each frame \textit{vs.} with the neural network inference. 
  }
  \label{tab:exp_dyn_owlii}
  \centering
  \scriptsize
  \setlength{\tabcolsep}{4.3pt} %

\resizebox{\textwidth}{!}{%
\begin{tabular}{@{}cl|cc|ccccc@{}}
\toprule

 && \multicolumn{2}{c}{Resources} &\multicolumn{5}{c}{10 Input Views} \\
 && FPS$\uparrow$ & t$\downarrow$ & \textit{mean} & Dancer &     Exercise &      Model & Basketball \\
 \multicolumn{1}{c}{\multirow{5}{*}{\rotatebox{90}{4D NeRFs}}}
 &DyNeRF~\cite{DyNeRF,ResFields}            &\multirow{5}{*}{<1}&1 day&29.70 &  28.22 & 30.64 & 29.95 & 30.00 \\
 &TNeRF~\cite{DyNeRF,ResFields}             &&1 day&30.39 &   29.12 & 31.00 & 30.71 & 30.71 \\
 &DNeRF~\cite{DNeRF,ResFields}              &&1.5 day&30.25 &   29.39 & 30.63 & 30.63 & 30.35 \\
 &Nerfies~\cite{Nerfies,ResFields}          &&1.5 day&30.70 & 29.57 & 31.08 & 30.53 & 31.60 \\
 &HyperNeRF~\cite{HyperNeRF,ResFields}      &&2 days&30.36 &       30.09 & 30.39 & 30.88 & 30.08 \\\noalign{\vskip 0.2em}
 
 \cdashline{1-9} \noalign{\vskip 0.2em}
 
 \multicolumn{1}{c}{\multirow{7}{*}{\rotatebox{90}{Splatting}}}
&\text{4D-GS}~\cite{yang4DGS}&120+&10h            &\cellcolor[RGB]{\colorthird}28.05 & \cellcolor[RGB]{\colorthird}28.11 & \cellcolor[RGB]{\colorthird}29.09 & \cellcolor[RGB]{\colorthird}29.06 & 25.94 \\
&\text{Deformable3DGS}~\cite{yang2023deformable3dgs}&120+/30&8h   &27.76 & 27.86 & 28.78 & 26.47 & \cellcolor[RGB]{\colorthird}27.95 \\
&\text{4DGaussians}~\cite{wu4dgaussiansRealTime}&120+/50&2h      &\cellcolor[RGB]{\colorsecond}29.80 & \cellcolor[RGB]{\colorsecond}28.46 & \cellcolor[RGB]{\colorsecond}30.21 & \cellcolor[RGB]{\colorsecond}30.69 & \cellcolor[RGB]{\colorsecond}29.82 \\\noalign{\vskip 0.2em}
\cdashline{2-9} \noalign{\vskip 0.2em}
& \text{SplatFields4D (30k it)} &120+/30&2h&30.88 &    30.46 & 30.78 & 31.14 & 31.15 \\
& \text{SplatFields4D (40k it)} &120+/30&3h&30.96 &    30.57 & 30.85 & 31.20 & 31.24 \\
& \text{SplatFields4D (100k it)} & 120+/30 &7h &31.12 &  30.79 & 30.99 & 31.33 & 31.39 \\
&\text{SplatFields4D (200k it)}&120+/30&14h&\cellcolor[RGB]{\colorfirst}31.32 & \cellcolor[RGB]{\colorfirst}31.05 & \cellcolor[RGB]{\colorfirst}31.16 & \cellcolor[RGB]{\colorfirst}31.50 & \cellcolor[RGB]{\colorfirst}31.58 \\
\midrule

 && \multicolumn{2}{c}{} &\multicolumn{5}{c}{8 Input Views} \\
 \multicolumn{1}{c}{\multirow{7}{*}{\rotatebox{90}{Splatting}}}
&\text{4D-GS}~\cite{yang4DGS}&120+&10h            &\cellcolor[RGB]{\colorthird}26.20 & \cellcolor[RGB]{\colorthird}26.99 & \cellcolor[RGB]{\colorthird}26.34 & \cellcolor[RGB]{\colorthird}27.41 & 24.07 \\
&\text{Deformable3DGS}~\cite{yang2023deformable3dgs}&120+/30&8h   &26.06 & 26.77 & 26.24 & 25.61 & \cellcolor[RGB]{\colorthird}25.62 \\
&\text{4DGaussians}~\cite{wu4dgaussiansRealTime}&120+/50&2h      &\cellcolor[RGB]{\colorsecond}28.16 & \cellcolor[RGB]{\colorsecond}27.34 & \cellcolor[RGB]{\colorsecond}28.10 & \cellcolor[RGB]{\colorsecond}29.60 & \cellcolor[RGB]{\colorsecond}27.62 \\\noalign{\vskip 0.2em}
\cdashline{2-9} \noalign{\vskip 0.2em}

& \text{SplatFields4D (30k it)} &120+/30&2h&29.46 &    29.38 & 28.84 & 29.80 & 29.83 \\
& \text{SplatFields4D (40k it)} &120+/30&3h&29.53 &    29.46 & 28.90 & 29.85 & 29.90 \\
&\text{SplatFields4D (100k it)}&120+/30&7h &29.66 &  29.66 & 29.01 & 29.96 & 30.02 \\
&\text{SplatFields4D (200k it)}&120+/30&14h   &\cellcolor[RGB]{\colorfirst}29.84 & \cellcolor[RGB]{\colorfirst}29.92 & \cellcolor[RGB]{\colorfirst}29.16 & \cellcolor[RGB]{\colorfirst}30.10 & \cellcolor[RGB]{\colorfirst}30.18 \\
\midrule

&& \multicolumn{2}{c}{} &\multicolumn{5}{c}{6 Input Views} \\
  \multicolumn{1}{c}{\multirow{7}{*}{\rotatebox{90}{Splatting}}}
&\text{4D-GS}~\cite{yang4DGS}&120+&10h            &21.42 & 22.89 & 20.80 & 21.60 & 20.40 \\
&\text{Deformable3DGS}~\cite{yang2023deformable3dgs}&120+/30&8h   &\cellcolor[RGB]{\colorthird}24.46 & \cellcolor[RGB]{\colorthird}25.37 & \cellcolor[RGB]{\colorthird}24.31 & \cellcolor[RGB]{\colorthird}24.12 & \cellcolor[RGB]{\colorthird}24.02 \\
&\text{4DGaussians}~\cite{wu4dgaussiansRealTime}&120+/50&2h      &\cellcolor[RGB]{\colorsecond}26.52 & \cellcolor[RGB]{\colorsecond}26.13 & \cellcolor[RGB]{\colorsecond}26.27 & \cellcolor[RGB]{\colorsecond}27.34 & \cellcolor[RGB]{\colorsecond}26.36 \\\noalign{\vskip 0.2em}
\cdashline{2-9} \noalign{\vskip 0.2em}
& \text{SplatFields4D (30k it)} &120+/30&2h&28.04 &    28.36 & 27.31 & 28.44 & 28.07 \\
& \text{SplatFields4D (40k it)} &120+/30&3h&28.10 &    28.43 & 27.35 & 28.48 & 28.13 \\
& \text{SplatFields4D (100k it)} &120+/30&7h&28.22 &  28.61 & 27.44 & 28.56 & 28.25 \\
&\text{SplatFields4D (200k it)} &120+/30&14h   &\cellcolor[RGB]{\colorfirst}28.36 & \cellcolor[RGB]{\colorfirst}28.84 & \cellcolor[RGB]{\colorfirst}27.54 & \cellcolor[RGB]{\colorfirst}28.67 & \cellcolor[RGB]{\colorfirst}28.39 \\
\midrule

&& \multicolumn{2}{c}{} &\multicolumn{5}{c}{4 Input Views} \\
 \multicolumn{1}{c}{\multirow{7}{*}{\rotatebox{90}{Splatting}}}
&\text{4D-GS}~\cite{yang4DGS}&120+&10h            &17.40 & 17.70 & 16.86 & 18.35 & 16.71 \\
&\text{Deformable3DGS}~\cite{yang2023deformable3dgs}&120+/30&8h   &\cellcolor[RGB]{\colorthird}20.04 & \cellcolor[RGB]{\colorthird}21.42 & \cellcolor[RGB]{\colorthird}19.56 & \cellcolor[RGB]{\colorthird}19.71 & \cellcolor[RGB]{\colorthird}19.45 \\
&\text{4DGaussians}~\cite{wu4dgaussiansRealTime}&120+/50&2h      &\cellcolor[RGB]{\colorsecond}21.31 & \cellcolor[RGB]{\colorsecond}21.49 & \cellcolor[RGB]{\colorfirst}21.05 & \cellcolor[RGB]{\colorfirst}21.90 & \cellcolor[RGB]{\colorsecond}20.80 \\\noalign{\vskip 0.2em}
\cdashline{2-9} \noalign{\vskip 0.2em}
&\text{SplatFields4D (30k it)} &120+/30&2h&21.88 &    22.60 & 20.73 & 21.83 & 22.34 \\
&\text{SplatFields4D (40k it)} &120+/30&3h&21.89 &    22.63 & 20.74 & 21.83 & 22.35 \\
&\text{SplatFields4D (100k it)} & 120+/30 & 7h &21.92 &  22.73 & 20.75 & 21.82 & 22.36 \\
&\text{SplatFields4D (200k it)}&120+/30&14h   &\cellcolor[RGB]{\colorfirst}21.95 & \cellcolor[RGB]{\colorfirst}22.83 & \cellcolor[RGB]{\colorsecond}20.76 & \cellcolor[RGB]{\colorsecond}21.83 & \cellcolor[RGB]{\colorfirst}22.39 \\
\bottomrule
\end{tabular}
}
\end{table}

\begin{table}
  \caption{
    \textbf{Flow model ablation study}. Multi-view reconstruction task from Tab.~\ref{tab:exp_dyn_owlii}. See Sec.~\ref{subsec:exp_dyn} for discussion. 
    Symbol ``-'' denotes failed runs}
  \label{tab:exp_dyn_owlii_ablation}
  \centering
  \setlength{\tabcolsep}{3.2pt} %

\scriptsize
\resizebox{\textwidth}{!}{%
\begin{tabular}{@{}l|cccccccc@{}}

\toprule

& \multicolumn{2}{c}{10  Views}   & \multicolumn{2}{c}{8  Views}   & \multicolumn{2}{c}{6  Views}  & \multicolumn{2}{c}{4  Views} \\
& SSIM$\uparrow$ & PSNR$\uparrow$   & SSIM$\uparrow$ & PSNR$\uparrow$   & SSIM$\uparrow$ & PSNR$\uparrow$  & SSIM$\uparrow$ & PSNR$\uparrow$ \\ \midrule

DCT~\cite{DCTNeRF}      & 95.10 & 28.72& 94.53 & 27.34& 94.00 & 26.42& 90.94 & 21.68\\
DCT+ResFields~\cite{ResFields}           & 96.56 & 30.99& 96.05 & 29.66& 95.46 & 28.22& \cellcolor[RGB]{\colorthird}91.47 & \cellcolor[RGB]{\colorsecond}22.12\\
offset~\cite{wu4dgaussiansRealTime} & 95.20 & 28.91& 94.74 & 27.83& 94.15 & 26.73& 90.85 & 21.63\\
offset+ResFields~\cite{ResFields}        & \cellcolor[RGB]{\colorsecond}96.75 &\cellcolor[RGB]{\colorthird}31.24& \cellcolor[RGB]{\colorfirst}96.28 & \cellcolor[RGB]{\colorsecond}29.83& \cellcolor[RGB]{\colorsecond}95.68 & \cellcolor[RGB]{\colorsecond}28.38& 91.41 & 21.62\\
SE3~\cite{yang2023deformable3dgs} & 95.33 & 29.05& 94.78 & 27.95& 94.39 & 27.09& 91.13 & \cellcolor[RGB]{\colorthird}22.09\\
SE3+ResFields~\cite{ResFields}           & \cellcolor[RGB]{\colorfirst}96.81 & \cellcolor[RGB]{\colorfirst}31.32& \cellcolor[RGB]{\colorfirst}96.28 & \cellcolor[RGB]{\colorfirst}29.84& \cellcolor[RGB]{\colorfirst}95.69 & \cellcolor[RGB]{\colorthird}28.36& \cellcolor[RGB]{\colorsecond}91.60 & 21.95\\

scaled SE3~\cite{zhang2024degrees} & 95.05 & 28.78 &92.42 & 23.88 & - & - &- & - \\
scaled SE3+ResFields~\cite{ResFields} & \cellcolor[RGB]{\colorthird}96.74 &\cellcolor[RGB]{\colorsecond}31.30 &\cellcolor[RGB]{\colorthird}96.08 &\cellcolor[RGB]{\colorsecond}29.83 &\cellcolor[RGB]{\colorthird}95.63 &\cellcolor[RGB]{\colorfirst}28.64 &\cellcolor[RGB]{\colorfirst}92.54 &\cellcolor[RGB]{\colorfirst}23.55 \\

\bottomrule
\end{tabular}
}
\end{table}

\textbf{Multi-view dynamic reconstruction.} 
Following~\cite{ResFields}, we use 4 sequences from Owlii~\cite{xu2017owlii}. We opt for the dataset as it has realistic and more complex motion compared to the commonly utilized synthetic sequences~\cite{DNeRF}. 
Each sequence is 100 frames long and comprises multi-view video streams, where we vary the number of input views from 4 to 10 to study the robustness of our and baseline models. For evaluation, we use 100 images of a rotating camera around the performer, where each image comes from a different time step. 

We provide a comprehensive comparison of the SplatFields4D against the respective baselines in Tab.~\ref{tab:exp_dyn_owlii}. Dynamic NeRF methods~\cite{DyNeRF,DNeRF,Nerfies,HyperNeRF} improved with ResFields~\cite{ResFields} generally require significantly longer training times, while recent dynamic 3DGS-based methods~\cite{yang4DGS,yang2023deformable3dgs,wu4dgaussiansRealTime} showcase suboptimal modeling capabilities and performance in the considered sparse scenario. Our method demonstrates a clear metric improvement over the baselines while retaining the key properties of 3DGS, such as interactive rendering speed and compatibility with the existing visualization pipelines. We specifically emphasize the difference in performance between our and the recent closely related method~\cite{yang4DGS}, where our combination of the triplane features and MLP-based dynamics modeling proves to be more robust compared to the HexPlane-based~\cite{HexPlane} approach in the case of rapid motion and sparse camera setups.

\textbf{Splat flow model ablation.} We compare the flow field $f_{\mathbf{p}}$ modeling deformations via DCT basis~\cite{li2023dynibar}, translation vectors~\cite{DNeRF}, SE(3) transformation~\cite{Nerfies}, and via scaled SE(3) transformation~\cite{zhang2024degrees}. We also ablate the impact of implementing neural fields via ResFields \cite{ResFields} to further increase the modeling capacity of our pipeline without affecting its training speed. 
The results (Tab.~\ref{tab:exp_dyn_owlii_ablation}) suggest that modeling flow as SE(3) achieves slightly better quality when the number of views is large. %
We further observe that implementing neural fields via ResFields further boosts the reconstruction quality across all setups.

\section{Conclusion} 
In this work, we proposed an effective optimization strategy that introduces spatial and connectivity biases into the 3D Gaussian splats during optimization process by modeling them through a continuous neural field. We demonstrated that our optimization strategy considerably enhances reconstruction quality in the sparse setups, without the need for any external, data-driven priors. Furthermore, we introduced an effective extension of our method for reconstructing dynamic sequences and demonstrated state-of-the-art results under sparse views.

\textbf{Limitations and future work.} The performance of our method noticeably diminishes in extremely sparse and highly dynamic scenarios, such as those involving rapid motion with as few as four views, exemplified by the Owlii dataset~\cite{xu2017owlii}. This performance is inferior when compared to the best-performing NeRF-based methods in similar sparse configurations~\cite{ResFields}. Therefore, further exploration is required to narrow the performance gap between 3DGS- and NeRF-based methods in sparse settings. Future work should also consider incorporating learning-based priors~\cite{jain2021putting,yang2024gaussianobject,yu2021pixelnerf,ssdnerf} as promising directions for advancement.

\clearpage
\appendix
\setcounter{page}{1}
\setcounter{table}{0}
\setcounter{figure}{0}
\counterwithin{figure}{section}
\counterwithin{table}{section}
\renewcommand{\thetable}{\thesection.\arabic{table}}
\renewcommand{\thefigure}{\thesection.\arabic{figure}}
\renewcommand{\theequation}{\thesection.\arabic{equation}}

\title{SplatFields: Neural Gaussian Splats for\\Sparse 3D and 4D Reconstruction \newline --Supplementary Material--}
\titlerunning{SplatFields}
\authorrunning{M. Mihajlovic et al.}

\author{}
\institute{}

\maketitle

\subsubsection{Overview.} 
We provide additional details related to training (Sec.~\ref{app_subsec:training}), implementation (Sec.~\ref{app_subsec:impl}), and reported results (Sec.~\ref{app_subsec:results}). 
We refer the reader to the supplementary video for more qualitative results. 

\section{Training} \label{app_subsec:training} 
\subsubsection{Optimization.}
Given a set of calibrated multi-view input images and an initial collection of $K$ Gaussian splats $\mathbf{G} = \{\mathcal{G}_k\}_{k=1}^K$, differential rasterizer $\mathcal{R}$ based on Gaussian Splatting (Sec.~\ref{sec:preliminaries}) propagates image changes to the scene parameters $\mathbf{G}$. 
This feedback loop is used to optimize the scene parameters by imposing the photometric loss between the rendered $I$ and the input image $I^*$:
\begin{equation} \label{eq:optimization}
    \arg \min_{\mathbf{G}} \mathcal{L} (\mathcal{R}(\mathbf{G}), I^*),
\end{equation}
where the initial collection of splats is initialized either randomly~\cite{lassner2021pulsar}, by SfM \cite{schoenberger2016sfm}, or by the visual hull~\cite{franco2003exact}. 

\subsubsection{Loss definition.} We follow the training scheme used by 3DGS (Sec.~\ref{sec:preliminaries}) and optimize the rendering objective (Eq.~\ref{eq:optimization}) via the Adam optimizer \cite{Adam:ICLR:2015}. 
We also employ the mask loss between rendered and ground-truth masks for the object-level scenes:
\begin{equation} \label{eq:gaussian_loss_mask}
    \mathcal{L} = (1-\lambda_1)\mathcal{L}_1 + \lambda_1 \mathcal{L}_{\text{D-SSIM}} + \lambda_2 \mathcal{L}_{\text{MASK}} + \lambda_3 \mathcal{L}_{\text{norm}},
\end{equation}
where $\mathcal{L}_{\text{MASK}}$ is the $\mathcal{L}_1$ loss between the rendered opacity and the ground truth mask akin to \cite{qian20233dgs} and $\mathcal{L}_{\text{norm}}$ is the splat norm regularization term from Sec.~\ref{sec:splatFields}. 

Hyperparameter $\lambda_1$ is empirically set to $0.2$, $\lambda_2$ is set to $0.1$ for object-level scenes~\cite{mildenhall2020nerf,xu2017owlii} and to $0$ for unbounded scenes, while $\lambda_3$ is set to $0.01$ for static reconstruction and to $0$ for all of the other experiments. 
We further employ the exponential learning rate decay that starts from $8 \times 10^{-4}$ until it reaches $1.6 \times 10^{-6}$ at 40k iterations.

\section{Implementation Details} \label{app_subsec:impl} 

\subsection{Spatial Autocorrelation} \label{app_subsec:morans} 
To quantify the spatial similarity of nearby features, we measure local spatial autocorrelation via Moran's I \cite{moran1950notes} between the features of splats in their local neighborhoods. 
Specifically, for each splat $\mathcal{G}_k$ and its attribute (color, opacity, and covariance) we query $N$ nearest neighbors $[\mathcal{X}_{i}]_{i=1}^{N}$ with associated locations $loc(\mathcal{X}_{i}) \in \mathbb{R}^{3}$ and measure Moran's I (Eq.~\ref{eq:moransI}) of its attributes $attr(\mathcal{X}_i) \in \mathbb{R}$: 

\begin{equation}\label{eq:moransI}
    I = \E_{\mathcal{X} \in \mathbf{G}}[I(\mathcal{X})],
\end{equation}
where 
\begin{equation}
    I(\mathcal{X}) = 
    \frac{N}{\sum_{i=1}^{N} \sum_{j=1}^{N} W_{ij}(\mathcal{X})}
    \frac{\sum_{i=1}^N \sum_{j=1}^N W_{ij}(\mathcal{X}) attr(\mathcal{X}_i) attr(\mathcal{X}_j)} {\sum_{i=1}^N attr(\mathcal{X}_i)^2},
\end{equation}
\begin{equation}
W_{ij}(\mathcal{X}) = \begin{cases*} 
    \| \textit{loc}(X_i) - \textit{loc}(X_j) \|^{-1}_{2} & if $i \neq j$ \,, \\
    0 & otherwise.
 \end{cases*}
\end{equation}

For attributes with more than one feature dimension (\eg color and covariance matrix), we average Moran's I across all feature dimensions. In all of the experiments, we set $N=5$. 

\textbf{Moran's Loss $\mathcal{L}_{Moran}$} (in Tab.~\ref{tab:exp_static_blender},~\ref{tab:app:exp_static_blender},~\ref{tab_app:exp_static_blender}) enhances 3DGS as a straightforward baseline that incorporates the \textit{spatial bias} by enforcing a higher Moran’s I score  and is implemented as the negative autocorrelation score: 
\begin{equation}
    \mathcal{L}_{Moran} = \lambda_{Moran}\left(1 - \E_{\mathcal{X} \in \mathbf{G}}[I(\mathcal{X})]\right)\,,
\end{equation}
where $\lambda_{Moran}$ is empirically set to $0.01$. 

\subsection{SplatFields} 
In the following, we describe the network architectures. 

\textit{CNN Generator $g_\theta$} consists of three CNN decoders to produce three axis-aligned feature planes $\mathbf{F}$. 
Each decoder, takes as input a $20\times20$-resolution noise $\mathbf{\epsilon} \in \mathbb{R}^{20 \times 20 \times 8}$ with $8$ channels to produce the $160\times160$-resolution feature plane with 16 channels ($\mathbb{R}^{160 \times 160 \times 16}$) through up-sampling blocks with residual connections. 
First, the noise is expanded to 32 channels via an up-sampling CNN layer, which is then processed by a single attention layer and propagated through a ResNet block with two CNN layers to output an intermediate feature ($20 \times 20 \times 32$). 
This feature is then propagated through four up-sampling blocks until the feature resolution of $160 \times 160 \times 32$ which is then down-scaled to 16 channels via a single CNN layer to form the final tri-plane representation $\mathbf{F} \in \mathbb{R}^{3 \times 160 \times 160 \times 16}$. 
Each up-sampling block consists of two CNN ResNet blocks (each with two CNN layers) and one up-sampling CNN layer. 
Then the splat center $\mathbf{p}_k$ is projected onto each axis-aligned feature plane to obtain feature vectors via bi-linear interpolation. 
These features are then concatenated along the feature dimension and propagated through a tiny 2-layer MLP with 48 neurons to produce the point feature $\mathbf{f}_k \in \mathbb{R}^{48}$.

\textit{Deform MLP $f_{\Theta}$} takes as input the splat center $\mathbf{p}_k$ and the feature $\mathbf{f}_k$. 
The splat location is first positionally encoded~\cite{mildenhall2020nerf} with 4 levels and propagated through an 8-layer MLP with 128 neurons that deforms the splat center  $\hat{\mathbf{p}}_k$  by predicting its residual akin to \cite{prokudin2023dynamic,DNeRF}. 

\textit{Color Field $f_{\Theta_{\mathbf{c}}}$} takes as input the deformed query point (positionally encoded with 4 levels) along with $\mathbf{f}_k$ and propagates them through a 6-layer MLP with 128 neurons, where the last layer takes as input the viewing direction akin to NeRF~\cite{mildenhall2020nerf}. 

\textit{Scale $f_{\Theta_{\mathbf{s}}}$ and Opacity $f_{\Theta_{\alpha}}$ Fields } take the same input as the color MLP and are implemented as 5-layer 64-neuron MLPs. 
The output of the opacity MLP is activated by the sigmoid function. 

\textit{Rotation Field $f_{\Theta_{\mathbf{O}}}$} is implemented as a 4-layer MLP that takes the same input as the color MLP and predicts a four-dimensional vector that is normalized to produce the quaternion representation. 

\textit{Flow Field $f_{\Theta_{\mathbf{p}}}$} is utilized only for the 4D reconstruction. It takes as input the deformed splat center $\hat{\mathbf{p}}_k$ (positionally encoded with 4 levels) and the feature vector $\mathbf{f}_k$ and propagates them through an architecture similar to Deform MLP to model the forward flow. In the paper, we consider different types of modeling the flow: DCT~\cite{li2023dynibar}, SE(3)~\cite{Nerfies}, scaled SE(3)~\cite{zhang2024degrees}, and offsets~\cite{prokudin2023dynamic,DNeRF}. See Sec.~\ref{subsec:exp_dyn} for further details. 

All of the MLP fields take time (positionally encoded with 4 levels) as an additional input and are implemented as ResField MLPs~\cite{ResFields}. 
We empirically set the ResFields' rank to 40 for the multi-view dynamic reconstruction on Owlii~\cite{xu2017owlii} and to 0 for the monocular reconstruction~\cite{nerfds} as the scenes are semi-static. 

\section{Experiment Details} \label{app_subsec:results} 

\begin{table}
  \caption{
    \textbf{Sparse static scene reconstruction} of Blender~\cite{mildenhall2020nerf} scenes. 
    Reported numbers indicate PSNR metric on the novel views (``-'' denotes failed runs). Colors denote the \colorbox[RGB]{\colorfirst}{1st}, \colorbox[RGB]{\colorsecond}{2nd}, and \colorbox[RGB]{\colorthird}{3rd} best-performing model. See Sec.~\ref{subsec:exp_static} for discussion
    }
  \label{tab:app:exp_static_blender}
  \centering
  \scriptsize
  \setlength{\tabcolsep}{3.9pt} %
\resizebox{\textwidth}{!}{%
\begin{tabular}{@{}l|ccccccccc@{}}
\toprule
&\multicolumn{9}{c}{12 Input Views} \\
  & \textit{mean} & Toy &       Ficus & Hotdog &        Chair & Mic &   Ship &  Drums & Materials \\
SparseNeRF~\cite{sparsenerf}& - &    23.02 & 18.19 & - & 26.20 & 23.26 & 20.81 & 19.21 & 20.80 \\
SparseNeRF \textit{wo.} depth & 22.92 &    24.00 & 18.84 & 27.52 & 27.11 & 23.35 & 21.84 & 19.17 & 21.50 \\ 
\text{SuGaR}~\cite{sugar} &21.78 &   23.77 & 23.08 & 22.36 & 25.72 & 18.72 & 21.09 & 19.55 & 19.94 \\
\text{ScaffoldGS}~\cite{scaffoldgs} &23.82 &      23.65 & 22.78 & 26.34 & 25.80 & \cellcolor[RGB]{\colorthird}28.28 & 21.17 & 20.47 & 22.06 \\
\text{Mip3DGS}~\cite{mipgs} &24.86 & 24.65 & 25.62 & 26.53 & 26.25 & \cellcolor[RGB]{\colorsecond}28.40 & 22.52 & 21.98 & \cellcolor[RGB]{\colorfirst}22.94 \\
\text{3DGS}~\cite{GaussianSplatting} &25.29 &    25.14 & 25.92 & 27.51 & 27.10 & \cellcolor[RGB]{\colorfirst}29.02 & 22.79 & \cellcolor[RGB]{\colorthird}22.10 & 22.71 \\
\text{Light3DGS}~\cite{lightgaussian} &25.39 & 25.08 & \cellcolor[RGB]{\colorfirst}27.53 & 27.10 & 27.40 & 28.04 & \cellcolor[RGB]{\colorthird}23.02 & 22.07 & \cellcolor[RGB]{\colorsecond}22.90 \\ 
\text{2DGS}~\cite{2dgs} &\cellcolor[RGB]{\colorsecond}25.62 &    \cellcolor[RGB]{\colorsecond}25.50 & 25.62 & \cellcolor[RGB]{\colorsecond}29.24 & \cellcolor[RGB]{\colorfirst}28.52 & 28.07 & \cellcolor[RGB]{\colorsecond}23.08 & \cellcolor[RGB]{\colorsecond}22.19 & \cellcolor[RGB]{\colorthird}22.75 \\
\noalign{\vskip 0.2em}\cdashline{1-10}\noalign{\vskip 0.2em}
\text{3DGS \textit{w.} $\mathcal{L}_{Moran}$} &\cellcolor[RGB]{\colorthird}25.44 &  \cellcolor[RGB]{\colorthird}25.26 & \cellcolor[RGB]{\colorsecond}26.55 & \cellcolor[RGB]{\colorthird}28.96 & \cellcolor[RGB]{\colorsecond}27.91 & 27.87 & 22.33 & 21.98 & 22.65 \\
\text{SplatFields3D} &\cellcolor[RGB]{\colorfirst}25.80 &    \cellcolor[RGB]{\colorfirst}26.98 & \cellcolor[RGB]{\colorthird}26.27 & \cellcolor[RGB]{\colorfirst}29.45 & \cellcolor[RGB]{\colorthird}27.42 & 27.60 & \cellcolor[RGB]{\colorfirst}23.78 & \cellcolor[RGB]{\colorfirst}22.55 & 22.32 \\
\midrule
&\multicolumn{9}{c}{10 Input Views}\\

SparseNeRF~\cite{sparsenerf}                   & -     & 22.64 & 18.27 & -     & 25.30 & 23.27 & 20.29 & 18.61 & 19.72 \\
SparseNeRF \textit{wo.} depth                   & 22.58 & 23.89 & 18.75 & 27.56 & \cellcolor[RGB]{\colorsecond}26.42 & 23.23 & 21.68 & 18.20 & \cellcolor[RGB]{\colorsecond}20.87 \\
\text{SuGaR}~\cite{sugar}                      & 21.10 & 22.78 & 22.42 & 23.60 & 24.25 & 17.93 & 20.35 & 19.11 & 18.40 \\
\text{ScaffoldGS}~\cite{scaffoldgs}            & 22.63 & 21.98 & 22.68 & 24.37 & 24.15 & \cellcolor[RGB]{\colorsecond}27.76 & 20.39 & 19.64 & 20.08 \\
\text{Mip3DGS}~\cite{mipgs}                    & 23.65 & 23.49 & 24.97 & 25.27 & 24.49 & \cellcolor[RGB]{\colorthird}27.69 & 21.38 & 21.23 & \cellcolor[RGB]{\colorthird}20.66 \\
\text{3DGS}~\cite{GaussianSplatting}           & 24.11 & 23.79 & 25.54 & 26.16 & 25.28 & \cellcolor[RGB]{\colorfirst}28.39 & \cellcolor[RGB]{\colorsecond}21.87 & 21.34 & 20.51 \\
\text{Light3DGS}~\cite{lightgaussian}          & \cellcolor[RGB]{\colorthird}24.21 & \cellcolor[RGB]{\colorsecond}23.94 & \cellcolor[RGB]{\colorfirst}26.95 & 25.62 & 25.91 & 27.45 & 21.82 & \cellcolor[RGB]{\colorthird}21.38 & 20.60 \\
\text{2DGS}~\cite{2dgs}                        & \cellcolor[RGB]{\colorsecond}24.42 & 24.06 & 25.17 & \cellcolor[RGB]{\colorsecond}27.92 & \cellcolor[RGB]{\colorfirst}26.96 & 27.53 & \cellcolor[RGB]{\colorthird}21.83 & \cellcolor[RGB]{\colorsecond}21.58 & 20.27 \\
\noalign{\vskip 0.2em}\cdashline{1-10}\noalign{\vskip 0.2em}
\text{3DGS \textit{w.}$\mathcal{L}_{Moran}$}  & \cellcolor[RGB]{\colorthird}24.21 & \cellcolor[RGB]{\colorthird}23.91 & \cellcolor[RGB]{\colorsecond}26.09 & \cellcolor[RGB]{\colorthird}27.65 & 25.86 & 27.07 & 21.38 & 21.26 & 20.46 \\
\text{SplatFields3D}                           & \cellcolor[RGB]{\colorfirst}24.94 & \cellcolor[RGB]{\colorfirst}26.51 & \cellcolor[RGB]{\colorthird}25.59 & \cellcolor[RGB]{\colorfirst}28.29 & \cellcolor[RGB]{\colorthird}25.92 & 27.36 & \cellcolor[RGB]{\colorfirst}23.12 & \cellcolor[RGB]{\colorfirst}21.86 & \cellcolor[RGB]{\colorfirst}20.88 \\
\midrule
&\multicolumn{9}{c}{8 Input Views} \\

SparseNeRF~\cite{sparsenerf}                    & -     & 22.33 & 17.97 & -     & 23.81 & 23.01 & 19.85 & 17.85 & \cellcolor[RGB]{\colorsecond}20.02 \\
SparseNeRF \textit{wo.} depth                   & 22.20 & \cellcolor[RGB]{\colorsecond}24.06 & 18.42 & \cellcolor[RGB]{\colorsecond}27.09 & 25.12 & 23.04 & \cellcolor[RGB]{\colorsecond}21.23 & 17.94 & \cellcolor[RGB]{\colorfirst}20.74 \\
\text{SuGaR}~\cite{sugar}                       & 20.62 & 21.91 & 22.33 & 23.01 & 23.30 & 18.60 & 19.59 & 18.66 & 17.55 \\
\text{ScaffoldGS}~\cite{scaffoldgs}             & 21.53 & 20.95 & 21.35 & 23.77 & 22.77 & 26.40 & 18.88 & 18.96 & 19.17 \\
\text{Mip3DGS}~\cite{mipgs}                     & 22.37 & 22.05 & 23.23 & 24.24 & 23.57 & 26.32 & 19.91 & 20.10 & 19.55 \\
\text{3DGS}~\cite{GaussianSplatting}            & 22.93 & 22.55 & 23.69 & 25.57 & 24.43 & \cellcolor[RGB]{\colorfirst}27.37 & 19.98 & 20.33 & 19.49 \\
\text{Light3DGS}~\cite{lightgaussian}           & 22.98 & 22.67 & \cellcolor[RGB]{\colorfirst}24.98 & 24.79 & 24.40 & \cellcolor[RGB]{\colorthird}26.59 & \cellcolor[RGB]{\colorthird}20.60 & \cellcolor[RGB]{\colorthird}20.41 & 19.41 \\
\text{2DGS}~\cite{2dgs}                         & \cellcolor[RGB]{\colorthird}23.04 & 22.19 & 23.63 & \cellcolor[RGB]{\colorthird}26.76 & \cellcolor[RGB]{\colorsecond}25.46 & 26.24 & 20.16 & \cellcolor[RGB]{\colorsecond}20.60 & 19.25 \\
\noalign{\vskip 0.2em}\cdashline{1-10}\noalign{\vskip 0.2em}
\text{3DGS \textit{w.} $\mathcal{L}_{Moran}$}   & \cellcolor[RGB]{\colorsecond}23.19 & \cellcolor[RGB]{\colorthird}22.79 & \cellcolor[RGB]{\colorsecond}24.56 & 26.57 & \cellcolor[RGB]{\colorthird}25.14 & \cellcolor[RGB]{\colorsecond}26.97 & 19.79 & \cellcolor[RGB]{\colorthird}20.41 & 19.32 \\
\text{SplatFields3D}                            & \cellcolor[RGB]{\colorfirst}23.98 & \cellcolor[RGB]{\colorfirst}24.71 & \cellcolor[RGB]{\colorthird}23.97 & \cellcolor[RGB]{\colorfirst}27.87 & \cellcolor[RGB]{\colorfirst}25.64 & 26.49 & \cellcolor[RGB]{\colorfirst}22.15 & \cellcolor[RGB]{\colorfirst}21.12 & \cellcolor[RGB]{\colorthird}19.85 \\

\midrule
&\multicolumn{9}{c}{6 Input Views} \\
SparseNeRF~\cite{sparsenerf}                    & -     & \cellcolor[RGB]{\colorthird}20.86 & 18.03 & -     & \cellcolor[RGB]{\colorthird}22.75 & 22.40 & \cellcolor[RGB]{\colorsecond}19.33 & 16.24 & \cellcolor[RGB]{\colorsecond}19.54 \\
SparseNeRF \textit{wo.} depth                   & \cellcolor[RGB]{\colorthird}20.86 & \cellcolor[RGB]{\colorfirst}22.62 & 17.63 & \cellcolor[RGB]{\colorsecond}25.84 & 22.65 & 20.72 & \cellcolor[RGB]{\colorfirst}19.85 & 17.25 & \cellcolor[RGB]{\colorfirst}20.30 \\
\text{SuGaR}~\cite{sugar}                       & 19.07 & 19.89 & 20.61 & 20.80 & 21.92 & 18.26 & 17.72 & 16.86 & 16.53 \\
\text{ScaffoldGS}~\cite{scaffoldgs}             & 19.65 & 18.21 & 20.72 & 19.48 & 22.20 & 24.31 & 16.47 & 17.21 & 18.62 \\
\text{Mip3DGS}~\cite{mipgs}                     & 20.04 & 19.39 & 21.81 & 19.70 & 21.72 & 24.44 & 17.02 & 17.72 & 18.52 \\
\text{3DGS}~\cite{GaussianSplatting}            & 20.62 & 19.80 & 22.25 & 21.16 & \cellcolor[RGB]{\colorthird}22.75 & \cellcolor[RGB]{\colorfirst}25.21 & 17.58 & 17.77 & 18.48 \\
\text{Light3DGS}~\cite{lightgaussian}           & 20.76 & 20.25 & \cellcolor[RGB]{\colorfirst}23.12 & 20.66 & 22.69 & \cellcolor[RGB]{\colorsecond}24.89 & 17.83 & \cellcolor[RGB]{\colorthird}18.02 & 18.63 \\
\text{2DGS}~\cite{2dgs}                         & 20.74 & 19.38 & 21.93 & 23.85 & \cellcolor[RGB]{\colorsecond}23.26 & 24.48 & 16.92 & 17.91 & 18.17 \\
\noalign{\vskip 0.2em}\cdashline{1-10}\noalign{\vskip 0.2em}
\text{3DGS \textit{w.} $\mathcal{L}_{Moran}$}   & \cellcolor[RGB]{\colorsecond}21.03 & 20.34 & \cellcolor[RGB]{\colorsecond}23.05 & \cellcolor[RGB]{\colorthird}23.92 & 22.50 & 24.64 & 17.20 & \cellcolor[RGB]{\colorsecond}18.14 & 18.48 \\
\text{SplatFields3D}                            & \cellcolor[RGB]{\colorfirst}22.26 & \cellcolor[RGB]{\colorsecond}22.41 & \cellcolor[RGB]{\colorthird}22.26 & \cellcolor[RGB]{\colorfirst}26.19 & \cellcolor[RGB]{\colorfirst}25.03 & \cellcolor[RGB]{\colorthird}24.84 & \cellcolor[RGB]{\colorsecond}19.33 & \cellcolor[RGB]{\colorfirst}18.97 & \cellcolor[RGB]{\colorthird}19.05 \\
\midrule
&\multicolumn{9}{c}{4 Input Views} \\
SparseNeRF~\cite{sparsenerf}                        & -     & \cellcolor[RGB]{\colorfirst}20.94 & 17.48 & \cellcolor[RGB]{\colorsecond}23.81 & \cellcolor[RGB]{\colorfirst}21.41 & 21.52 & -     & \cellcolor[RGB]{\colorthird}15.37 & \cellcolor[RGB]{\colorsecond}17.03 \\
SparseNeRF \textit{wo.} depth                       & \cellcolor[RGB]{\colorthird}17.87 & \cellcolor[RGB]{\colorsecond}19.31 & 17.05 & \cellcolor[RGB]{\colorthird}23.54 & \cellcolor[RGB]{\colorsecond}20.26 & 11.56 & \cellcolor[RGB]{\colorfirst}17.86 & 13.64 & \cellcolor[RGB]{\colorfirst}19.77 \\
\text{SuGaR}~\cite{sugar}                           & 16.94 & 16.96 & 19.30 & 19.36 & 19.07 & 17.47 & 15.22 & 14.73 & 13.38 \\
\text{ScaffoldGS}~\cite{scaffoldgs}                 & 16.86 & 15.40 & 19.58 & 17.31 & 18.40 & 20.54 & 14.70 & 15.27 & 13.69 \\
\text{Mip3DGS}~\cite{mipgs}                         & 16.94 & 16.23 & 19.60 & 16.98 & 18.38 & 20.56 & 14.64 & 14.92 & 14.21 \\
\text{3DGS}~\cite{GaussianSplatting}                & 17.37 & 16.44 & 19.72 & 18.65 & 18.72 & 20.75 & 15.43 & 15.08 & 14.15 \\
\text{Light3DGS}~\cite{lightgaussian}               & 17.70 & 16.94 & \cellcolor[RGB]{\colorsecond}20.35 & 18.56 & 18.96 & \cellcolor[RGB]{\colorthird}21.53 & \cellcolor[RGB]{\colorthird}15.67 & \cellcolor[RGB]{\colorsecond}15.44 & 14.19 \\ 
\text{2DGS}~\cite{2dgs}                             & 17.58 & 16.32 & 19.69 & 20.67 & \cellcolor[RGB]{\colorthird}19.39 & 21.17 & 14.45 & 14.84 & 14.14 \\
\noalign{\vskip 0.2em}\cdashline{1-10}\noalign{\vskip 0.2em}
\text{3DGS \textit{w.} $\mathcal{L}_{Moran}$}       & \cellcolor[RGB]{\colorsecond}18.13 & 17.06 & \cellcolor[RGB]{\colorfirst}20.53 & 22.10 & 18.25 & \cellcolor[RGB]{\colorfirst}22.06 & 15.18 & 15.32 & 14.53 \\
\text{SplatFields3D}                                & \cellcolor[RGB]{\colorfirst}19.16 & \cellcolor[RGB]{\colorthird}18.89 & \cellcolor[RGB]{\colorthird}20.19 & \cellcolor[RGB]{\colorfirst}24.31 & 19.31 & \cellcolor[RGB]{\colorsecond}21.73 & \cellcolor[RGB]{\colorsecond}16.83 & \cellcolor[RGB]{\colorfirst}16.35 & \cellcolor[RGB]{\colorthird}15.69 \\
\bottomrule

\end{tabular}
}
\end{table}

\begin{table}
    \caption{
      \textbf{Sparse static scene reconstruction.} 
      Synthetic Blender~\cite{mildenhall2020nerf} dataset, reported numbers indicate SSIM metric on the novel views. 
      See Sec.~\ref{subsec:exp_static} for discussion
      }
    \label{tab_app:exp_static_blender}
    \centering
    \scriptsize
    \setlength{\tabcolsep}{3.9pt} %

\resizebox{\textwidth}{!}{%
  \begin{tabular}{@{}l|ccccccccc@{}}
  \toprule
  
&\multicolumn{9}{c}{12 Input Views} \\
 & \textit{mean} & Toy &       Ficus & Hotdog &        Chair & Mic &   Ship &  Drums & Materials \\
SparseNeRF~\cite{sparsenerf}                    & -     & 86.07 & 84.57 & -     & 90.45 & 92.23 & 76.28 & 83.65 & 85.40 \\
SparseNeRF \textit{wo.} depth                   & 87.54 & 88.64 & 85.10 & 93.96 & 91.77 & 92.55 & 77.69 & 83.69 & 86.94 \\
\text{SuGaR}~\cite{sugar}                       & 85.60 & 86.24 & 89.09 & 90.29 & 90.94 & 85.86 & \cellcolor[RGB]{\colorthird}76.81 & 82.77 & 82.76 \\
\text{ScaffoldGS}~\cite{scaffoldgs}             & 87.47 & 86.22 & 90.65 & 92.06 & 90.75 & 96.21 & 73.02 & 85.75 & 85.10 \\
\text{Mip3DGS}~\cite{mipgs}                     & 89.78 & 87.81 & 93.87 & 93.15 & 92.46 & \cellcolor[RGB]{\colorsecond}96.90 & 76.42 & 89.67 & 87.96 \\
\text{3DGS}~\cite{GaussianSplatting}            & 90.01 & 88.57 & 94.11 & 93.45 & 93.48 & \cellcolor[RGB]{\colorfirst}96.98 & 76.11 & 89.81 & 87.55 \\
\text{Light3DGS}~\cite{lightgaussian}           & 90.30 & 88.51 & \cellcolor[RGB]{\colorfirst}95.31 & 93.45 & \cellcolor[RGB]{\colorthird}93.52 & 96.64 & 76.57 & \cellcolor[RGB]{\colorthird}89.98 & \cellcolor[RGB]{\colorfirst}88.42 \\
\text{2DGS}~\cite{2dgs}                         & \cellcolor[RGB]{\colorsecond}91.09 & \cellcolor[RGB]{\colorsecond}90.18 & 94.20 & \cellcolor[RGB]{\colorsecond}94.85 & \cellcolor[RGB]{\colorfirst}94.93 & \cellcolor[RGB]{\colorthird}96.77 & \cellcolor[RGB]{\colorsecond}79.09 & \cellcolor[RGB]{\colorsecond}90.41 & \cellcolor[RGB]{\colorsecond}88.27 \\
\noalign{\vskip 0.2em}\cdashline{1-10}\noalign{\vskip 0.2em}
\text{3DGS \textit{w.} $\mathcal{L}_{Moran}$}   & \cellcolor[RGB]{\colorthird}90.45 & \cellcolor[RGB]{\colorthird}89.31 & \cellcolor[RGB]{\colorsecond}94.62 & \cellcolor[RGB]{\colorthird}94.40 & \cellcolor[RGB]{\colorsecond}94.07 & 96.52 & 76.71 & 89.97 & \cellcolor[RGB]{\colorthird}88.01 \\
\text{SplatFields3D}                            & \cellcolor[RGB]{\colorfirst}91.18 & \cellcolor[RGB]{\colorfirst}91.06 & \cellcolor[RGB]{\colorthird}94.36 & \cellcolor[RGB]{\colorfirst}95.55 & 92.42 & 96.17 & \cellcolor[RGB]{\colorfirst}81.03 & \cellcolor[RGB]{\colorfirst}90.90 & 87.98 \\
\midrule
&\multicolumn{9}{c}{10 Input Views} \\
SparseNeRF~\cite{sparsenerf}                    & -     & 85.65 & 84.30 & -     & 89.57 & 92.27 & 75.51 & 82.52 & 83.82 \\
SparseNeRF \textit{wo.} depth                   & 86.95 & \cellcolor[RGB]{\colorsecond}88.45 & 84.65 & \cellcolor[RGB]{\colorsecond}94.06 & 91.26 & 92.39 & \cellcolor[RGB]{\colorsecond}77.25 & 82.12 & \cellcolor[RGB]{\colorfirst}85.42 \\
\text{SuGaR}~\cite{sugar}                       & 83.83 & 84.27 & 88.55 & 90.52 & 88.77 & 84.39 & 74.27 & 80.78 & 79.10 \\
\text{ScaffoldGS}~\cite{scaffoldgs}             & 85.45 & 83.27 & 90.14 & 89.66 & 88.56 & 95.77 & 70.66 & 83.46 & 82.10 \\
\text{Mip3DGS}~\cite{mipgs}                     & 88.00 & 85.68 & 93.15 & 91.76 & 90.34 & \cellcolor[RGB]{\colorsecond}96.44 & 73.98 & 88.22 & 84.41 \\
\text{3DGS}~\cite{GaussianSplatting}            & 88.27 & 86.30 & 93.65 & 92.19 & 91.23 & \cellcolor[RGB]{\colorfirst}96.49 & 73.89 & 88.48 & 83.93 \\
\text{Light3DGS}~\cite{lightgaussian}           & 88.76 & 86.66 & \cellcolor[RGB]{\colorfirst}94.82 & 92.41 & \cellcolor[RGB]{\colorthird}91.70 & 96.27 & 74.37 & \cellcolor[RGB]{\colorthird}88.75 & \cellcolor[RGB]{\colorthird}85.12 \\
\text{2DGS}~\cite{2dgs}                         & \cellcolor[RGB]{\colorsecond}89.51 & \cellcolor[RGB]{\colorthird}88.42 & 93.61 & \cellcolor[RGB]{\colorthird}93.79 & \cellcolor[RGB]{\colorfirst}93.26 & \cellcolor[RGB]{\colorthird}96.37 & \cellcolor[RGB]{\colorthird}76.52 & \cellcolor[RGB]{\colorsecond}89.38 & 84.72 \\
\noalign{\vskip 0.2em}\cdashline{1-10}\noalign{\vskip 0.2em}
\text{3DGS \textit{w.} $\mathcal{L}_{Moran}$}   & \cellcolor[RGB]{\colorthird}88.94 & 87.53 & \cellcolor[RGB]{\colorsecond}94.24 & 93.32 & \cellcolor[RGB]{\colorsecond}91.88 & 95.98 & 75.10 & 88.74 & 84.69 \\
\text{SplatFields3D}                            & \cellcolor[RGB]{\colorfirst}90.32 & \cellcolor[RGB]{\colorfirst}91.34 & \cellcolor[RGB]{\colorthird}93.70 & \cellcolor[RGB]{\colorfirst}95.09 & 91.14 & 95.95 & \cellcolor[RGB]{\colorfirst}80.25 & \cellcolor[RGB]{\colorfirst}89.85 & \cellcolor[RGB]{\colorsecond}85.26 \\
\midrule
&\multicolumn{9}{c}{8 Input Views} \\
SparseNeRF~\cite{sparsenerf}                    & -     & 84.85 & 83.98 & -     & 88.39 & 92.06 & \cellcolor[RGB]{\colorthird}74.32 & 81.44 & \cellcolor[RGB]{\colorsecond}83.66 \\
SparseNeRF \textit{wo.} depth                   & 86.30 & \cellcolor[RGB]{\colorfirst}88.27 & 83.97 & \cellcolor[RGB]{\colorsecond}93.65 & 90.23 & 92.15 & \cellcolor[RGB]{\colorsecond}76.01 & 81.38 & \cellcolor[RGB]{\colorfirst}84.78 \\
\text{SuGaR}~\cite{sugar}                       & 82.74 & 82.04 & 87.95 & 89.41 & 87.60 & 85.08 & 72.80 & 79.84 & 77.23 \\
\text{ScaffoldGS}~\cite{scaffoldgs}             & 83.62 & 80.54 & 88.25 & 89.04 & 86.23 & 95.01 & 68.49 & 81.49 & 79.91 \\
\text{Mip3DGS}~\cite{mipgs}                     & 86.24 & 82.93 & 91.03 & 90.84 & 89.21 & \cellcolor[RGB]{\colorthird}95.77 & 71.86 & 86.15 & 82.14 \\
\text{3DGS}~\cite{GaussianSplatting}            & 86.63 & 84.05 & 91.56 & 91.63 & 90.49 & \cellcolor[RGB]{\colorfirst}95.99 & 70.96 & 86.62 & 81.76 \\
\text{Light3DGS}~\cite{lightgaussian}           & 87.11 & 84.44 & \cellcolor[RGB]{\colorfirst}93.01 & 91.47 & 90.17 & 95.76 & 72.16 & 86.97 & 82.92 \\
\text{2DGS}~\cite{2dgs}                         & \cellcolor[RGB]{\colorsecond}87.72 & 84.80 & \cellcolor[RGB]{\colorthird}91.74 & \cellcolor[RGB]{\colorthird}93.13 & \cellcolor[RGB]{\colorfirst}91.79 & 95.72 & 74.06 & \cellcolor[RGB]{\colorsecond}87.89 & 82.65 \\
\noalign{\vskip 0.2em}\cdashline{1-10}\noalign{\vskip 0.2em}
\text{3DGS \textit{w.} $\mathcal{L}_{Moran}$}   & \cellcolor[RGB]{\colorthird}87.35 & \cellcolor[RGB]{\colorthird}85.15 & \cellcolor[RGB]{\colorsecond}92.57 & 92.50 & \cellcolor[RGB]{\colorsecond}91.04 & \cellcolor[RGB]{\colorsecond}95.88 & 72.03 & \cellcolor[RGB]{\colorthird}87.26 & 82.36 \\
\text{SplatFields3D}                            & \cellcolor[RGB]{\colorfirst}88.94 & \cellcolor[RGB]{\colorsecond}88.04 & 91.69 & \cellcolor[RGB]{\colorfirst}94.68 & \cellcolor[RGB]{\colorthird}90.91 & 95.48 & \cellcolor[RGB]{\colorfirst}78.76 & \cellcolor[RGB]{\colorfirst}88.50 & \cellcolor[RGB]{\colorthird}83.46 \\
\midrule
&\multicolumn{9}{c}{6 Input Views} \\
SparseNeRF~\cite{sparsenerf}                    & -     & \cellcolor[RGB]{\colorthird}83.25 & 83.70 & -     & \cellcolor[RGB]{\colorthird}87.76 & 91.53 & \cellcolor[RGB]{\colorsecond}72.96 & 78.58 & \cellcolor[RGB]{\colorsecond}83.47 \\
SparseNeRF \textit{wo.} depth                   & 84.42 & \cellcolor[RGB]{\colorfirst}85.68 & 82.33 & \cellcolor[RGB]{\colorsecond}92.51 & 87.47 & 90.72 & \cellcolor[RGB]{\colorthird}72.50 & 79.58 & \cellcolor[RGB]{\colorfirst}84.59 \\
\text{SuGaR}~\cite{sugar}                       & 79.85 & 77.67 & 85.51 & 86.44 & 84.98 & 84.24 & 68.93 & 76.35 & 74.68 \\
\text{ScaffoldGS}~\cite{scaffoldgs}             & 80.34 & 74.13 & 87.01 & 82.79 & 84.99 & 93.10 & 62.49 & 78.61 & 79.59 \\
\text{Mip3DGS}~\cite{mipgs}                     & 83.09 & 77.68 & 89.34 & 86.81 & 86.46 & 94.58 & 66.45 & 82.08 & 81.34 \\
\text{3DGS}~\cite{GaussianSplatting}            & 83.56 & 78.58 & \cellcolor[RGB]{\colorthird}89.79 & 87.81 & 87.35 & \cellcolor[RGB]{\colorfirst}94.81 & 66.71 & 82.45 & 80.94 \\
\text{Light3DGS}~\cite{lightgaussian}           & 84.34 & 79.64 & \cellcolor[RGB]{\colorfirst}91.08 & 88.67 & 87.49 & \cellcolor[RGB]{\colorsecond}94.76 & 67.63 & 83.09 & 82.33 \\
\text{2DGS}~\cite{2dgs}                         & \cellcolor[RGB]{\colorthird}84.43 & 78.54 & 89.71 & \cellcolor[RGB]{\colorthird}90.36 & \cellcolor[RGB]{\colorsecond}88.16 & \cellcolor[RGB]{\colorthird}94.59 & 68.63 & \cellcolor[RGB]{\colorsecond}83.87 & 81.61 \\
\noalign{\vskip 0.2em}\cdashline{1-10}\noalign{\vskip 0.2em}
\text{3DGS \textit{w.} $\mathcal{L}_{Moran}$}   & \cellcolor[RGB]{\colorsecond}84.54 & 80.20 & \cellcolor[RGB]{\colorsecond}90.84 & 90.15 & \cellcolor[RGB]{\colorthird}87.76 & 94.50 & 67.49 & \cellcolor[RGB]{\colorthird}83.47 & 81.89 \\
\text{SplatFields3D}                            & \cellcolor[RGB]{\colorfirst}86.62 & \cellcolor[RGB]{\colorsecond}84.05 & 89.56 & \cellcolor[RGB]{\colorfirst}93.62 & \cellcolor[RGB]{\colorfirst}89.53 & 94.50 & \cellcolor[RGB]{\colorfirst}74.14 & \cellcolor[RGB]{\colorfirst}85.03 & \cellcolor[RGB]{\colorthird}82.55 \\
\midrule
&\multicolumn{9}{c}{4 Input Views} \\
SparseNeRF~\cite{sparsenerf}                    & -     & \cellcolor[RGB]{\colorfirst}83.38 & 82.98 & \cellcolor[RGB]{\colorsecond}90.95 & \cellcolor[RGB]{\colorfirst}85.14 & 90.49 & -     & \cellcolor[RGB]{\colorsecond}76.46 & \cellcolor[RGB]{\colorsecond}79.48 \\
SparseNeRF \textit{wo.} depth                   & 78.66 & \cellcolor[RGB]{\colorsecond}78.80 & 80.86 & \cellcolor[RGB]{\colorthird}90.57 & 82.26 & 72.23 & \cellcolor[RGB]{\colorsecond}69.29 & 71.97 & \cellcolor[RGB]{\colorfirst}83.28 \\
\text{SuGaR}~\cite{sugar}                       & 75.61 & 72.07 & 83.08 & 83.68 & 80.66 & 82.48 & 63.46 & 70.73 & 68.69 \\
\text{ScaffoldGS}~\cite{scaffoldgs}             & 74.99 & 68.04 & 84.68 & 76.92 & 77.58 & 90.24 & 59.60 & 71.20 & 71.68 \\
\text{Mip3DGS}~\cite{mipgs}                     & 77.67 & 71.47 & 85.73 & 82.68 & 81.00 & 91.42 & 61.17 & 74.27 & 73.62 \\
\text{3DGS}~\cite{GaussianSplatting}            & 78.12 & 72.17 & 85.97 & 83.78 & 81.49 & 91.55 & 62.05 & 74.80 & 73.18 \\
\text{Light3DGS}~\cite{lightgaussian}           & \cellcolor[RGB]{\colorthird}79.38 & 73.58 & \cellcolor[RGB]{\colorsecond}86.93 & 85.98 & 81.91 & \cellcolor[RGB]{\colorsecond}92.25 & 63.27 & 75.88 & 75.22 \\
\text{2DGS}~\cite{2dgs}                         & 79.26 & 72.84 & 86.04 & 86.62 & \cellcolor[RGB]{\colorthird}82.54 & \cellcolor[RGB]{\colorthird}92.04 & \cellcolor[RGB]{\colorthird}63.57 & \cellcolor[RGB]{\colorthird}76.25 & 74.14 \\
\noalign{\vskip 0.2em}\cdashline{1-10}\noalign{\vskip 0.2em}
\text{3DGS \textit{w.} $\mathcal{L}_{Moran}$}   & \cellcolor[RGB]{\colorsecond}79.65 & 73.79 & \cellcolor[RGB]{\colorfirst}87.05 & 87.79 & 82.01 & \cellcolor[RGB]{\colorfirst}92.39 & 63.31 & 75.95 & 74.88 \\
\text{SplatFields3D}                            & \cellcolor[RGB]{\colorfirst}82.26 & \cellcolor[RGB]{\colorthird}78.23 & \cellcolor[RGB]{\colorthird}86.17 & \cellcolor[RGB]{\colorfirst}92.10 & \cellcolor[RGB]{\colorsecond}83.85 & 91.92 & \cellcolor[RGB]{\colorfirst}70.40 & \cellcolor[RGB]{\colorfirst}78.67 & \cellcolor[RGB]{\colorthird}76.77 \\

\bottomrule
\end{tabular}
}
\end{table}

\textbf{Static reconstruction on Blender (Sec.~\ref{subsec:exp_static}).} 
We compare SplatFields with SparseNeRF~\cite{sparsenerf} and with recent 3DGS methods: SuGaR~\cite{sugar}, Mip3DGS~\cite{mipgs}, 3DGS~\cite{GaussianSplatting}, 2DGS~\cite{2dgs}, and Light3DGS~\cite{lightgaussian} on Blender~\cite{mildenhall2020nerf}. 
Mip3DGS~\cite{mipgs}, 3DGS~\cite{GaussianSplatting}, and 2DGS~\cite{2dgs} are run for 40k iterations like SplatFields, while Light3DGS~\cite{lightgaussian}, SuGaR~\cite{sugar}, and SparseNeRF~\cite{sparsenerf} are run with their default configurations as they have a particular training scheme. 
All of the methods are initialized from the randomly sampled points inside the visual hull of the objects and are further supervised with the mask loss implemented as the $\mathcal{L}_1$ distance between the ground truth and the rendered opacity akin to~\cite{qian20233dgs}. 

We further provide extended comparisons of Tab.~\ref{tab:exp_static_blender} in Tab.~\ref{tab:app:exp_static_blender}-\ref{tab_app:exp_static_blender} for varying number of views ranging from 4 to 12. 
Consistently with the main paper, SplatFields demonstrates superior metric reconstruction quality over the baseline methods across varying number of input views. 

\textbf{Static reconstruction on DTU (Sec.~\ref{subsec:exp_static}).} 
We compare with NeRF (VolRecon~\cite{volrecon}, ZeroRF~\cite{zerorf}) and splatting (3DGS~\cite{GaussianSplatting}, 2DGS~\cite{2dgs}) methods on DTU~\cite{DTU} on the task of 3-view reconstruction (Tab.~\ref{tab:app:dtu_comparison}). 
All of the baselines are run with the default configurations, with the difference that the splatting-based baselines adopt the mask loss for fair comparisons. 
\begin{table}
    \scriptsize
    \setlength{\tabcolsep}{1.25pt} %
    \centering
    \caption{\textbf{Static three-view reconstruction} on the DTU dataset~\cite{DTU}. 
    SplatFields demonstrates more accurate reconstructions compared to the NeRF- (VolRecon~\cite{volrecon}, ZeroRF~\cite{zerorf}) and splatting-based (3DGS~\cite{GaussianSplatting}, 2DGS~\cite{2dgs}) baselines; the displayed metric is PSNR$\uparrow$
    }
\resizebox{\textwidth}{!}{%
    \begin{tabular}{l|c|ccccccccccccccc}
    \toprule
    & \textit{mean} & \multicolumn{15}{c}{Scene ID Number (PSNR$\uparrow$)} \\
    & PSNR$\uparrow$ & 105 &    106 &       110 &       114 &       118 &       122 &       24 &        37 &        40 &        55 &        63 &        65 &        69 &        83 &        97 \\
    \hline
    VolRecon    & 11.42 & 9.03  & 15.19 & 16.45 & 11.66 & 17.96 & 18.29 & 7.86  & 6.30  & 7.86  & 12.90 & 6.54  & 10.06 & 14.84 & 7.77  & 8.64 \\
    ZeroRF      & 19.10 & \cellcolor[RGB]{\colorsecond}21.36 & 14.30 & \cellcolor[RGB]{\colorfirst}20.96 & 18.86 & \cellcolor[RGB]{\colorthird}19.24 & \cellcolor[RGB]{\colorsecond}23.45 & 15.78 & 15.23 & 16.06 & 21.02 & \cellcolor[RGB]{\colorsecond}23.62 & 18.31 & 15.05 & \cellcolor[RGB]{\colorfirst}24.17 & 19.13 \\
    3DGS        & \cellcolor[RGB]{\colorthird}19.40 & 20.07 & \cellcolor[RGB]{\colorthird}17.06 & 17.04 & \cellcolor[RGB]{\colorsecond}20.56 & 18.25 & 20.23 & \cellcolor[RGB]{\colorsecond}18.76 & \cellcolor[RGB]{\colorsecond}19.82 & \cellcolor[RGB]{\colorthird}18.29 & \cellcolor[RGB]{\colorthird}21.03 & 22.63 & \cellcolor[RGB]{\colorthird}20.02 & \cellcolor[RGB]{\colorthird}15.86 & 21.64 & \cellcolor[RGB]{\colorthird}19.76 \\
    2DGS        & \cellcolor[RGB]{\colorsecond}20.70 & \cellcolor[RGB]{\colorthird}21.25 & \cellcolor[RGB]{\colorfirst}19.23 & \cellcolor[RGB]{\colorthird}19.17 & \cellcolor[RGB]{\colorthird}19.90 & \cellcolor[RGB]{\colorsecond}19.75 & \cellcolor[RGB]{\colorthird}22.04 & \cellcolor[RGB]{\colorfirst}19.71 & \cellcolor[RGB]{\colorfirst}20.22 & \cellcolor[RGB]{\colorsecond}19.56 & \cellcolor[RGB]{\colorfirst}21.95 & \cellcolor[RGB]{\colorthird}23.16 & \cellcolor[RGB]{\colorsecond}22.37 & \cellcolor[RGB]{\colorfirst}17.64 & \cellcolor[RGB]{\colorthird}23.13 & \cellcolor[RGB]{\colorfirst}21.47 \\ 
    \noalign{\vskip 0.2em} \cdashline{1-17} \noalign{\vskip 0.2em}
    SplatFields        & \cellcolor[RGB]{\colorfirst}21.07 & \cellcolor[RGB]{\colorfirst}21.93 & \cellcolor[RGB]{\colorsecond}19.11 & \cellcolor[RGB]{\colorsecond}19.77 & \cellcolor[RGB]{\colorfirst}22.03 & \cellcolor[RGB]{\colorfirst}21.35 & \cellcolor[RGB]{\colorfirst}24.49 & \cellcolor[RGB]{\colorthird}18.43 & \cellcolor[RGB]{\colorsecond}19.82 & \cellcolor[RGB]{\colorfirst}19.67 & \cellcolor[RGB]{\colorsecond}21.45 & \cellcolor[RGB]{\colorfirst}23.79 & \cellcolor[RGB]{\colorfirst}22.64 & \cellcolor[RGB]{\colorsecond}17.54 & \cellcolor[RGB]{\colorsecond}23.52 & \cellcolor[RGB]{\colorsecond}20.52 \\
    \bottomrule
        \end{tabular}
}
    \label{tab:app:dtu_comparison}
\end{table}

\begin{table}[t]
\scriptsize
  \caption{
    \textbf{Monocular reconstruction} of dynamic sequences from the NeRF-DS dataset~\cite{nerfds} with recent state-of-the-art methods.  
    The forward slash in FPS indicates the rendering speed with the inference of neural network \textit{vs.} without when the rendering primitives are extracted and stored for each frame
  }
      \vspace{-1em}
\label{tab_app:exp_nerfds}
  \centering
  \setlength{\tabcolsep}{2.7pt} %

\resizebox{\textwidth}{!}{%
\begin{tabular}{@{}l|cc|cccccccc@{}}
\toprule

&\multicolumn{2}{c}{Resources}& \multicolumn{8}{c}{LPIPS$\downarrow$ ($\times 10^2$)} \\
& FPS $\uparrow$ & t $\downarrow$ & \textit{mean} & Sieve & Plate & Bell & Press & Cup & As & Basin \\
3D-GS~\cite{GaussianSplatting}           &120+&15 min& 29.20 & 22.47 & 40.93 & 25.03 & 29.04 & 25.48 & 29.94 & 31.53 \\ 
TiNeuVox~\cite{TiNeuVox}        &< 1 & 30 min & 27.66 & 31.76 & 33.17 & 25.68 & 30.01 & 36.43 & 39.67 & 26.90 \\ 
4DGaussians~\cite{wu4dgaussiansRealTime} &120+/50&30 min & 21.06 & 16.39 & \cellcolor[RGB]{\colorthird}23.80 & 21.84 & 21.68 & 19.06 & 22.06 & 22.57 \\
HyperNeRF~\cite{HyperNeRF}       &< 1&1 day& 19.90 & 16.45 & 29.40 & 20.52 & \cellcolor[RGB]{\colorsecond}19.59 & \cellcolor[RGB]{\colorthird}16.50 & \cellcolor[RGB]{\colorthird}17.77 & \cellcolor[RGB]{\colorthird}19.11 \\ 

Deformable3DGS~\cite{yang2023deformable3dgs} &120+/30& 1 h & \cellcolor[RGB]{\colorthird}19.79 & \cellcolor[RGB]{\colorsecond}15.30 & 25.04 & \cellcolor[RGB]{\colorfirst}15.93 & 29.89 & \cellcolor[RGB]{\colorfirst}15.38 & 17.88 & \cellcolor[RGB]{\colorsecond}19.10 \\ 
NeRF-DS~\cite{nerfds} &< 1&1 day & \cellcolor[RGB]{\colorsecond}18.16 & \cellcolor[RGB]{\colorfirst}14.72 & \cellcolor[RGB]{\colorfirst}19.96 & \cellcolor[RGB]{\colorthird}18.67 & \cellcolor[RGB]{\colorthird}20.47 & 17.37 & \cellcolor[RGB]{\colorfirst}17.41 & \cellcolor[RGB]{\colorfirst}18.55 \\ \noalign{\vskip 0.2em} \cdashline{1-11} \noalign{\vskip 0.2em}
SplatFields4D  &120+/30&1 h& \cellcolor[RGB]{\colorfirst}17.86 & \cellcolor[RGB]{\colorfirst}14.72 & \cellcolor[RGB]{\colorsecond}22.43 & \cellcolor[RGB]{\colorsecond}16.10 & \cellcolor[RGB]{\colorfirst}19.26 & \cellcolor[RGB]{\colorsecond}15.67 & \cellcolor[RGB]{\colorsecond}17.71 & \cellcolor[RGB]{\colorthird}19.11 \\ 

\midrule

& \multicolumn{2}{c}{} & \multicolumn{8}{c}{PSNR$\uparrow$} \\
& FPS $\uparrow$ & t $\downarrow$ & \textit{mean} & Sieve & Plate & Bell & Press & Cup & As & Basin \\
3D-GS~\cite{GaussianSplatting}           &120+&15 min& 20.29                              & 23.16                                & 16.14                             & 21.01                             & 22.89                             & 21.71                             & 22.69                             & 18.42 \\
TiNeuVox~\cite{TiNeuVox}        &< 1 & 30 min& 21.61                              & 21.49                                & \cellcolor[RGB]{\colorfirst}20.58 & 23.08                             & 24.47                             & 19.71                             & 21.26                             & \cellcolor[RGB]{\colorfirst}20.66 \\
4DGaussians &120+/30&30 min &23.68 &     26.77 & 20.51 & 24.25 & 25.55 & 23.69 & 25.50 & 19.47 \\
HyperNeRF~\cite{HyperNeRF}       &< 1&1 day& 23.45                              & \cellcolor[RGB]{\colorsecond}25.43   & 18.93                             & 23.06                             & \cellcolor[RGB]{\colorfirst}26.15 & \cellcolor[RGB]{\colorthird}24.59 & \cellcolor[RGB]{\colorthird}25.58 & \cellcolor[RGB]{\colorsecond}20.41 \\
Deformable3DGS~\cite{yang2023deformable3dgs}  &120+/30&1 h& \cellcolor[RGB]{\colorthird}23.54  & 25.16                                & 19.97                             & \cellcolor[RGB]{\colorsecond}25.02& 24.18                             & \cellcolor[RGB]{\colorsecond}24.64& \cellcolor[RGB]{\colorfirst}26.26 & 19.57 \\ 
NeRF-DS~\cite{nerfds} &< 1&1 day & \cellcolor[RGB]{\colorsecond}23.60 & \cellcolor[RGB]{\colorfirst}25.78    & \cellcolor[RGB]{\colorsecond}20.54& \cellcolor[RGB]{\colorthird}23.19 & \cellcolor[RGB]{\colorsecond}25.72& \cellcolor[RGB]{\colorfirst}24.91 & 25.13 & 19.96 \\ \noalign{\vskip 0.2em} \cdashline{1-11} \noalign{\vskip 0.2em}
SplatFields4D  &120+/30&1 h& \cellcolor[RGB]{\colorfirst}23.84  & \cellcolor[RGB]{\colorthird}25.35    & \cellcolor[RGB]{\colorthird}20.36& \cellcolor[RGB]{\colorfirst}25.51 & \cellcolor[RGB]{\colorthird}25.43 & 24.29                             & \cellcolor[RGB]{\colorsecond}26.21& \cellcolor[RGB]{\colorthird}19.71 \\

\midrule
&&& \multicolumn{8}{c}{SSIM$\uparrow$} \\
& FPS $\uparrow$ & t $\downarrow$ & \textit{mean} & Sieve & Plate & Bell & Press & Cup & As & Basin \\
3D-GS~\cite{GaussianSplatting}          &120+&15 min& 78.16 & 82.03 & 69.70 & 78.85 & 81.63 & 83.04 & 80.17 & 71.70 \\
TiNeuVox~\cite{TiNeuVox}       &< 1 & 30 min & 82.34 & 82.65 & \cellcolor[RGB]{\colorsecond}80.27 &  \cellcolor[RGB]{\colorthird}82.42& 86.13 & 81.09 & 82.89 & \cellcolor[RGB]{\colorthird}81.45 \\
4DGaussians &120+/30&30 min &83.22 &     87.18 & 79.70 & 81.14 & 85.73 & 86.46 & 85.73 & 76.62 \\
HyperNeRF~\cite{HyperNeRF}      &< 1 &1 day& \cellcolor[RGB]{\colorthird}84.88  &  \cellcolor[RGB]{\colorsecond}87.98& 77.09 & 80.97 & \cellcolor[RGB]{\colorfirst}88.97 &  \cellcolor[RGB]{\colorthird}87.70& \cellcolor[RGB]{\colorfirst}89.49 & \cellcolor[RGB]{\colorfirst}81.99 \\
Deformable3DGS~\cite{yang2023deformable3dgs} &120+/30& 1 h & 84.05 & 87.58 & 79.14 & \cellcolor[RGB]{\colorsecond}84.52& 81.22 & \cellcolor[RGB]{\colorsecond}88.71& \cellcolor[RGB]{\colorthird}88.49 & 78.69 \\ 
NeRF-DS~\cite{nerfds} &< 1&1 day & \cellcolor[RGB]{\colorsecond}84.94 & \cellcolor[RGB]{\colorfirst}89.00 & \cellcolor[RGB]{\colorfirst}80.42 & 82.12 & \cellcolor[RGB]{\colorthird}86.18 & 87.41 & 87.78 & \cellcolor[RGB]{\colorsecond}81.66 \\ \noalign{\vskip 0.2em} \cdashline{1-11} \noalign{\vskip 0.2em}
SplatFields4D &120+/30&1 h& \cellcolor[RGB]{\colorfirst}85.17 & \cellcolor[RGB]{\colorthird}87.78  & \cellcolor[RGB]{\colorthird}80.26 & \cellcolor[RGB]{\colorfirst}84.74 & \cellcolor[RGB]{\colorsecond}86.64 & \cellcolor[RGB]{\colorfirst}88.73 & \cellcolor[RGB]{\colorsecond}88.59 & 79.44 \\

\bottomrule
\end{tabular}
}
\end{table}

\textbf{Monocular dynamic reconstruction (Sec.~\ref{subsec:exp_dyn}).}
Our method adopts annealing smooth training \cite{yang2023deformable3dgs} and is trained for 30k iterations after being initialized from static 3DGS ran for 3k iterations akin to \cite{yang2023deformable3dgs}. 
We run recent dynamic 3DGS methods \cite{wu4dgaussiansRealTime,yang2023deformable3dgs} with their default configurations, while results for the NeRF-based methods are adopted from the previous work \cite{nerfds,yang2023deformable3dgs}. 

We further provide additional metrics SSIM and PSNR in Tab.~\ref{tab_app:exp_nerfds}. Note that SSIM and PSNR are less reliable metrics due to noisy camera calibrations.

\begin{table}
  \caption{
    \textbf{Multi-view reconstruction} of dynamic sequences from the Owlii dataset ~\cite{mildenhall2020nerf} under varying number of input views. The reported metric is SSIM$\uparrow$ averaged across novel views. 
    See Sec.~\ref{subsec:exp_dyn} for discussion
  }
  \label{tab_app:exp_dyn_owlii}
  \centering
  \scriptsize
  \setlength{\tabcolsep}{3.5pt} %

\begin{tabular}{@{}l|ccccc@{}}
\toprule
&\multicolumn{5}{c}{10 Input Views}\\
 & \textit{mean} & Dancer &     Exercise &      Model & Basketball \\
4D-GS~\cite{yang4DGS} &\cellcolor[RGB]{\colorthird}95.34 &  \cellcolor[RGB]{\colorsecond}95.31 & \cellcolor[RGB]{\colorthird}95.96 & \cellcolor[RGB]{\colorthird}94.92 & \cellcolor[RGB]{\colorthird}95.16 \\
Deformable3DGS~\cite{yang2023deformable3dgs} &93.80 &   94.10 & 95.09 & 91.58 & 94.43 \\
4DGaussians~\cite{wu4dgaussiansRealTime} &\cellcolor[RGB]{\colorsecond}95.91 &      \cellcolor[RGB]{\colorthird}95.19 & \cellcolor[RGB]{\colorsecond}96.47 & \cellcolor[RGB]{\colorsecond}95.71 & \cellcolor[RGB]{\colorsecond}96.28 \\\noalign{\vskip 0.2em} \cdashline{1-6} \noalign{\vskip 0.2em}
SplatFields4D (30k it) &96.52 & 96.41 & 96.72 & 95.99 & 96.98 \\
SplatFields4D (40k it) &96.57 & 96.47 & 96.76 & 96.04 & 97.02 \\
SplatFields4D (100k it) &96.67 & 96.59 & 96.83 & 96.16 & 97.11 \\
SplatFields4D (200k it) &\cellcolor[RGB]{\colorfirst}96.81 &  \cellcolor[RGB]{\colorfirst}96.76 & \cellcolor[RGB]{\colorfirst}96.92 & \cellcolor[RGB]{\colorfirst}96.32 & \cellcolor[RGB]{\colorfirst}97.23 \\
\midrule
&\multicolumn{5}{c}{8 Input Views} \\
 & \textit{mean} & Dancer &     Exercise &      Model & Basketball \\
4D-GS~\cite{yang4DGS} &\cellcolor[RGB]{\colorthird}93.71 &  \cellcolor[RGB]{\colorthird}94.19 & \cellcolor[RGB]{\colorthird}93.94 & \cellcolor[RGB]{\colorthird}93.29 & \cellcolor[RGB]{\colorthird}93.40 \\
Deformable3DGS~\cite{yang2023deformable3dgs} &92.37 &   93.24 & 93.29 & 90.33 & 92.62 \\
4DGaussians~\cite{wu4dgaussiansRealTime} &\cellcolor[RGB]{\colorsecond}95.00 &       \cellcolor[RGB]{\colorsecond}94.39 & \cellcolor[RGB]{\colorsecond}95.45 & \cellcolor[RGB]{\colorsecond}94.92 & \cellcolor[RGB]{\colorsecond}95.26 \\\noalign{\vskip 0.2em} \cdashline{1-6}  \noalign{\vskip 0.2em}
SplatFields4D (30k it) &95.99 & 95.97 & 96.05 & 95.44 & 96.52 \\
SplatFields4D (40k it) &96.04 & 96.02 & 96.08 & 95.49 & 96.56 \\
SplatFields4D (100k it) &96.15 &        96.15 & 96.16 & 95.62 & 96.65 \\
SplatFields4D (200k it) &\cellcolor[RGB]{\colorfirst}96.28 &        \cellcolor[RGB]{\colorfirst}96.31 & \cellcolor[RGB]{\colorfirst}96.26 & \cellcolor[RGB]{\colorfirst}95.78 & \cellcolor[RGB]{\colorfirst}96.77 \\
\midrule
&\multicolumn{5}{c}{6 Input Views} \\
 & \textit{mean} & Dancer &     Exercise &      Model & Basketball \\
4D-GS~\cite{yang4DGS} &87.23 &  89.52 & 87.05 & 85.16 & 87.20 \\
Deformable3DGS~\cite{yang2023deformable3dgs} &\cellcolor[RGB]{\colorthird}90.95 &   \cellcolor[RGB]{\colorthird}91.73 & \cellcolor[RGB]{\colorthird}91.48 & \cellcolor[RGB]{\colorthird}89.11 & \cellcolor[RGB]{\colorthird}91.48 \\
4DGaussians~\cite{wu4dgaussiansRealTime} &\cellcolor[RGB]{\colorsecond}93.87 &       \cellcolor[RGB]{\colorsecond}93.58 & \cellcolor[RGB]{\colorsecond}94.45 & \cellcolor[RGB]{\colorsecond}93.05 & \cellcolor[RGB]{\colorsecond}94.40 \\\noalign{\vskip 0.2em} \cdashline{1-6} \noalign{\vskip 0.2em}
SplatFields4D (30k it) &95.40 & 95.62 & 95.51 & 94.53 & 95.95 \\
SplatFields4D (40k it) &95.45 & 95.67 & 95.54 & 94.59 & 95.99 \\
SplatFields4D (100k it) &95.56 &        95.81 & 95.62 & 94.70 & 96.09 \\
SplatFields4D (200k it) &\cellcolor[RGB]{\colorfirst}95.69 &        \cellcolor[RGB]{\colorfirst}95.99 & \cellcolor[RGB]{\colorfirst}95.71 & \cellcolor[RGB]{\colorfirst}94.85 & \cellcolor[RGB]{\colorfirst}96.21 \\
\midrule
&\multicolumn{5}{c}{4 Input Views} \\
 & \textit{mean} & Dancer &     Exercise &      Model & Basketball \\
4D-GS~\cite{yang4DGS} &78.94 &  80.81 & 78.72 & 78.09 & 78.15 \\
Deformable3DGS~\cite{yang2023deformable3dgs} &\cellcolor[RGB]{\colorthird}87.10 &   \cellcolor[RGB]{\colorthird}89.04 & \cellcolor[RGB]{\colorthird}87.67 & \cellcolor[RGB]{\colorthird}85.05 & \cellcolor[RGB]{\colorthird}86.63 \\
4DGaussians~\cite{wu4dgaussiansRealTime} &\cellcolor[RGB]{\colorsecond}89.50 &       \cellcolor[RGB]{\colorsecond}90.34 & \cellcolor[RGB]{\colorsecond}90.67 & \cellcolor[RGB]{\colorsecond}87.47 & \cellcolor[RGB]{\colorsecond}89.51 \\\noalign{\vskip 0.2em} \cdashline{1-6} \noalign{\vskip 0.2em}
SplatFields4D (30k it) &91.46 & 92.61 & 91.99 & 88.98 & 92.28 \\ 
SplatFields4D (40k it) &91.49 & 92.65 & 92.01 & 88.99 & 92.31 \\
SplatFields4D (100k it) &91.54 &        92.76 & 92.04 & 89.01 & 92.36 \\
SplatFields4D (200k it) &\cellcolor[RGB]{\colorfirst}91.60 &        \cellcolor[RGB]{\colorfirst}92.89 & \cellcolor[RGB]{\colorfirst}92.07 & \cellcolor[RGB]{\colorfirst}89.02 & \cellcolor[RGB]{\colorfirst}92.41 \\
\bottomrule

\end{tabular}

\end{table}

\textbf{Multi-view dynamic reconstruction (Sec.~\ref{subsec:exp_dyn})}.
Dynamic NeRF-based methods (DyNeRF~\cite{DyNeRF}, TNeRF~\cite{DyNeRF}, DNeRF~\cite{DNeRF}, HyperNeRF~\cite{HyperNeRF}) are trained with SDF parametrization as they are better suited for sparse view reconstruction. We use the implementations with ResField MLPs~\cite{ResFields} (256 neurons) and train them for 400k iterations, following the training scheme from \cite{ResFields}. 

For dynamic Gaussian splatting methods (4D-GS~\cite{yang4DGS}, Deformable3DGS~\cite{yang2023deformable3dgs}, 4DGaussians~\cite{wu4dgaussiansRealTime}), we use their default implementations and adopt additional mask loss with the weight of 0.1. All of these methods, including ours, are trained with a batch size of 5. 
4D-GS is trained with default 30k iterations. Deformable3DGS is trained until the full convergence of 200k iterations. 
4DGaussians is trained for 30k iterations, we noticed that longer training leads to overfitting and the loss becomes an invalid number. 

Additional SSIM metric is reported in Tab.~\ref{tab_app:exp_dyn_owlii}. 
Akin to the main paper, SplatFields demonstrates consistently better reconstruction quality across all scenes and varying number of input views. 

\textbf{Compute, memory overhead, and inference time.} %
Compared to the original 3DGS, our method requires longer training to converge ($\sim$10 min. for 3DGS vs. $\sim$70 min. for ours on the Toy scene) and consumes a greater amount of GPU memory ($\sim$5GB for 3DGS vs. about $\sim$8GB for ours). However, after training, the neural components can be discarded, \textit{leaving the inference speed and memory usage equivalent to that of 3DGS}. In dynamic setups, the training times of our method are comparable to other dynamic 3DGS methods that also employ neural networks. Our neural network architecture comprises $\sim$1M parameters for the static case. 
All the run-times reported in the paper are calculated on an NVIDIA RTX 3090. 

\textbf{CNNs \textit{vs.} MLPs on extremely sparse view setups.} CNN module enhances the capacity of the SplatFields, which may lead to slight overfitting in extremely sparse scenarios, such as a 4-view setup. However, as additional views are incorporated and the model receives more diverse inputs, the ability of CNNs to capture structural patterns become increasingly beneficial; this is demonstrated by the improved performance in denser view setups. Please also note that the CNN-based SplatFields model is still better than the vanilla 3DGS method in the 4-view setup (Tab.~\ref{tab:app:exp_static_blender}-\ref{tab_app:exp_static_blender}). 

\textbf{Spatial autocorrelation: sparse \textit{vs.} dense view setup. }  %
A simple experiment under different view setups on Toy~\cite{mildenhall2020nerf} demonstrates (Tab.~\ref{tab:app:moran}) a tendency that overfitting (high $\Delta$PSNR) corresponds to lower autocorrelation, especially for RGB.
\begin{table}
\centering
\caption{Moran's I on Toy~\cite{mildenhall2020nerf} for a varying number of views }
\begin{tabular}{lccccc}
\toprule
\#Views & 5 & 25 & 50 & 75 & 100 \\ \hline

Train PSNR  & 51.85 & 45.10 & 41.11 & 40.23 & 40.03 \\ 
Test PSNR $\uparrow$   & 18.07 & 30.12 & 33.75 & 35.02 & \textbf{35.34} \\ 
$\Delta$PSNR $\downarrow$  & \underline{33.78} & 14.98 &  7.36 & 5.21 & \textbf{4.69} \\ \hline

Moran RGB $\uparrow$    & \underline{0.467} & 0.588 & 0.634 & 0.655 & \textbf{0.661} \\
Moran Opacity $\uparrow$ & \underline{0.710} & 0.746 & 0.744 & \textbf{0.742} & 0.736 \\
Moran Covariance $\uparrow$   & 0.426 & \underline{0.414} & 0.452 & 0.465 & \textbf{0.476} \\

\bottomrule
\end{tabular}
\label{tab:app:moran}
\end{table}

\clearpage

\bibliographystyle{splncs04}
\bibliography{main}

\end{document}